\documentclass{article}

 \usepackage[preprint]{neurips_2025}

\usepackage[utf8]{inputenc} % allow utf-8 input
\usepackage[T1]{fontenc}    % use 8-bit T1 fonts
\usepackage{hyperref}       % hyperlinks
\usepackage{url}            % simple URL typesetting
\usepackage{booktabs}       % professional-quality tables
\usepackage{amsfonts}       % blackboard math symbols
\usepackage{nicefrac}       % compact symbols for 1/2, etc.
\usepackage{microtype}      % microtypography
\usepackage{xcolor}         % colors
\usepackage{amsmath}
\usepackage{amssymb}
\usepackage{amsthm}
\usepackage{enumitem}
\usepackage{colortbl}
\usepackage{bm}
\usepackage[ruled,linesnumbered]{algorithm2e}
\usepackage{graphicx}
\usepackage{comment}
\usepackage{multirow}
\usepackage{bigstrut}
\usepackage{rotating}
\usepackage{tabularx}
\usepackage{graphicx}
\usepackage{epigraph}
\usepackage{tikz-qtree}
\usetikzlibrary{trees,decorations.text,math,calc, positioning, arrows.meta}
\usepackage{tcolorbox}
\usepackage{changepage}
\usepackage{titlesec}
\usepackage{mdframed}
\usepackage{fvextra}
\tcbuselibrary{listings,breakable}
\definecolor{Red}{RGB}{190,20,42}
\definecolor{CY}{RGB}{233,246,254}

\newtcolorbox{textbox}{
    breakable,                 % 允许自动分页
    colback=gray!10,
    colframe=black,
    coltitle=black,
    rounded corners,
    left=10pt,
    right=10pt,
    top=10pt,
    bottom=10pt,
    before skip=10pt,
    after skip=10pt
}

\theoremstyle{definition}

\theoremstyle{theorem}

\theoremstyle{proof}

\theoremstyle{remark}

\titlespacing*{\section}{0pt}{4pt}{4pt}

\titlespacing*{\subsection}{0pt}{3pt}{3pt}

\titlespacing*{\subsubsection}{0pt}{3pt}{3pt}

\title{YuLan-OneSim: Towards the Next Generation of Social Simulator with Large Language Models}

\author{
    Lei Wang,~Heyang Gao,~Xiaohe Bo,~Xu Chen,~Ji-Rong Wen\\
    Gaoling School of Artificial Intelligence,
	Renmin University of China, Beijing, China\\
	\texttt{\{wanglei154, gaoheyang, xiaohe, xu.chen, jrwen\}@ruc.edu.cn}  \\
        $\vcenter{\vspace{0.3cm}\hbox{\includegraphics[width=1.1cm]{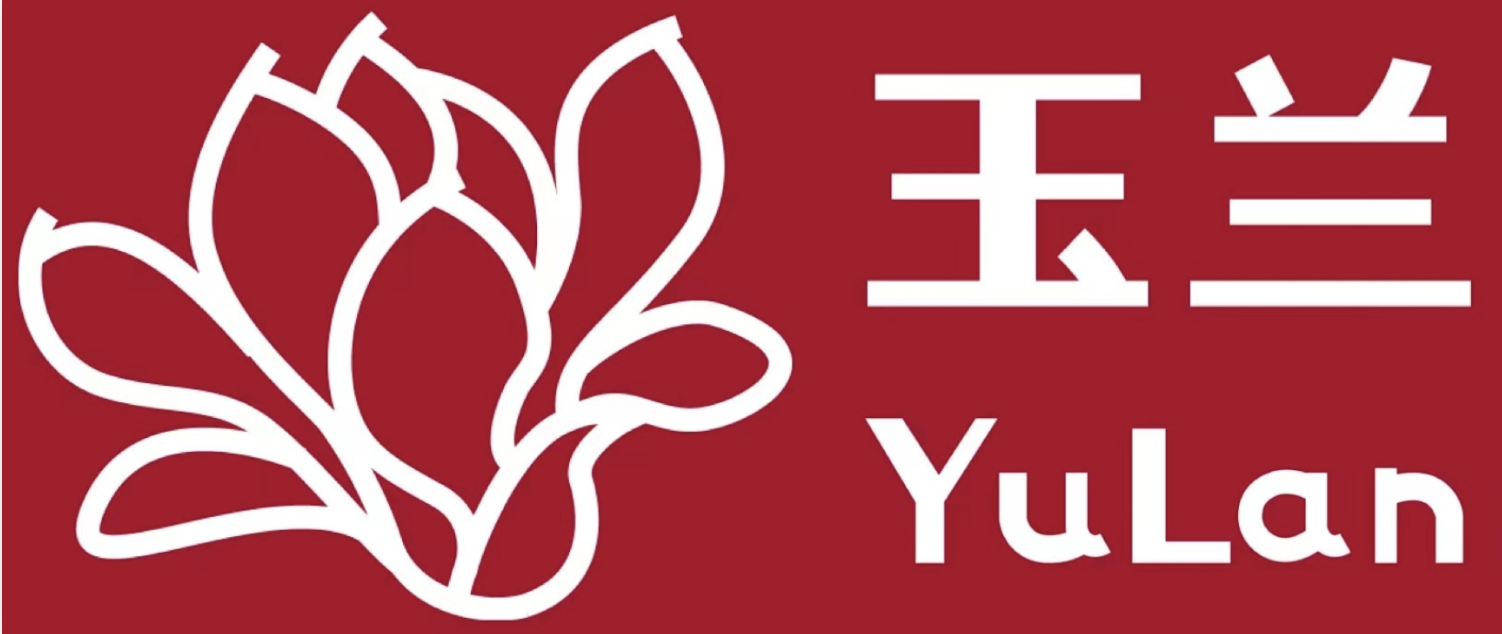}\hspace{0.1cm}}}$ 
        $\vcenter{\vspace{0.3cm}\hbox{\textbf{RUC YuLan Team}} }$\\
}

\begin{document}
	\maketitle
	\begin{abstract}
Leveraging large language model (LLM) based agents to simulate human social behaviors has recently gained significant attention. In this paper, we introduce a novel social simulator called YuLan-OneSim. Compared to previous works, YuLan-OneSim distinguishes itself in five key aspects:
(1) \textbf{Code-free scenario construction}: Users can simply describe and refine their simulation scenarios through natural language interactions with our simulator. All simulation code is automatically generated, significantly reducing the need for programming expertise.
(2) \textbf{Comprehensive default scenarios}: 
We implement 50 default simulation scenarios spanning 8 domains, including economics, sociology, politics, psychology, organization, demographics, law, and communication, broadening access for a diverse range of social researchers.
(3) \textbf{Evolvable simulation}: 
Our simulator is capable of receiving external feedback and automatically fine-tuning the backbone LLMs, significantly enhancing the simulation quality.
(4) \textbf{Large-scale simulation}: 
By developing a fully responsive agent framework and a distributed simulation architecture, our simulator can handle up to 100,000 agents, ensuring more stable and reliable simulation results.
(5) \textbf{AI social researcher}: 
Leveraging the above features, we develop an AI social researcher. Users only need to propose a research topic, and the AI researcher will automatically analyze the input, construct simulation environments, summarize results, generate technical reports, review and refine the reports—completing the social science research loop.
To demonstrate the advantages of YuLan-OneSim, we conduct experiments to evaluate the quality of the automatically generated scenarios, the reliability, efficiency, and scalability of the simulation process, as well as the performance of the AI social researcher.
To encourage further exploration in this field, we have fully released our project at https://github.com/RUC-GSAI/YuLan-OneSim.
\end{abstract}
	
	% % \begin{figure}[h]
	% % 	\centering
	% % 	\vspace{-0.2cm}
	% % 	\includegraphics[width=0.84\textwidth]{figure_abstract_cx_zeyu_0420.pdf}
	% % 	\vspace{-0.2cm}
	% % 	\caption{The importance of the memory module in LLM-based agents.}
	% % 	\vspace{-0.2cm}
	% % 	\label{imp}
	% % \end{figure}
	
	\setcounter{footnote}{0}
	\setcounter{tocdepth}{4}
	\clearpage
	\tableofcontents
	\clearpage

    \setlength{\epigraphwidth}{.98\textwidth}
	
    \section{Introduction}
    Social science has long played a critical role in the advancement of human civilization by providing profound insights into human behavior, societal structures, and cultural dynamics. For most social science studies, social simulation has emerged as a foundational methodology for investigating complex social behaviors and uncovering patterns that are often difficult to observe directly in real-world settings~\cite{squazzoni2014social}. 
    For decades, agent-based modeling (ABM) has been a prominent approach in social simulations~{\cite{heath2009survey,bianchi2015agent}}. However, this approach often fails to capture the intricacies of human cognitive processes and language-mediated interactions~{\cite{gao2024large}}.

    Recently, the advent of large language models (LLMs) has opened new avenues for advancing social simulation. 
    In particular, numerous studies have demonstrated that language is a powerful medium for representing human cognition. 
    As a result, by training on large-scale language corpora, LLMs can exhibit human-like intelligence across a wide range of tasks~{\cite{zhao2023survey,achiam2023gpt}}.     
    Motivated by these advancements, researchers have increasingly explored leveraging LLMs to build more realistic language-based social simulators. For example, GenSim introduces a general-purpose social simulation platform by providing essential and generic functions that enable users to create customized scenarios~\cite{gensim}.
    OASIS develops a large-scale social media simulator supporting millions of agents, similar to platforms like Twitter or Reddit~\cite{oasis}. 
    AgentSociety constructs a virtual urban environment where agents can simulate various social and economic activities in city-based scenarios~\cite{agentsociety}. 

    In this paper, we introduce a novel LLM agent based social simulator, called ``YuLan-OneSim''.
    % \footnote{The name of our simulator comes from the concepts of Yi and Xiang from the Book of Changes, which represents the wisdom of ancient Chinese people in predicting and analyzing social phenomena.
    % Our work, as an attempt to simulate society in the era of large language models, shares a similar essence with the ancient wisdom.}.
    Compared to previous works, our simulator has five key advantages:

    $\bullet$ \textbf{Code-free scenario construction}:
    To make our simulator more accessible to social science researchers, we enable users to construct simulation scenarios directly using natural language, substantially reducing the need for programming expertise.
    % The framework follows five key steps: requirement communication, agent type extraction, workflow extraction, code generation, and scenario data generation.
    
    $\bullet$ \textbf{Comprehensive default scenarios}: 
    To further support users in the social sciences, we implement a comprehensive scenario repository comprising 50 pre-defined scenarios spanning eight domains, including economics, sociology, politics, psychology, organizational studies, demography, law, and communication.

    $\bullet$ \textbf{Evolvable simulation}: 
    Previous LLM agent based simulators often fail to incorporate system or human feedback to improve simulation realism. To address this limitation, we propose a multi-agent framework, called Verifier–Reasoner–Refiner–Tuner (VR²T), to effectively integrate external feedback and evolve the backbone LLMs.

    $\bullet$ \textbf{Large-scale simulation}: 
    Supporting large-scale simulation is essential for achieving realistic and stable outcomes. To this end, we develop a fully responsive agent framework and a distributed simulation architecture tailored for social simulations. This design inherently supports parallel computing and enables the simultaneous simulation of up to 100,000 agents.

    $\bullet$ \textbf{AI social researcher}: 
    To complete the social science research loop, we develop an AI social researcher. Users only need to provide a research topic, and the AI researcher can autonomously analyze the input, construct simulation scenarios, summarize key findings, generate technical reports, and review and refine the reports.

    To realize the above features, our simulator is built based on four subsystems: the scenario auto-construction subsystem, simulation subsystem, feedback-driven evolving subsystem, and the AI social researcher subsystem. To begin with, the scenario auto-construction subsystem translates user requirements into executable code that directly drives the simulation. Based on this code, the simulation subsystem performs the corresponding simulation tasks and continuously oversees and evaluates the simulation process in real time. 
    When simulation results are found to be unreliable, the feedback-driven evolving subsystem incorporates external feedback and retrains the underlying LLMs to improve the overall performance. Built upon these subsystems, the AI social researcher can autonomously complete the entire social science research loop—from idea generation and simulation scenario construction to results analysis and report generation and refinement.

    In our experiments, we first assess the quality of the automatically generated simulation scenarios. Next, we evaluate the reliability, efficiency, and scalability of the simulation process. Finally, we examine the performance of the AI social researcher.
    In summary, the key contributions of this paper are as follows:

    $\bullet$ We develop a novel social simulator named "YuLan-OneSim", which supports code-free scenario construction, a comprehensive set of default scenarios, and evolvable and large-scale simulations.

    $\bullet$ Based on YuLan-OneSim, we develop an AI social researcher capable of autonomously completing the entire social science research loop—from initial idea generation to final report production.
    
    $\bullet$ To assess the efficiency and effectiveness of our simulator, we conduct extensive experiments to evaluate the performance of its different subsystems.

    $\bullet$ We have fully released our project at {https://github.com/RUC-GSAI/YuLan-OneSim}, aiming to advance LLM-based social simulation and deepen the integration between social science and artificial intelligence.

    In the following sections, we first review related work to contextualize the key contributions of YuLan-OneSim.
    We then present the details of our simulator, outlining the motivations and design choices behind its core subsystems.
    Subsequently, we conduct extensive experiments to evaluate the effectiveness and efficiency of our simulator.
    Finally, we conclude the paper and discuss potential directions for future improvements.

    \section{Related Work}

\subsection{Traditional Social Simulation}
    Social simulation has long been a valuable tool for uncovering and understanding social phenomena. Social science researchers have developed numerous classical rule-based models targeting specific scenarios.
    For example, Schelling~\cite{schelling1971dynamic} proposed a residential segregation model to simulate individuals’ willingness to relocate based on the racial composition of their neighbors. The results demonstrated that even when individuals had a relatively high tolerance for neighbors of different races, racial segregation still emerged at the macro level.
    Similarly, Axelrod~\cite{axelrod1997dissemination} introduced a cultural dissemination model, in which 100 agents, each possessing a set of cultural features, interacted with their neighbors and exchanged traits. The model revealed that, under certain conditions, local convergence could coexist with global cultural diversity, resulting in a stable multi-cultural equilibrium.
    Palmer et al.~\cite{palmer1999artificial} developed the artificial stock market, in which dozens of trading agents used genetic algorithms to learn and predict stock prices. Their interactions were found to generate realistic price fluctuations and speculative bubbles, highlighting how simple adaptive behaviors could give rise to complex market dynamics.
    In addition, RecSim~\cite{ie2019recsim} is a platform for simulating online user behaviors.
    It enables researchers to evaluate the long-term performance of recommendation algorithms within a controlled, simulated setting.

    While the above simulation methods have allowed researchers to observe and analyze various important social phenomena, they mostly rely on heuristic rules, which constrain the simulation outcomes to limited predefined spaces. Moreover, the agents' behaviors are typically driven by simplistic functions trained on small datasets, making it difficult to capture the complexity of real human cognitive processes.

    \subsection{LLM Agent based Social Simulation}
    Due to the strong capabilities of LLMs in simulating human behavior, recent years have seen a surge in studies exploring LLM-based agents for building social simulators.
    Chronologically, we categorize the work in this field into two stages. In the first stage, researchers focused on simulating specific scenarios, where the number of agents is typically smaller than 1,000.
    For example, Generative Agents~\cite{park2023generative} simulates 25 agents' daily life in a small town to observe their social behaviors. RecAgent~\cite{wang2025user} builds a virtual web environment and simulates 1000 users' recommendation, one-to-one chatting, and one-to-many posting behaviors. 
    S$^3$\cite{gao2023s3} simulates social networks by observing individual and group-level discussions and reactions to public events. EconAgent\cite{li2023econagent} employs 100 agents making decisions that drive various market operations, simulating macroeconomic phenomena. WarAgent~\cite{hua2023war} uses dozens of agents to simulate historical conflict relationships and war-related behaviors between nations.
    TrendSim~\cite{zhang2024trendsim} utilizes 1,000 agents to simulate the impact of poisoning attacks on trending topics in social media. Agent Hospital~\cite{li2024agent} models hospital operations with 50 agents, including doctors, nurses, and patients, simulating medical practice processes. FinCon~\cite{yu2024fincon} simulates real investment company structures through manager and analyst agents making investment decisions in financial markets. Agent4Rec~\cite{zhang2024generative} employs 1,000 agents to simulate users in recommendation systems, interacting with the system and evaluating recommendation algorithm effectiveness. UGI~\cite{xu2023urban} provides an embodied urban platform simulating agent transportation, economic activities, and social interactions within cities.
    % Social Simulacra~\cite{park2022social} creates a prototype of a populated community to study interactions among people in specified scenarios. 
    Chuang et al.\cite{chuang2023simulating} utilize LLM agents to simulate topic discussions and study opinion dynamics. 
    % Wang et al.\cite{wang2025investigating} simulate exchange behaviors between agents to study and extend social exchange theory. 
    Acerbi et al.\cite{acerbi2023large} observed human-like bias in information propagation processes simulated by LLM agents. ElectionSim\cite{zhang2024electionsim} designed a simulation framework to predict U.S. presidential election results.

    As research in this field continues to progress, two critical challenges have emerged: the generality and scalability of social simulators. Consequently, in the second stage, researchers have proposed several promising general-purpose social simulators capable of supporting large-scale agent populations.
    
    For example, GenSim~\cite{gensim} is a general simulation platform equipped with function sets facilitating customization of simulation scenarios and supporting up to 100,000 agents. AgentScope~\cite{gao2024agentscope} stands as a comprehensive multi-agent framework offering exceptional flexibility and scalability, demonstrating its robustness through successful implementation of simulations involving up to one million agents~\cite{pan2024very}.
    OASIS~\cite{oasis} also enables 1 million agents to simulate social media interactions in parallel. AgentSociety~\cite{agentsociety} builds a virtual city with real geographic information where 10,000 agents simulate urban activities such as polarization, information propagation, economic behaviors, and disaster response. SocioVerse~\cite{zhang2025socioverse} includes a user profile engine covering 10 million people, supporting thousands of agents simultaneously simulating survey responses to elections, news events, and economic inquiries. SOTOPIA~\cite{zhou2023sotopia} is an interactive open platform supporting agent-human interactions across diverse rich scenarios to evaluate agents' social performance. SOTOPIA-S4~\cite{zhou2025sotopia} further extends this by allowing users to describe and create agents for simulated dialogue scenarios using natural language.

    Unlike the studies mentioned above, YuLan-OneSim aims to advance LLM-based social simulation to the next stage—where simulation scenarios can be constructed automatically, the simulator can autonomously evolve through feedback integration, and the entire social science research cycle can be completed by an AI social researcher with minimal human input. A detailed comparison between YuLan-OneSim and previous studies is provided in Table~\ref{tab:related-work}.

\begin{table}[t]
\centering
\small
\caption{Comparison between YuLan-OneSim and previous works.
In the "\# Agents" column, we present the number of agents as reported in the original papers' experiments. 
In the "\# Environments (Env)" column, "-" means that the simulator does not provide environments for the agents to take actions, for example, the simulator is purely based on conversations.
The symbol $\bigcirc$ indicates that while SOTOPIA-S4 allows users to configure simulation scenarios using natural language, it leverages user inputs as LLM prompts instead of converting them into executable simulation code.
We use different colors to represent various stages in the field of LLM agent based social simulation.
}

\renewcommand\arraystretch{1.4}
\label{tab:related-work}
\scalebox{.9}{
\begin{tabular}{>{\columncolor{lightgray!25}}m{1.2cm}<{\centering}|
m{4.6cm}<{\centering}|
m{1.4cm}<{\centering}|
m{1.6cm}<{\centering}|
m{.8cm}<{\centering}|
m{1.5cm}<{\centering}|
m{1.5cm}<{\centering}}
\hline\hline
\rowcolor{lightgray!15}
{Date} & {Name \& Reference} & {Automatic Coding} & {\# Agents} & {\# Env} & {AI Social Researcher} & {Feedback Compatible} \\
\hline
% \rowcolor{yellow!10}
% 2022.08 & Social Simulacra~\cite{park2022social} & $\times$ & 100-1000 & 1 & $\times$ & $\times$ \\
\rowcolor{yellow!10}
2023.04 & Generative Agents~\cite{park2023generative} & $\times$ & 10-100 & 1 & $\times$ & $\times$ \\
\rowcolor{yellow!10}
2023.06 & RecAgent~\cite{wang2025user} & $\times$ & 1000-10000 & 3 & $\times$ & $\times$ \\
\rowcolor{yellow!10}
2023.07 & S$^3$~\cite{gao2023s3} & $\times$ & 1000-10000 & 1 & $\times$ & $\times$ \\
\rowcolor{yellow!10}
2023.10 & Agent4Rec~\cite{zhang2024generative} & $\times$ & 1000-10000 & 1 & $\times$ & $\times$ \\
\rowcolor{yellow!10}
2023.10 & EconAgent~\cite{li2023econagent} & $\times$ & 10-100 & 1 & $\times$ & $\times$ \\
% \rowcolor{yellow!10}
% 2023.10 & Acerbi et al.~\cite{acerbi2023large} & $\times$ & 10-100 & 1 & $\times$ & $\times$ \\
\rowcolor{yellow!10}
2023.11 & WarAgent~\cite{hua2023war} & $\times$ & 10-100 & 1 & $\times$ & $\times$ \\
% \rowcolor{yellow!10}
% 2023.10 & SOTOPIA~\cite{zhou2023sotopia} & $\times$ & 10-100 & 13 & $\times$ & $\times$ \\
\rowcolor{yellow!10}
2023.11 & MedAgents~\cite{tang2023medagents} & $\times$ & 10-100 & 1 & $\times$ & $\times$ \\
% \rowcolor{yellow!10}
% 2023.11 & Chuang et al.~\cite{chuang2023simulating} & $\times$ & 10-100 & 1 & $\times$ & $\times$ \\
\rowcolor{yellow!10}
2023.12 & UGI~\cite{xu2023urban} & $\times$ & 100-1000 & 5 & $\times$ & $\times$ \\
\rowcolor{yellow!10}
2024.05 & Agent Hospital~\cite{li2024agent} & $\times$ & 10-100 & 1 & $\times$ & $\times$ \\
\rowcolor{yellow!10}
2024.07 & FinCon~\cite{yu2024fincon} & $\times$ & 10-100 & 1 & $\times$ & $\times$ \\
\rowcolor{yellow!10}
2024.10 & TrendSim~\cite{zhang2024trendsim} & $\times$ & 1000-10000 & 1 & $\times$ & $\times$ \\
\rowcolor{yellow!10}
2024.10 & ElectionSim~\cite{zhang2024electionsim} & $\times$ & 1000-10000 & 1 & $\times$ & $\times$ \\
% \rowcolor{yellow!10}
% 2025.02 & Wang et al.~\cite{wang2025investigating} & $\times$ & 10-100 & 1 & $\times$ & $\times$ \\
\hline
% \rowcolor{blue!10}
% 2024.05 & AgentScope~\cite{gao2024agentscope} & $\times$ & $\geq 10000$ & 1 & $\times$ &  $\times$ \\
\rowcolor{blue!10}
2024.10 & GenSim~\cite{gensim} & $\times$ & $\geq 10000$ & 4 & $\times$ & \checkmark \\
\rowcolor{blue!10}
2024.11 & OASIS~\cite{oasis} & $\times$ & $\geq 10000$ & 2 & $\times$ & $\times$ \\
\rowcolor{blue!10}
2025.02 & AgentSociety~\cite{agentsociety} & $\times$ & $\geq 10000$ & 4 & $\times$ & $\times$ \\
\rowcolor{blue!10}
% 2025.02 & ZJUCSS~\cite{ZJUCSS} & $\times$ & $>$10,000 & Multiple & $\times$ & $\times$ \\
\rowcolor{blue!10}
2025.04 & SocioVerse~\cite{zhang2025socioverse} & $\times$ & $\geq 10000$ & 3 & $\times$ & $\times$ \\
\rowcolor{blue!10}
2025.04 & SOTOPIA-S4~\cite{zhou2025sotopia} & $\bigcirc$ & 100-1000 & - & $\times$ & $\times$ \\
\hline
\rowcolor{green!10}
2025.05 & YuLan-OneSim & \checkmark & $\geq 10000$ & 50 & \checkmark & \checkmark \\
\hline\hline
\end{tabular}
}
\vspace{-0.5cm}
\end{table}

    \section{YuLan-OneSim}\label{sec3}
    
    \begin{figure}
    \centering
    \includegraphics[width=.95\linewidth]{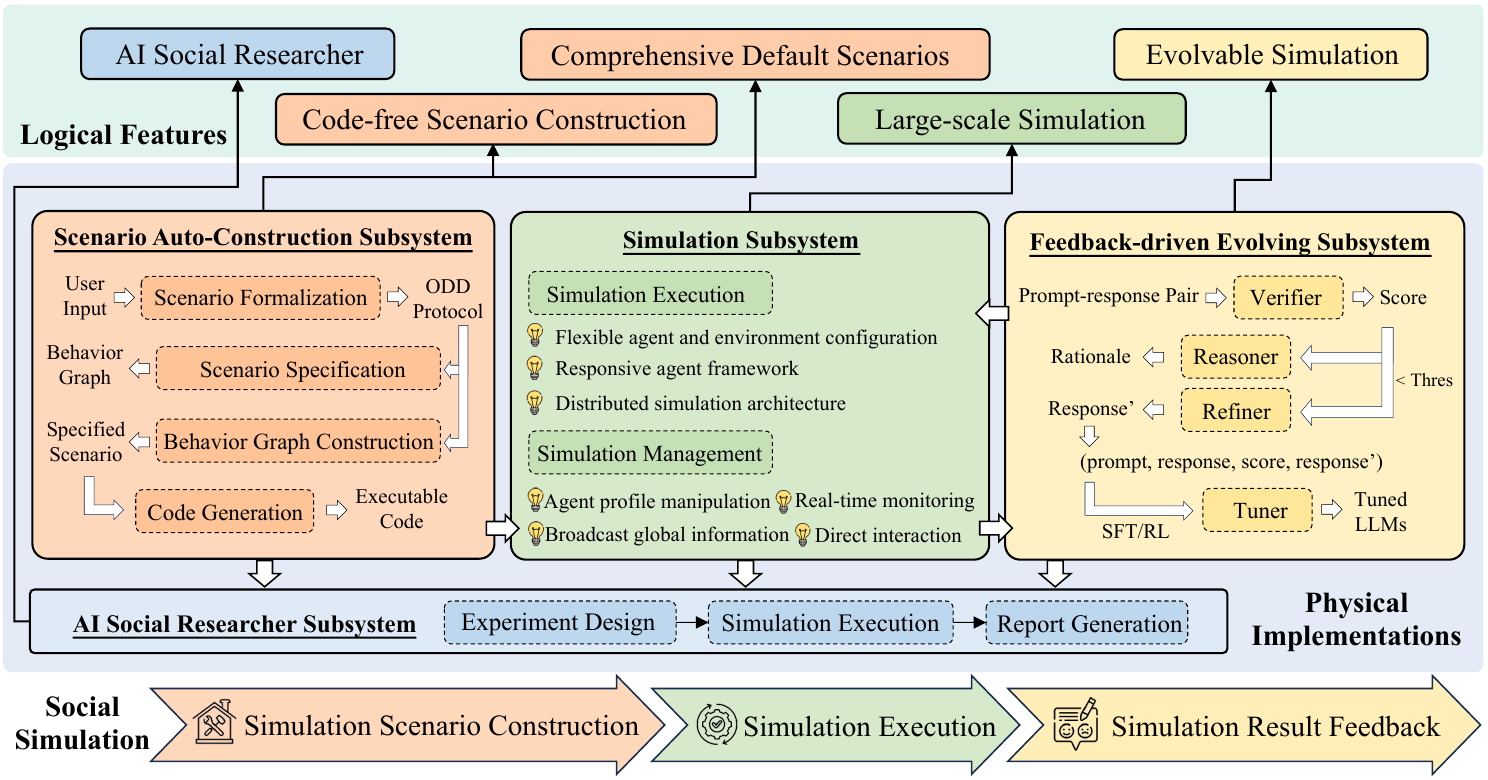}
    \caption{{
    The overall framework of our simulator.
    The upper subfigure highlights the key features of our simulator (i.e., logical characteristics), while the lower subfigure illustrates its core subsystems (i.e., physical implementations).
    The correspondences between the features and their supporting subsystems are labeled by solid arrows.
    }}
    \label{fig:overall}
    \vspace{-0.5cm}
    \end{figure}

    \subsection{Overview}
    The overall framework of our simulator is illustrated in Figure~\ref{fig:overall}.
    We divide the complete social simulation process into three phases:
    (1) Simulation scenario construction: In this phase, users are required to implement the entire simulation program, including defining agent profiles, specifying interaction behaviors, and other related components.
    (2) Simulation execution: During this phase, users run the simulation program, where the simulator should provide essential runtime support—such as large-scale simulation capabilities and other necessary system functionalities.
    (3) Simulation result feedback: In this phase, the simulator offers feedback to evaluate the performance of the simulation based on predefined metrics.
    Phases (2) and (3) can form an iterative ``execution–feedback–execution'' loop, enabling continuous refinement and enhancement of simulation quality.
    These three phases correspond to three subsystems in our simulator, respectively.
    In addition, we extend the usage paradigm of our simulator by introducing an AI social researcher subsystem. At this stage, phases (1)–(3) are integrated into a single cohesive module (denoted as X). The user provides a research question, which the AI social researcher translates into inputs compatible with module X. The corresponding outputs are then analyzed and synthesized into a final report.
    In the following sections, we detail the above-mentioned subsystems.
    
    \subsection{Scenario Auto-Construction Subsystem}
    Scenario construction serves as the foundation for social simulation. In previous studies, building scenarios typically required manually writing large amounts of code, which is labor-intensive and demands programming expertise.
    However, many social science researchers may lack programming experience, creating a gap between social simulators and their intended users.
    To bridge this gap, we introduce a scenario auto-construction subsystem that allows users to describe and refine their desired scenarios using natural language, while the system automatically generates the corresponding executable code.

    Although promising, developing automatic scenario construction methods remains inherently challenging, as agent types and interaction patterns are often highly complex and can vary significantly across different social scenarios.
    To address this challenge, we leverage the strong generalization capabilities of LLMs in programming and propose a four-step framework—comprising \textit{scenario formalization}, \textit{behavior graph construction}, \textit{code generation}, and \textit{agent specification}—to ensure the accuracy of the generated code. 
    More specifically, scenario formalization aims to translate users' natural language requirements into structured and comprehensive scenario descriptions, enabling LLMs to better understand and process user intent. 
    Building on the formalized scenario, behavior graph generation outlines the action logic of all agents by first extracting agent types and then generating their interaction patterns.
    Based on this behavior graph, code generation produces the corresponding executable code for simulation.
    Finally, users can specify the attributes and quantities of agents through the agent specification step.
    
    In the above framework, instead of having the LLM directly generate executable code, we first construct a behavior graph, which serves as an intermediate result to guide and simplify the code generation process. Importantly, all the above four steps are human-editable, allowing users to correct potential errors introduced by the LLM.
    In the following, we detail these five steps.

    \subsubsection{Scenario formalization}
    We adopt the Overview, Design Concepts, and Details (ODD) protocol to formalize user requirements. Specifically, the ODD protocol is a standardized framework for describing agent-based models~\cite{grimm2010odd}. The Overview defines the modeling purpose, key entities, and state variables. The Design Concepts outline the underlying modeling principles, including emergence, adaptation, and learning. The Details specify the initialization procedures, input data, and submodels. Together, these three components ensure sufficient modeling detail to support the execution and reproducibility of social simulations.

    To convert user requirements into ODD protocols, we develop an ODD-translation agent that interacts directly with users through conversation.
    The process begins with the user submitting an initial requirement.
    If the agent identifies any ambiguities or missing components essential for a comprehensive simulation, it asks the user for clarification.
    Once the agent determines that the provided information is sufficient, it concludes the dialogue and finalizes the corresponding ODD protocol. 

    A complete example of the ODD protocol as well as its generation process can be seen in Appendix~\ref{odd}. 
    % {ODD协议随对话逐渐完善的过程}

   \subsubsection{Behavior graph construction}
    The behavior graph defines how agents interact with each other and with the environment. Formally, let the behavior graph be $\mathcal{G} = \{\mathcal{N}, \mathcal{E}\}$, where $\mathcal{N}$ and $\mathcal{E}$ denote the sets of nodes and edges, respectively. Each node $n \in \mathcal{N}$ represents an agent action. 
    Given that agents of the same type share identical behavior logic, we abstract actions at the agent-type level. This means that each node $n$ represents an action associated with an agent type, rather than a specific agent.
    Each edge $e \in \mathcal{E}$ indicates that the action at the source node can trigger the action at the destination node.
    
    For example, in the job market scenario, there are two nodes called \textit{evaluate\_job\_applications} and \textit{screen\_candidates}, which represent the JobSeeker agent's action "evaluating job postings" and the Employer agent's action "evaluating job postings", respectively.
    These two nodes are connected by an edge called \textit{JobPostingEvaluationEvent}, which triggers the Employer agent to receive the JobSeeker agent's application and proceed to screen it (node \textit{screen\_candidates}), after the JobSeeker agent evaluates and applies to the job posting (node \textit{evaluate\_job\_applications}).

    To manage the initialization and termination of the simulation process, we introduce a start node and an end node. Specifically, the simulation begins when the start node emits a StartEvent, which activates the initial agent behaviors. As the simulation progresses, events propagate through the behavior graph, triggering agent actions and generating subsequent events, until the process reaches the end node.

    To automatically generate the behavior graph based on the formalized ODD protocol described in the previous section, we first use LLMs to extract all relevant agent types along with their corresponding profiles. Subsequently, the following steps are performed to construct the behavior graph:
    
    $\bullet$ \textbf{Action extraction}: We use LLMs to extract agent actions from the ODD protocol, where we record the target node name and necessary preconditions for execution. 
    
    $\bullet$ \textbf{Event extraction}: For each action, we identify the corresponding output events, including the name of the next action it points to and a description of the event itself. Each action may generate multiple different events based on different conditions.

    With the above two steps, we can generate an initial behavior graph. To ensure that the generated graph is reasonable, we further conduct the following two validation steps.
    
    $\bullet$ \textbf{Structure validation}: We examine the structure of the behavior graph, verifying that each node lies on a path from the Start node to the End node. This ensures that every action can be properly triggered and executed during simulation.
    
    $\bullet$ \textbf{Semantic validation}: We verify the alignments between the extracted behavior graph and the ODD protocol description, checking for errors in action and event descriptions.

    If either of the above validation steps fails, the identified issues and modification suggestions from the LLM are recorded. These are then used to regenerate the behavior graph, and the process iterates until the graph successfully passes both structural and semantic validation.

    While the above behavior graph establishes the interaction logic between agents, running a successful simulation requires specific data structures and variables. As a result, we improve the details of actions and events using LLMs through the following steps:
    
    $\bullet$ \textbf{Event variable specification}: We generate detailed information for all variables carried by each event, including their data types, default values, and descriptive metadata.
    
    $\bullet$ \textbf{Action input/output definition}: Based on the event specifications, we refine each action's execution requirements by defining:
    (1) Input variables required for the action to execute
    (2) Output variables produced by the action
    (3) Data types for all variables
    (4) Variable sources (whether from the environment, agent state, or incoming events)

    These steps enhance the behavior graph by defining how data is transformed and propagated throughout the graph. To ensure correctness, we employ LLMs to inspect the data-enhanced behavior graph, identifying any issues and providing suggestions for improvement. In practice, we iteratively refine the graph until it passes all validation checks, thereby ensuring data consistency and completeness.

    \subsubsection{Code generation}
    As described above, we have generated the behavior graph, which provides logical representations and data structures of the simulation scenario. However, converting this abstract graph into executable code remains a challenging task—even for state-of-the-art LLMs. To enhance code generation quality, our simulator incorporates three strategies, which are detailed below:

    $\bullet$ \textbf{General code structure.} Although simulation scenarios may differ significantly from one another, we can extract general functions that apply across various contexts. We have constructed base classes for agent and environment, containing abstracted foundational and common functionalities encapsulated as easy-to-use functions. Thanks to this infrastructure, when generating code, LLMs only need to create a corresponding handler function for each action to handle incoming events, focusing exclusively on the logic and content of the current action being generated—specifically, how to generate appropriate events in response, which agents should receive these events, and the prompts for the current action. For handler functions, we provide generic template code that LLMs can modify by replacing specific portions based on the details of the current action. For Events, benefiting from the detailed information defined in the data structure—such as variable names and data types—the event components can be directly assembled without requiring LLM involvement. 
    This modular code assembly approach significantly reduces the complexity of the generation process and improves overall efficiency.
    
    $\bullet$ \textbf{Graph-guided code generation.} Tasking an LLM with directly generating extensive simulation code would be extraordinarily difficult. Therefore, we employ breadth-first search (BFS) to traverse the simulation scenario's behavior graph. For each node visited, we first generate the code of events produced by the action corresponding to the node. We then provide the LLM with the scenario description, the agent type associated with the current action, the action's description, and code about both incoming and outgoing events. This enables the LLM to generate handler code for the current action based on the handler template. We then expand to successor nodes, progressively generating code for each node until all the code is produced.
    
    $\bullet$ \textbf{Iterative code validation and refinement.} The generated complete code undergoes comprehensive analysis by the LLM to identify potential errors. For any problematic agent or event code, issues and modification suggestions are recorded. Using these issues and suggestions, the LLM repairs problematic code segments and regenerates them. The revised code undergoes another validation cycle, and this iterative process continues until the code successfully passes validation checks.

    In addition to generating the code required to run the simulation, our system also automatically programs the metrics used for real-time monitoring.
    Specifically, we prompt LLMs to generate metric definitions based on the corresponding ODD protocols. These definitions specify the necessary variables and the computational logic required to calculate each metric.
    Using the generated definitions, we then produce metric-specific code, which is executed to monitor the simulation in real time.

    \begin{figure}[t]
    \centering
    \includegraphics[width=1.\linewidth]{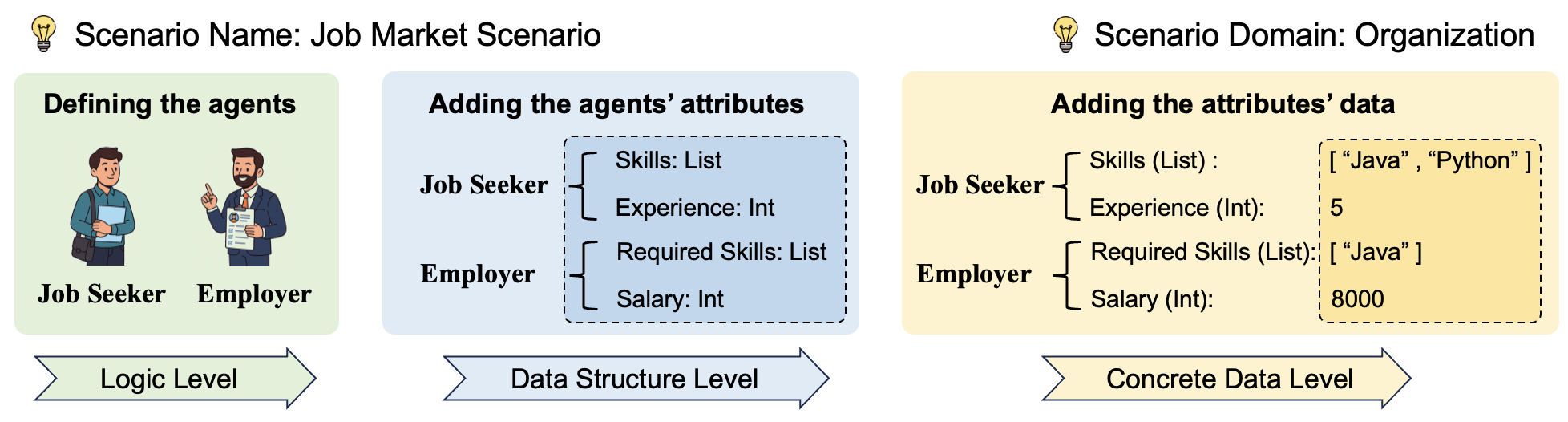}
    \caption{A toy example of the progressive scenario construction in the job market simulation: from the logical level to the data structure level, and finally to the concrete data level.}
    \label{fig:structure}
    \vspace{-0.cm}
    \end{figure}

    \begin{table}[t]
    \centering
    \caption{The default scenarios implemented in our simulator}
    \renewcommand\arraystretch{1.2}
    \label{default-scen}
    \scalebox{.9}{
    \begin{tabular}{m{2.2cm}<{\centering}|
    m{1.8cm}<{\centering}|
    m{10.0cm}<{\raggedright}}
    \hline\hline
    \textbf{Domain} & \textbf{\# Scenarios}& \textbf{Scenario Names} \\
    \hline
    Economics &  6 & Auction market dynamics, Bank reserves, Cash flow, Collective action problem, Customer satisfaction and loyalty model, Rational choice theory \\
    Sociology &  6 & Cultural capital theory, Norm formation theory, Social capital theory, Social relations theory, Social stratification network, Theory of planned behavior \\
    Political &  6 & Electoral polarization system, Public opinion polling, Rebellion, Selective exposure theory, Simple policy implementation model, Voting \\
    Psychology &  6 & Antisocial personality theory, Attribution theory, Cognitive dissonance theory, Conformity behavior model, Emotional contagion model, Metacognition theory \\
    Organization &  6 & Decision theory, Hawthorne studies, Hierarchy of needs, Labor market matching process, Organizational change and adaptation theory, Scientific management theory \\
    Demographics &  7 & Community health mobilization theory, Epidemic transmission network, Health belief model, Health inequality, Life course theory, Reciprocal altruism theory, SIR model \\
    Law &  7 & Case law model, Court trial simulation, Self defense and excessive defense, Social contract theory, Tort law and compensation, Unjust enrichment, Work hours and overtime in labor law \\
    Communication &  6 & Agenda setting theory, Cultural globalization, Diffusion of innovations, Information cascade and silence, Two step flow model, Uses and gratifications theory \\
    \hline\hline
    \end{tabular}
    }
    \vspace{-0.5cm}
    \end{table}

    \subsubsection{Scenario specification}
    In our simulator, the code and data are fully decoupled. 
    In the above sections, we have detailed the code generation process. Here, we move to the specification of the simulation data. 
    In general, we specify three types of data: 
    environment data, agent profile data, and agent relation data.
    For the environment data, we initialize the necessary variables based on the scenario description. If the scenario does not contain environment data, we skip this step.
    For the agent profile data, we instruct LLMs to generate an agent profile schema, which defines the agent attributes along with their data types and sampling methods. We support two sampling methods: LLM-based generation, where the LLM creates values for specific attributes, and random sampling within predefined ranges.
    For the agent relation data, we first have LLMs generate a relationship schema between different agent types, specifying whether relationships are unidirectional or bidirectional, and defining the nature of interpersonal relationships. Based on this schema, we randomly generate a relationship network. Then, we validate the generated relationships to ensure that no agent remains isolated and the events can flow normally through the behavior graph according to the defined topology. This validation guarantees that the social network structure supports the intended simulation dynamics while maintaining realistic relationship patterns.
    
    The above four steps progressively construct the simulation scenario—from the logical level to the data structure level, and ultimately to the concrete data level ({see Figure~\ref{fig:structure} for an example}). This staged process ensures smoother and more stable generation, thereby enhancing the overall quality of scenario construction.
    
    \subsubsection{Default scenario repository}
    To further support social science researchers, we have developed a default scenario repository comprising 50 scenarios across eight domains: economics, sociology, politics, psychology, management, public health, law, and communication (see Table~\ref{default-scen}). Each scenario includes a detailed description, an agent behavior graph, executable simulation code, and agent profiles. All scenarios are ready to run, enabling researchers to observe simulation results without any additional development. For more details, please refer to Appendix~\ref{app:detail}.

    \subsection{Simulation Subsystem}
    The simulation subsystem is responsible for executing and managing the simulation process.
    From the execution perspective, it supports flexible configurations of the agents and environments. The dynamic interactions between agents and their environments are realized based on a responsive framework as well as a distributed simulation architecture, which readily accommodates large-scale agent simulations.
    From the management perspective, users can flexibly manipulate agent profiles, broadcast global information, and interact with agents through conversation. Additionally, we implement a monitoring module to observe simulation performance in real time.

    \subsubsection{Simulation execution}
    
    $\bullet$ \textbf{Flexible agent and environment configuration}. 
    The agent in our simulator is composed of four modules: profile, memory, planning, and action.
    The profile module defines an agent's identity and characteristics, enabling it to assume different roles across various simulation scenarios. The profile module is highly customizable and can accommodate diverse profile data types. It is structured into two categories of attributes: public and private. Public attributes—such as name, gender, and occupation—are directly accessible to other agents during interactions. Private attributes, including personality traits and preferences, represent the agent's internal characteristics that are not directly visible to other agents but intrinsically influence behavior. This public-private profile design better reflects real-world interactions, enhancing simulation realism. Additionally, profile attributes can dynamically evolve as the simulation progresses.
    The memory module is designed to store and process the historical interaction records between agents and the environment.
    Our simulator supports extensive memory customization through the decoupled implementation of three submodules: memory strategy, storage, and operations. The strategy defines the memory organization method, such as historical sliding windows or combined long- and short-term memory mechanisms. Storage defines the memory container, such as vector databases or knowledge graphs. Operations define various memory manipulations, including addition, retrieval, reflection, merging, and forgetting. Furthermore, the memory module allows users to define different metrics for scoring and ranking memories—such as importance, recency, and relevance—for use during retrieval and reflection processes. Through flexible customization and combination of these modules, users can implement diverse agent memory mechanisms. 
    The planning module determines the agents' behavioral patterns. 
    Our simulator incorporates three planning approaches: Chain-of-Thought (COT), Belief-Desire-Intention (BDI), and Theory-of-Mind (TOM). COT emphasizes single-step reasoning, BDI excels at long-term goal planning, and TOM is designed for interpersonal interaction scenarios, taking greater account of interaction partners' perspectives.
    The action module converts the agent profile, memory, planning, and current contextual information into a specific agent behavior. 

    The environment functions as the central hub of the simulation, controlling the entire simulation lifecycle. As previously mentioned, the environment has two special actions: Start and End. Start sends initialization events to agents to begin the simulation. The environment controls the simulation progress through two modes: round and tick. In the round mode, the environment uniformly sends StartEvents at the beginning of each round and waits for all agents to send completion events before ending the current round and beginning the next. This mode is suitable for turn-based simulations or scenarios requiring strict process control. In the tick mode, the environment does not wait for all agents to complete their actions but instead sends StartEvents at regular intervals, allowing agents to act continuously. This ensures constant event flow in the environment and is suitable for periodically occurring simulation scenarios. 
    The environment also maintains public variables and information in the simulation scenario, such as real-time bidding prices in auction scenarios or government policies in virus transmission scenarios. This information changes accordingly based on agent actions. The environment periodically collects data on agent actions, transmitted events, and other information from the scenario and stores it for analysis.

    $\bullet$ \textbf{Responsive agent framework}.
    Real-world social dynamics are rarely perfectly synchronized; individuals and groups respond to events and information asynchronously. To capture this essential characteristic, we designed our simulator based on a fully responsive framework.
    Specifically, we adopt an event-driven, reactive paradigm centered on an asynchronous event bus. This design reflects the fundamentally reactive nature of social interactions. Rather than enforcing a rigid, step-by-step execution flow, our simulator operates through the propagation and handling of discrete Event objects. Each event encapsulates a meaningful occurrence within the simulation—such as an agent action, an environmental change, or a communication attempt—and carries contextual information including the originator, intended recipient(s), event type, and relevant payload.
    Agents and the environment function as reactive components that subscribe to specific event types relevant to their roles. When an event is dispatched by the central event bus, all subscribed components are notified and asynchronously trigger their corresponding internal logic. This decoupling of event producers and consumers mirrors the autonomy of actors in real-world social systems, enabling a more natural representation of concurrent activities and complex causal relationships.
    Moreover, this asynchronous, message-passing architecture inherently supports parallelism and modularity, simplifying the design of sophisticated agent behaviors and facilitating future extensibility.

    $\bullet$ \textbf{Distributed simulation architecture}.
    Supporting large-scale simulation is a fundamental requirement for achieving more realistic and stable simulation results. To this end, we base our simulator on a robust Master-Worker distributed architecture. This design distributes the computational load while maintaining centralized coordination for global consistency. Specifically, our distributed simulation framework has the following features:
    
    (1) \textbf{Centralized coordination (Master node):} We deploy a master node to serve as the orchestrator. It manages the lifecycle of the worker nodes, maintains a dynamic registry of agent locations across the distributed system, holds the canonical state of the global simulation environment, and arbitrates access to shared resources, ensuring coherent system-wide evolution.
    
    (2) \textbf{Distributed computation (Worker nodes):} The bulk of the simulation workload – executing individual agent logic and processing local interactions – is delegated to multiple worker nodes. Each worker hosts a subset of the agent population, allowing for parallel execution of agent behaviors.
    
    (3) \textbf{High-performance inter-node communication:} Communication between the master and workers, and potentially between different workers, relies on the efficient gRPC framework~\cite{wang1993grpc}. This ensures low-latency, high-throughput exchange of events, state updates, and control signals, which is critical for maintaining performance in a distributed setting. Optimizations such as batching non-critical data transmissions are employed to further reduce communication overhead.
    
    (4) \textbf{Topology-aware agent allocation:} Recognizing that communication latency is a key bottleneck in distributed systems, we employ a sophisticated agent allocation strategy before the simulation begins. The allocator analyzes the specified agent profiles and their anticipated interaction patterns (e.g., based on predefined relationships or agent types). It intelligently assigns agents to different worker nodes, aiming to co-locate agents that are likely to interact frequently, thereby minimizing costly cross-node event transmissions and optimizing overall simulation throughput. 
    
    (5) \textbf{Seamless event routing:} The event bus transparently handles event dispatch across the distributed topology. Events targeted at agents residing on the same worker are processed locally. For events destined for remote agents, our system employs a hybrid routing approach: initially, these events are routed through the master node, which forwards them to the appropriate destination based on its agent location registry. However, to optimize communication efficiency, the system implements a peer-to-peer (P2P) communication mechanism between worker nodes. After the first interaction between agents on different workers (mediated by the master), the system caches the connection information, enabling subsequent communications to occur directly between the relevant worker nodes without master intervention. This adaptive routing strategy significantly reduces communication overhead and potential bottlenecks at the master node, particularly for frequently interacting agent clusters distributed across different workers.
    
    (6) \textbf{Seamless environment access:} 
    For simplicity, we let the agents interact with the environment through a proxy interface. This proxy presents a consistent API regardless of whether the simulation is running in distributed or single-machine mode. The proxy environment encapsulates all necessary RPC calls to the master node behind the scenes, ensuring all agents operate on a consistent view of the environment state. This design allows developers to write agent code once and deploy it unchanged in both local development and large-scale distributed scenarios.
   
    By integrating the above responsive agent framework with the distributed simulation architecture, our simulator provides a powerful foundation for exploring large-scale social phenomena. This design allows researchers to model complex systems with numerous entities, pushing the boundaries of computational social science towards greater realism and scale.

    \subsubsection{Simulation management}

    $\bullet$ \textbf{Agent profile manipulation.} 
    One can freely configure agent profiles before simulation, and also can modify profiles in real-time after the simulation has begun. This capability enables researchers to study how attribute changes influence agent behaviors throughout the simulation lifecycle.

    $\bullet$ \textbf{Broadcast global information.}
    During simulations, the users can broadcast messages to all agents at any time, allowing observation of how global event changes impact the simulation system. This facilitates the study of large-scale interventions such as policy changes or crisis scenarios.

    $\bullet$ \textbf{Direct interaction.} 
    The system allows flexible interactions with any agent during the simulation process. For example, one can interview agents to understand their decision-making processes or guide agents to adopt specific behaviors, enabling controlled experiments with our simulator.

    $\bullet$ \textbf{Real-time monitoring.}
    For a given simulation scenario, the environment can automatically collect data from relevant agents and compute scenario-specific metrics, which are then presented in graphical formats within our simulator.
    In standalone mode, the environment directly collects the relevant monitoring information. In distributed deployments, the environment queries agent data by sending data collection events to agents. These strategies ensure consistent monitoring capabilities regardless of deployment configuration, providing researchers with immediate visibility into simulation dynamics and emerging patterns.
    
    \subsection{Feedback-driven Evolving Subsystem}
    Due to the inherent limitations of LLMs (e.g., hallucination and bias), the simulation result at each step may be unreliable. Even more concerning, errors introduced in earlier stages of the simulation can accumulate and amplify as the process progresses. 
    To alleviate this problem, we incorporate an error correction mechanism based on system or human feedback. 
    Specifically, we design a multi-agent framework comprising a verifier, a reasoner, a refiner, and a tuner, which collaboratively label simulation samples and re-train the backbone LLMs based on the annotated results (see Figure~\ref{fig:overall}).
    To begin with, the system generates a prompt–response pair, denoted as $(\text{p}, \text{r})$.
    The verifier evaluates the correctness of $\text{r}$ with respect to $\text{p}$ using predefined metrics and assigns a score $\text{s}$.
    Next, the reasoner analyzes the rationale behind the score $\text{s}$ and produces an explanation, denoted as $\text{res}$. 
    If the score $\text{s}$ falls below a predefined threshold $\text{thres}$, the refiner generates a corrected response $\text{r'}$ based on the combination of $\text{p}$, $\text{r}$, $\text{s}$, and $\text{res}$.
    Following this process, for all underperforming prompt–response pairs, we collect the quadruplets $(\text{p}, \text{r}, \text{s}, \text{r'})$.
    At last, the tuner fine-tunes the backbone LLMs using supervised fine-tuning (SFT) or reinforcement learning (RL). For the former method, we only use the data $(\text{p}, \text{r'})$, while for the latter one, we leverage the complete data $(\text{p}, \text{r}, \text{s}, \text{r'})$.
    
    In the above error correction mechanism, the verifier, reasoner, and refiner can each be implemented using either LLMs or real humans. As feedback accumulates from an increasing number of users, the simulator is expected to evolve toward more reliable and human-aligned simulations.
    In fact, we believe that in many domains of social science, real-world data is often difficult to access or collect due to factors such as privacy concerns, ethical constraints, and limited availability—challenges that differ significantly from those in typical AI domains. As a result, directly training LLMs on real-world social data can be impractical. To address this challenge, our simulator provides a "social science gym," enabling users to infuse their social intelligence into the underlying LLMs in a controlled and customizable manner.

    \subsection{AI Social Researcher Subsystem}
\begin{figure}[t]
    \centering
    \includegraphics[width=.9\linewidth]{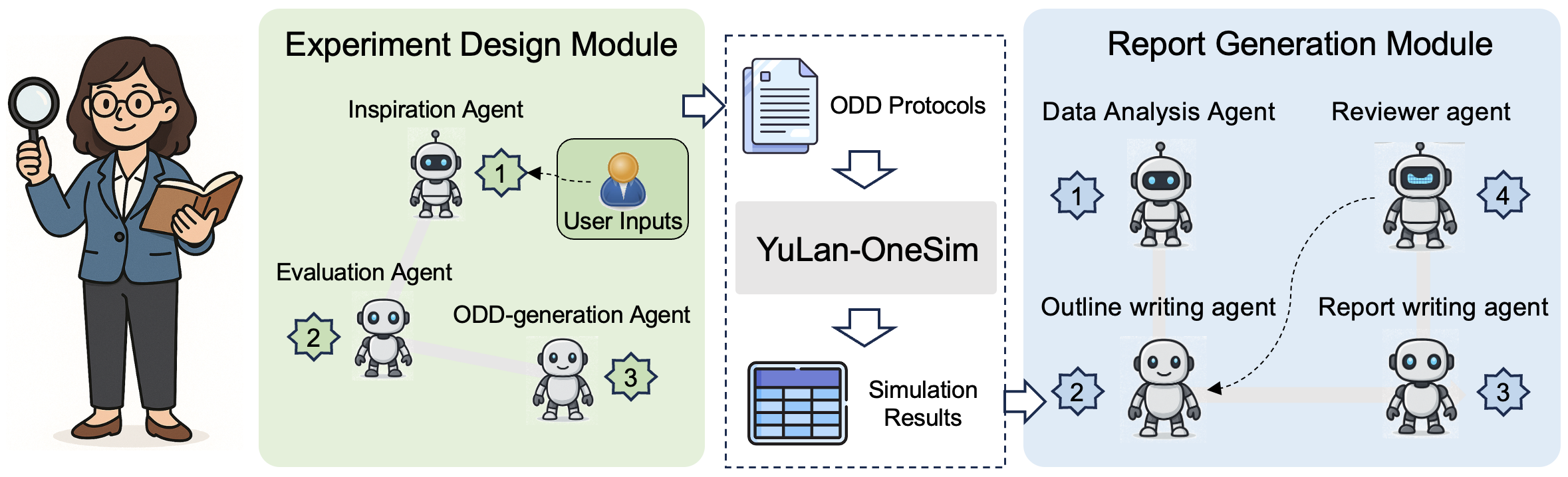}
    \caption{Illustration of the working process of the AI Social Researcher. There are two main modules: the Experiment design module and the Report generation module. YuLan-OneSim serves as the central tool for executing simulations and enabling a complete loop of social science research.
    }
    \label{fig:AIS}
    \vspace{-0.5cm}
    \end{figure}

    In the above sections, we systematically introduce the complete social simulation process based on our simulator.
    In this section, we treat the simulator as a black-box tool and develop an AI social researcher to further extend its usage.
    In general, the user simply needs to input a research topic (e.g., {\emph{The spread of rumors on social media platforms}}), and the AI social researcher analyzes the topic to generate a corresponding set of simulation-ready ODD protocols. These protocols are then used to initiate the simulator. Finally, the outputs produced by the simulation are synthesized into detailed technical reports.
    To achieve the above goal, we implement the AI social researcher based on two modules: the \textit{Experiment design module} and the \textit{Report generation module} (see Figure~\ref{fig:AIS}), each consisting of specialized collaborative agents.
    In the following, we detail these modules.

    % {三个阶段，从研究问题到ODD到模拟结果}

    \subsubsection{Experiment design module}
    This module aims to transform the user-provided research topic into a fully specified simulation scenario. This transformation is accomplished based on three agents: 
    (1) \textbf{Inspiration Agent}: This agent serves as the creative ideation component.  Given a brief and potentially ambiguous social science topic, it generates 3--5 candidate research questions and corresponding simulation scenarios. Each question targets a distinct aspect of the topic and is designed to be feasible for agent-based modeling. The associated scenarios include detailed characterizations of agent types, interaction mechanisms, and manipulable parameters, all constrained to manageable complexity.
    (2) \textbf{Evaluation Agent}: This agent acts as a critical assessor. It evaluates each proposed scenario using six criteria: simulation feasibility, complexity management, research value, agent design quality, parameter space definition, and potential insights. Each scenario is scored (1--10) per criterion with accompanying justifications. The agent then selects the most promising scenario for further elaboration, providing a rationale for its choice.
    (3) \textbf{ODD-generation Agent}: This agent converts the selected scenario into a standard simulation specification using the ODD protocol. 
    It assigns the scenario to one of eight predefined domains (\textit{e.g.}, Sociology, Politics, Economics), generates a scenario name, and constructs a comprehensive ODD description. 
    All information is formatted as strings using JSON-compatible structures to ensure both compatibility and simplicity. Based on the final output, our simulator can be directly initialized to run the simulation.

    \subsubsection{Report generation module}
    The ODD protocol generated by the above module is used to invoke our simulator, and based on the simulation results, a structured and insightful research report is synthesized through the report generation module.
    Specifically, the above goal is achieved by the following agents:
    (1) \textbf{Data analysis agent}: This agent interprets simulation results using visualizations, metrics, metadata, and ODD scenario descriptions. It extracts significant patterns, correlations, and emergent behaviors, and contextualizes them relative to the simulation objectives, producing a set of actionable insights.
    (2) \textbf{Outline writing agent}: Based on the scenario and analytical insights, this agent generates a detailed, hierarchical outline for the research report. 
    The outline includes sections for research objectives, simulation setup, experimental results, and conclusions, with bullet-point guidance on key content in each section.
    (3) \textbf{Report writing agent}: This agent composes the full report in LaTeX, following the above-generated outline. It sequentially generates each section according to the \textit{Research objectives}, which clearly articulate the research aims, contextualize their relevance, and establish their connection to the simulation setup, the \textit{Simulation setup}, which provides a precise technical description of the model, including agents, environments, behaviors, and evaluation metrics, the \textit{Experimental results}, which presents and interprets findings, emphasizing data-driven insights with reference to figures and quantitative outcomes, and the \textit{Conclusion}, which synthesizes the findings, discusses theoretical and practical implications, acknowledges limitations, and suggests avenues for future work.
    (4) \textbf{Reviewer agent}: This agent critically reviews the generated report across four dimensions: analytical insight, structural coherence, content quality, and practical utility. Each section is evaluated individually, and scores (1--5) are assigned per criterion. The reviewer agent provides constructive feedback and concrete suggestions for improvement.

    After the initial draft is generated, the report is reviewed by the Reviewer agent, which provides detailed, section-specific feedback. The system then revises the report accordingly in multiple iterations, incorporating the feedback until the content meets predefined standards of quality in insight, structure, content, and utility. The final result is a complete research report ready for academic presentation.

    In summary, the AI Social Researcher subsystem implements a structured, modular workflow for automated social simulation research. It integrates multiple LLM-based agents to handle scenario generation, evaluation, formalization, data analysis, and report writing. The system enables researchers to move efficiently from vague ideas to detailed simulations and comprehensive reports with minimal manual effort.

    \section{Experiments}\label{sec4}
    To evaluate the effectiveness and efficiency of our simulator, we conduct extensive experiments, focusing on the following four research questions (\textbf{RQ}):
    
    \textbf{RQ 1}: How high is the quality of our scenario auto-construction subsystem?

    \textbf{RQ 2}: Are the simulation results of YuLan-OneSim reliable?

    \textbf{RQ 3}: Is the simulation process of YuLan-OneSim efficient and scalable?

    \textbf{RQ 4}: How effective is our AI social researcher?

    In the following sections, we address each of these questions in detail.

    \subsection{Evaluation on the Scenario Auto-Construction Subsystem}

    In this section, we conduct experiments using all the default scenarios in our simulator (see Appendix~\ref{app:detail} for more details), and the following metrics are leveraged to evaluate the quality of the scenario construction process:
    
    $\bullet$ \textbf{Behavior Graph Rating (BG-Rating)}: 
    This metric evaluates the quality of the intermediate behavior graphs that serve as the foundation for generating the final simulation code. 
    We score the graph in the range of [1,5] based on the criteria in {Table~\ref{cre:bg}}.

    $\bullet$ \textbf{Code Rating (C-Rating)}: 
    This metric assesses the quality of the final generated simulation code.
    We score the code in the range of [1,5] based on the criteria in {Table~\ref{cre:simu}}.

    $\bullet$ \textbf{Generation Time (G-Time)}: This metric evaluates the efficiency of code generation—that is, the time required to generate the simulation code.

    $\bullet$ \textbf{Generation Tokens(G-Tokens), Files \& Lines}: 
    These metrics report the total number of tokens, files, and lines of generated code.

    \begin{table}[t]
    \centering
    \caption{Criteria for evaluating the behavior graph}
    \renewcommand\arraystretch{1.2}
    \label{cre:bg}
    \begin{tabular}{m{1.cm}<{\centering}|
    m{12.0cm}<{\raggedright}}
    \hline\hline
    \textbf{Score} & \textbf{Description} \\
    \hline
    5 &  Fully accurately identifies all core steps. Extracted steps are completely consistent with the original description. Step sequence matches exactly 80–100\% of key steps are correctly identified.\\\hline
    4 &  Accurately identifies the vast majority of core steps. Extracted steps are highly consistent with the original description. Step sequence is mostly correct 60–80\% of key steps are correctly identified. \\\hline
    3 &  Identifies most core steps. Extracted steps are generally consistent with the original description. Some inaccuracies in step sequence 40–60\% of key steps are correctly identified. \\\hline
    2 &  Identifies some core steps but with significant omissions. Extracted steps are partially inconsistent with the original description. Step sequence is mostly disordered 20–40\% of key steps are correctly identified. \\\hline
    1 & Fails to identify core steps of the workflow. Extracted steps are severely inconsistent with the original description. Step sequence is completely disordered. Fewer than 20\% of key steps are correctly identified. \\
    \hline\hline
    \end{tabular}
    \end{table}

    \begin{table}[t]
    \centering
    \caption{Criteria for evaluating the simulation code}
    \renewcommand\arraystretch{1.2}
    \label{cre:simu}
    \begin{tabular}{m{1.cm}<{\centering}|
    m{12.0cm}<{\raggedright}}
    \hline\hline
    \textbf{Score} & \textbf{Description} \\
    \hline
    5 &  The generated code fully matches the workflow. No syntax errors. Runs flawlessly. Accurately reflects the workflow logic. Code quality: 80–100\%. Clear code structure with good readability and maintainability. No manual modification required \\\hline
    4 &  The generated code highly conforms to the workflow. Virtually no syntax errors. Runs smoothly. Accurately reflects the workflow logic. Code quality: 60–80\%. Requires minor manual adjustments to fix a few edge cases or detail issues. \\\hline
    3 &  The generated code generally conforms to the workflow. Contains some syntax or logic errors. Partially executable. Roughly reflects the workflow logic. Code quality: 40–60\%. Requires manual revision of key algorithms and logic, fixing multiple code defects, and significant refactoring by developers \\\hline
    2 &  The generated code is partially related to the workflow. Contains severe syntax and logic errors. Difficult to execute. Reflects only a small portion of the workflow logic. Code quality: 20–40\%. Requires rewriting 60–90\% of the code. Major refactoring of core logic is needed. Multiple critical algorithm and business logic issues exist. \\\hline
    1 & The generated code is unrelated to the workflow. Severe syntax errors. Cannot run at all. Does not reflect any workflow logic. Code quality: below 20\%. Requires rewriting over 90\% of the code. Almost unusable, essentially starting from scratch. Contains serious syntax and logical flaws. Developers need to completely redesign the code architecture. \\
    \hline\hline
    \end{tabular}
    \end{table}

    For all the above metrics, we report the average results across different scenarios within each domain. We empirically adopt {GPT-4o} as the code generation LLM due to its strong performance.
    
\begin{figure}[t]
    \centering
    \includegraphics[width=.95\textwidth]{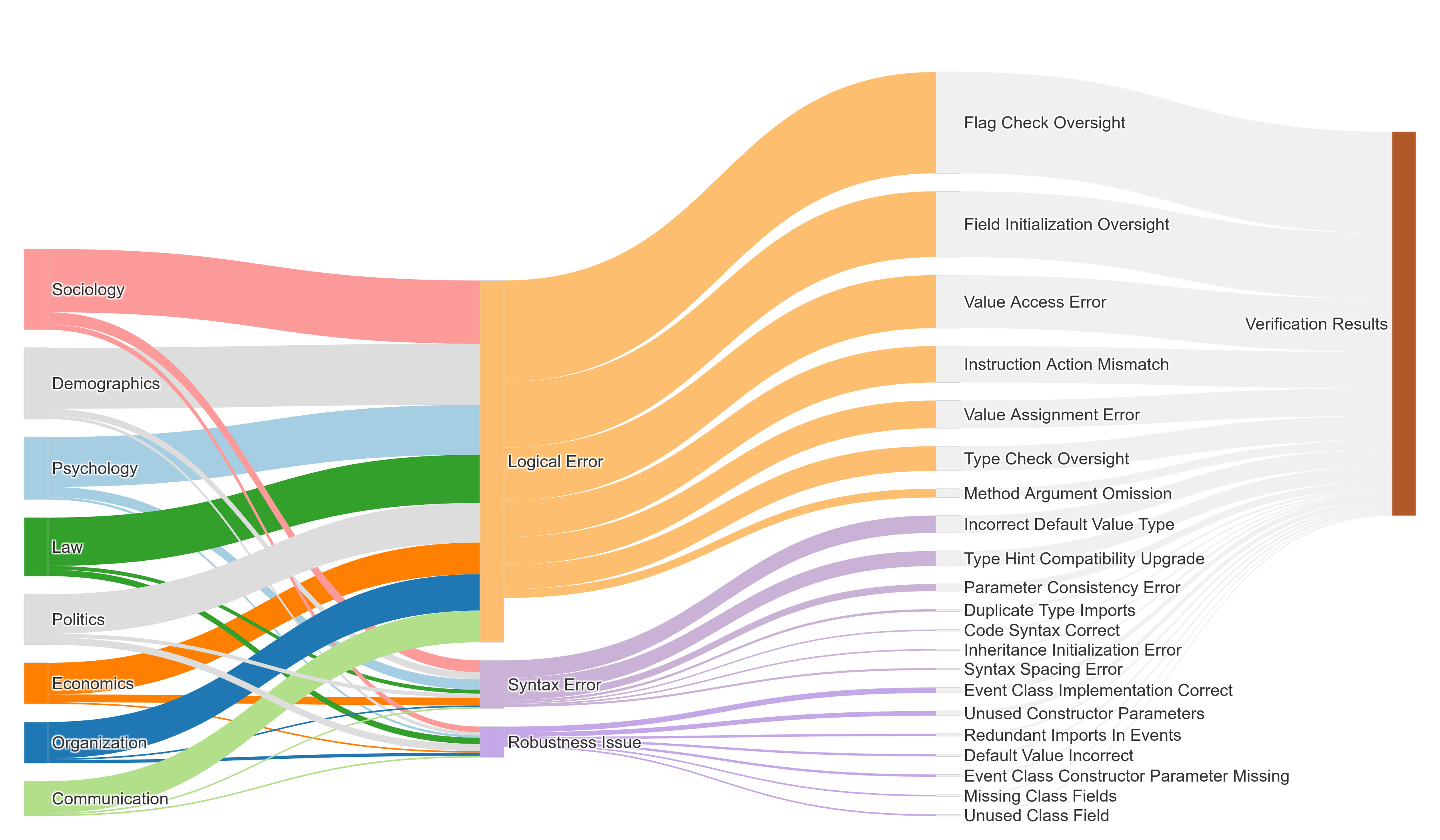}
    \caption{Illustration on the distribution of various error types—including logical errors, syntax errors, and robustness issues—across different domains.}
    \label{fig:domain_error_sankey}
    \vspace{-0.5cm}
    \end{figure}

    From the result shown in Table~\ref{exp:auto_coding_eval}, we can see:
    from the efficiency perspective, our simulator generates code at an average speed of 50 tokens per second, which is over 14 times faster than the average human programmer (approximately 3.5 tokens per second\footnote{https://saboorqadri.medium.com/what-is-the-average-programmers-typing-speed-c5f56cfd6287}).
    From the effectiveness perspective, the average ratings of the generated behavior graph and simulation code are 4.82 and 4.20, respectively, suggesting the strong potential of our simulator for automated simulation scenario construction.
    To better understand how our simulator makes mistakes, we further present the specific errors made by our simulator in Figure~\ref{fig:domain_error_sankey}.
    We observe that the majority of errors produced by our simulator are logical errors, including issues such as value access errors, instruction–action mismatches, incorrect value assignments, and type-check oversights. These errors are typically straightforward to correct using standard debugging techniques and require minimal manual intervention. However, their prevalence suggests that targeted improvements in logical validation could significantly enhance the overall reliability of the code generation process. Syntax errors and robustness issues are also present in the simulator’s output, but they represent a much smaller fraction of the total errors. While the current results are encouraging, our implementation still exhibits limitations when dealing with more complex logical dependencies and edge cases. To address these challenges, we plan to incorporate advanced error detection and repair mechanisms in future work, aiming to further reduce the need for manual code correction.

    \textbf{Ablation Study}.
    To evaluate the role of different components in our auto-construction subsystem, we perform ablation studies comparing the full subsystem with three variants: 
    \underline{\textit{w/o Graph}}, which skips behavior graph generation and relies solely on LLM coding; 
    \underline{\textit{w/o G-Valid}}, which omits validation of the behavior graph; and \underline{\textit{w/o C-Refinement}}, which excludes code evaluation and refinement.
    For each of the eight default domains, we randomly sample one scenario for evaluation. As shown in Table~\ref{tab:ablation}, removing the behavior graph leads to a substantial drop in code quality, underscoring its importance. Likewise, the weaker performance of w/o G-Valid highlights the necessity of graph validation. Finally, the results of w/o C-Refinement demonstrate that iterative code refinement further enhances performance.
    
\begin{table}[t]
\centering
\caption{Evaluation results on the scenario construction}
\renewcommand\arraystretch{1.2}
\label{exp:auto_coding_eval}
\scalebox{.90}{
\begin{tabular}{m{2.1cm}<{\raggedright}|
m{2.2cm}<{\centering}|
m{2.2cm}<{\centering}|
m{1.3cm}<{\centering}|
m{1.5cm}<{\centering}|
m{1.3cm}<{\centering}|
m{1.3cm}<{\centering}}\hline\hline
\textbf{Domain} & \textbf{BG-Rating} & \textbf{C-Rating} & \textbf{G-Time} & \textbf{G-Tokens} & \textbf{Files} & \textbf{Lines}\\\hline
Communication & 5.00$\pm$0.00 & 4.17$\pm$0.37 & 362.51 & 20978.0 & 13.33 & 588.33  \\
Demographics & 4.86$\pm$0.35 & 4.29$\pm$0.45 & 844.19 & 25342.8 & 14.25 & 667.75  \\
Economics & 5.00$\pm$0.00 & 4.00$\pm$0.37 & 294.80 & 20487.4 & 12.60 & 581.60  \\
Law & 4.57$\pm$0.49 & 4.29$\pm$0.45 & 402.41 & 20760.8 & 13.83 & 615.67  \\
Organization & 4.50$\pm$0.50 & 4.00$\pm$0.00 & 412.77 & 17595.8 & 13.40 & 674.00  \\
Politics & 5.00$\pm$0.00 & 4.33$\pm$0.47 & 177.05 & 12491.7 & 12.33 & 458.50  \\
Psychology & 4.83$\pm$0.37 & 4.17$\pm$0.37 & 310.83 & 16056.8 & 12.88 & 540.25  \\
Sociology & 4.83$\pm$0.37 & 4.33$\pm$0.47 & 261.15 & 16122.7 & 12.57 & 518.29  \\\hline
{Average} & 4.82$\pm$0.20 & 4.20$\pm$0.13 & 358.95 & 18080.7 & 13.71 & 570.66 \\\hline\hline
\end{tabular}
}
\vspace{-0.5cm}
\end{table}

\begin{table}[t]
\centering
\caption{{Ablation study on different components of our scenario auto-construction subsystem}}
\renewcommand\arraystretch{1.1}
\label{tab:ablation}
\scalebox{1}{
\begin{tabular}{l|c|c|c|c|c|c|c|c|c}
\hline \hline
\textbf{Method} & \textbf{Econ.} & \textbf{Soc.} & \textbf{Poli.} & \textbf{Psy.} & \textbf{Org.} & \textbf{Demo.} & \textbf{Law} & \textbf{Comm.} & \textbf{Avg.} \\
\hline
w/o Graph & 2 & 3 & 3 & 3 & 3 & 3 & 3 & 3 & 2.88 \\
w/o G-Valid & 2 & 3 & 3 & \textbf{4} & 3 & 3 & 3 & 3 & 3.00 \\
w/o C-Refinement & 3 & \textbf{4} & 3 & \textbf{4} & 3 & 3 & 3 & \textbf{4} & 3.38 \\
YuLan-OneSim & \textbf{4} & \textbf{4} & \textbf{4} & \textbf{4} & \textbf{4} & \textbf{4} & \textbf{4} & \textbf{4} & \textbf{4.00} \\
\hline \hline
\end{tabular}
}
\end{table}
    
    \subsection{Evaluation on the Simulation Reliability}
    In this section, we evaluate the reliability of our simulator from two perspectives: (1) Social theory validation — evaluating whether well-established social theories can be verified within our simulation environments; and (2) Real-world data alignment — examining the degree to which our simulated data align with real-world observations.

    \subsubsection{Social theory verification}\label{sec:emergence}
    Previous research in social science has produced a large number of theories, which typically serve as high-level abstractions of real-world phenomena. If these theories can also be verified within a simulated environment, it indicates that the simulation is reliable.

    In this experiment, we use Axelrod’s cultural dissemination scenario as a case to study and evaluate our simulator~\cite{axelrod1997dissemination}. Specifically, agents are arranged in an N×N grid, with each agent possessing F cultural features, each of which can take on one of q possible states. Agents interact with their adjacent neighbors, and the probability of interaction increases with cultural similarity. During these interactions, agents may adopt certain features from their neighbors, thereby modifying their own cultural profiles. This scenario is designed to explore how cultural traits diffuse and evolve within a population.
    Following Axelrod's setting~\cite{axelrod1997dissemination}, we set N=10, F=5, and q=5, but the agents' interactions are determined by LLMs based on their profiles.

    \begin{figure}[t]
        \centering
        \includegraphics[width=\textwidth]{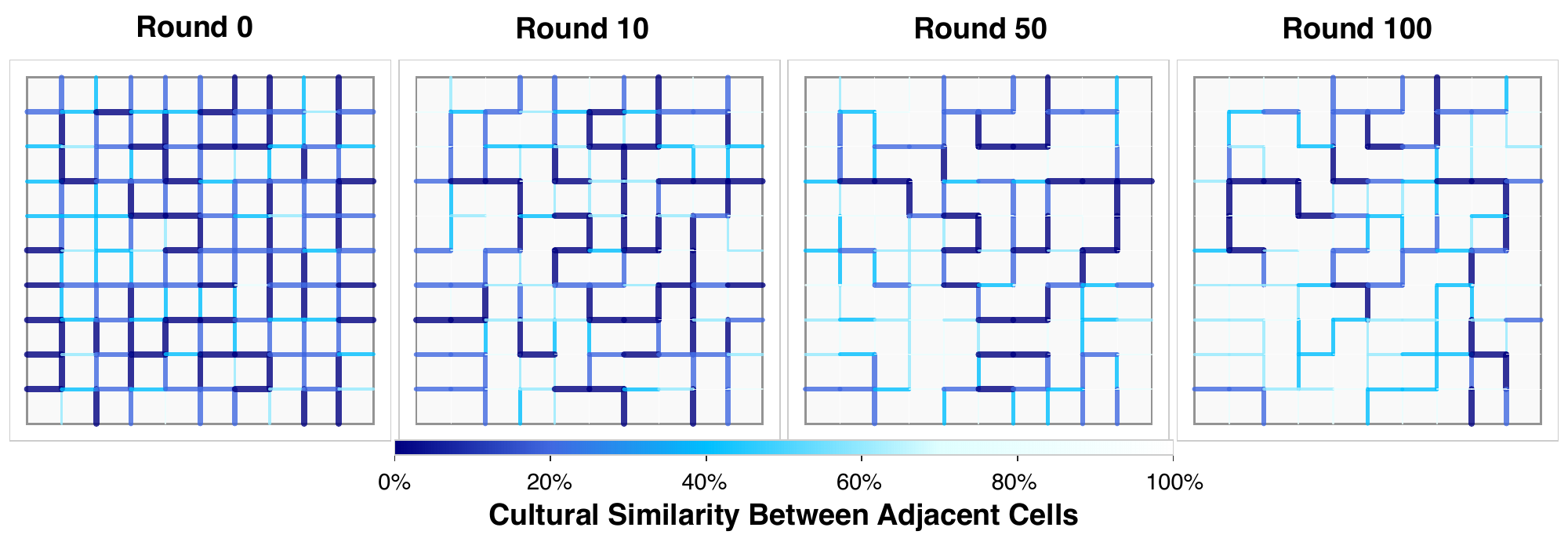}
        \caption{Cultural Similarity Maps across simulation stages. The four panels show the evolution of cultural similarity between adjacent agents at rounds 0, 25, 50, and 100. Darker connections indicate lower cultural similarity, revealing the formation of distinct cultural regions over time. The color gradient corresponds to similarity percentages from 0\% (darkest) to 100\% (lightest).}
        \label{fig:culture_map}
    \end{figure}
    
    The simulation results are presented in Figure \ref{fig:culture_map}. We can see: as the simulation progresses, distinct cultural boundaries gradually emerge.
    Within each cultural region, neighboring agents exhibit high similarity, indicated by darker connections, while the boundaries between regions remain clearly visible through lighter connections. This visual pattern effectively captures the core insight of Axelrod’s theory—local interactions promote regional cultural homogeneity, while the overall cultural diversity is preserved at a global level.

    Beyond the above qualitative analysis, we also conduct quantitative studies using the following two metrics to gain deeper insights into how Axelrod’s theoretical conclusions emerge during the simulation process:
    
    $\bullet$ \textbf{Local Convergence (LC):} This metric measures the average cultural similarity between all neighboring agent pairs. Mathematically, it is defined as:
    \begin{equation}
        LC = \frac{1}{|E|} \sum_{(i,j) \in E} \frac{|F_i \cap F_j|}{|F|}
        \end{equation}
    where $E$ is the set of all adjacent agent pairs, $F_i$ and $F_j$ represent the cultural feature sets of agents $i$ and $j$ respectively, and $|F|$ is the total number of features. Higher values indicate greater cultural homogeneity among neighboring agents, signifying the formation of local cultural regions.
        
    $\bullet$ \textbf{Global Polarization (GP):} This metric quantifies the number of distinct cultural regions in the system, normalized by the total number of agents. It is calculated as:
        \begin{equation}
        GP = \frac{|C|}{N^2}
        \end{equation}
    where $|C|$ is the number of distinct cultural regions (connected clusters of agents sharing identical cultural profiles) and $N^2$ is the total number of agents. Higher values indicate more diverse cultural regions within the system, reflecting the degree of cultural diversity at the global level.

    \begin{figure}[t]
        \centering
        \includegraphics[width=\textwidth]{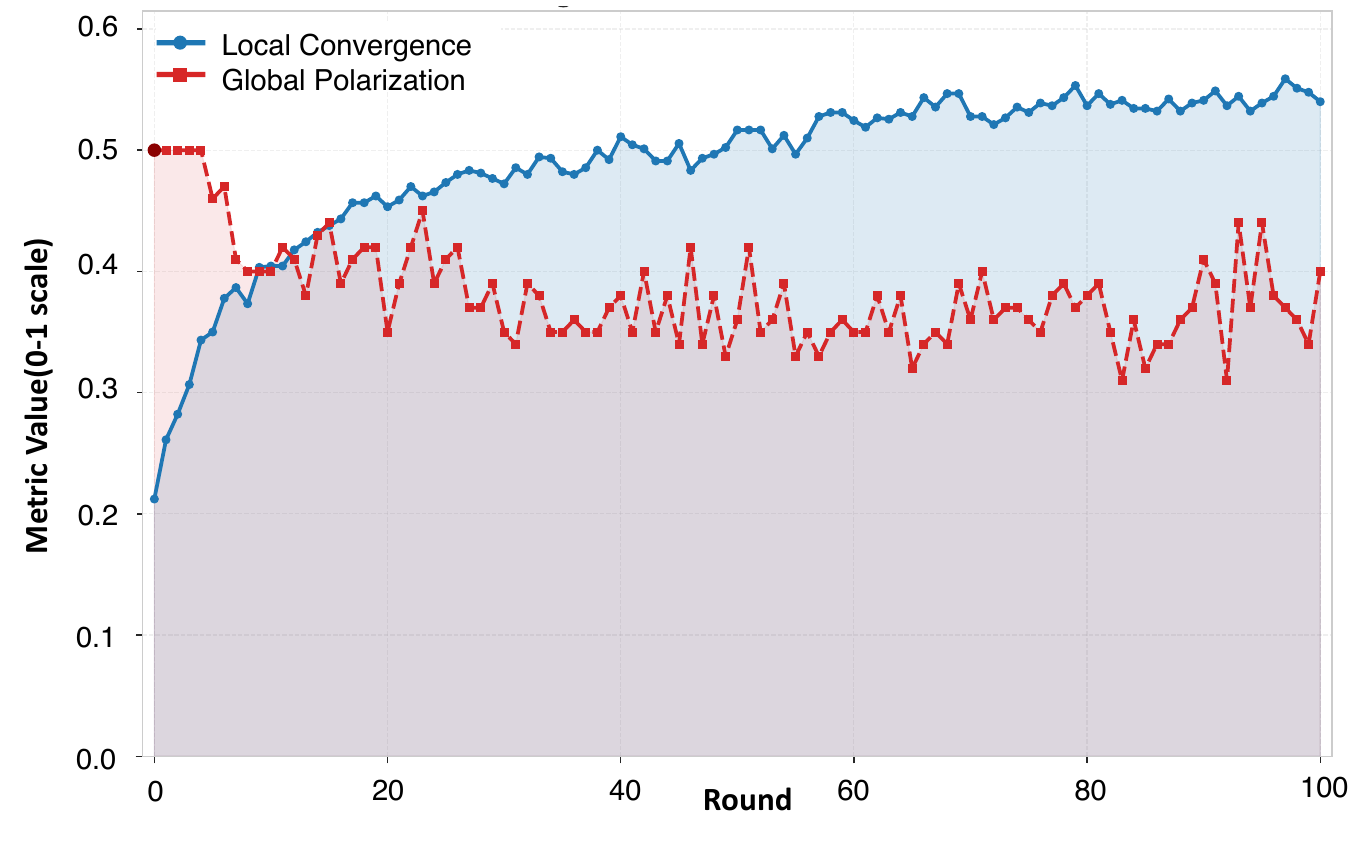}
        \caption{Local Convergence vs. Global Polarization over 100 simulation rounds. The blue solid line represents Local Convergence, which increases steadily as neighboring agents become culturally similar. The red dashed line represents Global Polarization, which initially rises during early diversification before stabilizing at a lower level. The crossover point (marked) indicates where local cultural similarity begins to dominate while maintaining distinct global cultural regions.}
        \label{fig:polarization}
    \vspace{-0.2cm}
    \end{figure}
    
    In Figure \ref{fig:polarization}, we present the evolution of local convergence and global polarization over 100 rounds of simulation.
    We can see: initially, local convergence increases while global polarization decreases, indicating that agents start to interact and form early cultural clusters. Around round 15, we observe a crossover point where local convergence continues to rise while global polarization begins to stabilize. This pattern demonstrates that local regions become increasingly homogeneous internally while maintaining distinct boundaries between different cultural groups.
    
    The stabilization of global polarization at approximately 0.35-0.40 indicates that the system sustains multiple cultural regions despite ongoing local convergence, which reaches over 0.50 by round 60 and continues to increase gradually. Notably, the occasional fluctuations in global polarization after round 40 suggest temporary merging and splitting of cultural regions as the system dynamically evolves.
    
    These results clearly demonstrate the coexistence of local convergence and global polarization—a key phenomenon predicted by Axelrod's theory. Our simulation not only replicates this theoretical expectation but also provides quantitative measures of these dynamics, offering deeper insights into the temporal evolution of cultural dissemination processes.

\begin{figure}[t]
    \centering
    \includegraphics[width=\textwidth]{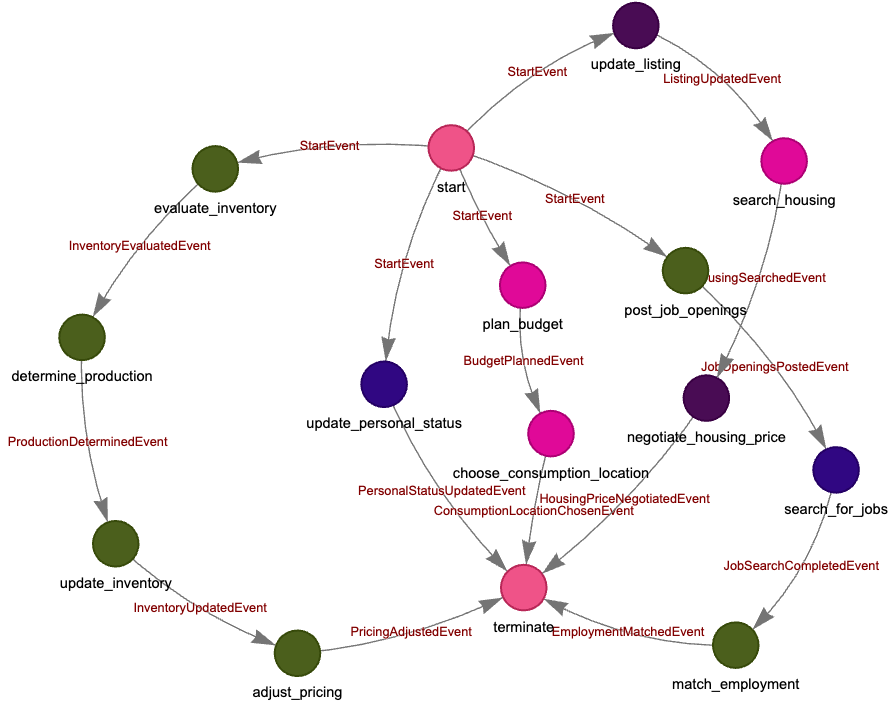}
    \caption{Agent Behavior Graph for the Brazilian Real Estate Market Simulation. The directed graph illustrates the interaction network between agents, with the central start node initiating various processes including updating listings, planning budgets, posting job openings, and updating personal status. Each node represents a specific agent action, while the directed edges indicate event flows between actions.}
    \label{fig:c-behavior_graph}
    \vspace{-0.3cm}
\end{figure}

\begin{figure}[t]
    \centering
    \includegraphics[width=0.95\textwidth]{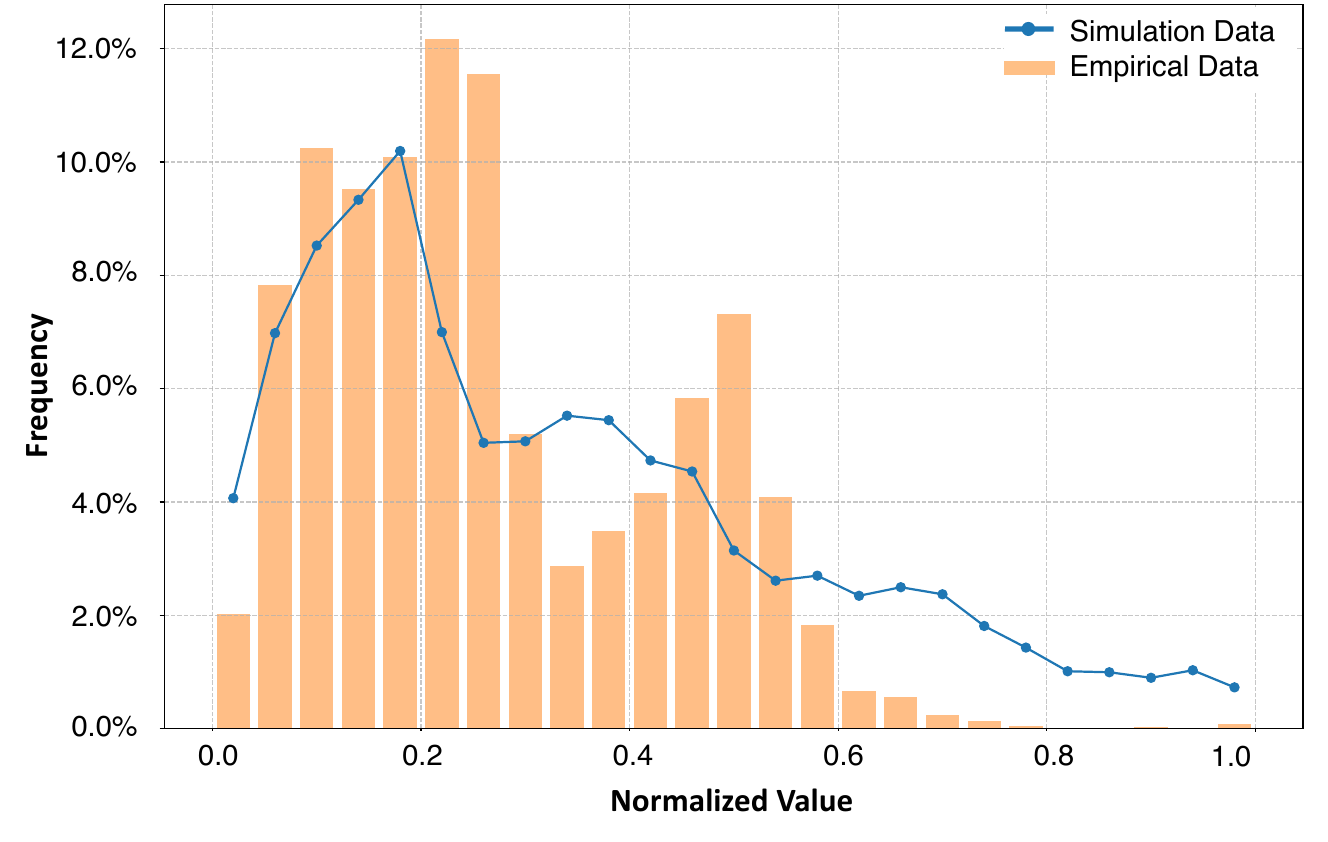}
    \caption{Normalized Frequency Distribution: Simulation vs Empirical Data. The bar chart represents the empirical data of Brazilian housing prices, while the blue line shows our simulation results. Both distributions are normalized to facilitate comparison despite different absolute price ranges.}
    \label{fig:house_price}
\end{figure}

    \subsubsection{Real-world data fitting}
     Beyond the above social theory verification, this section evaluates whether the simulation result of YuLan-OneSim can align well with the real-world data.
     
     Specifically, we focus on the scenario of Brazilian real estate market\cite{policyspace}. Our simulation incorporated six types of agents: individuals, families, government, banks, retail companies, and real estate companies. We simultaneously modeled three interconnected markets: the talent recruitment market, the retail goods market, and the real estate market.

     In our simulation, individuals submit resumes to companies, which then recruit them as employees and pay salaries. Individuals form families, which engage in consumption by purchasing products from retail companies or renting and buying properties. Retail companies produce goods and set pricing strategies, while real estate companies make decisions about housing construction and pricing. The agent behavior graph for these interactions is illustrated in Figure \ref{fig:c-behavior_graph}.

     We conducted our simulation with each round representing one month, running for a total of 12 rounds to model a full year. To validate our simulation against real-world data, we compared our results with the actual distribution of rental prices in Brazil from 2020 as reported in \cite{policyspace}.
   
     Figure \ref{fig:house_price} presents a comparison between our simulated housing price distribution and the empirical price distribution in Brazil. The simulation reproduces several key characteristics of the real-world data, particularly the multi-modal pattern with primary and secondary peaks at similar normalized values. The close alignment between simulated and actual distributions in the lower price ranges (0.05-0.25) indicates that our model captures the core market dynamics for the most common housing segment. While our simulation effectively reproduces the overall distribution shape and the long-tail characteristic typical of housing markets, there are some discrepancies in the mid-range values (0.45-0.55) where our model slightly underestimates frequency compared to empirical data. This difference likely stems from our simplified representation of the complex real-world factors that influence housing prices, such as neighborhood desirability, infrastructure quality, and historical valuation patterns. Nevertheless, these results demonstrate that our simulator can approximate real-world economic distributions with reasonable accuracy despite the inherent complexity of such systems, validating its potential for social science research applications.

    \subsection{Evaluation on the Simulation Efficiency and Scalability}
    In this section, we evaluate the efficiency and scalability of our simulator. Specifically, we continue using the Axelrod’s cultural dissemination scenario from Section~\ref{sec:emergence}, but scale it up by setting N=320, F=5, and q=10, resulting in a simulation with approximately 100,000 agents.
    We conduct experiments on a server equipped with 8 A100 40GB GPUs and 96 CPU cores. We employ our distributed simulation architecture with one master node and 96 worker nodes. For efficient LLM inference, we deploy 8 instances of the Qwen2.5-7B-Instruct~\cite{yang2024qwen2} using vLLM~\cite{kwon2023efficient}, distributing the computational load across the available GPUs to maximize throughput.
    We run our simulator for 3 rounds and report the time cost and events count for each round, as well as the average across all rounds.

    From Table~\ref{tab:performance}, we can see: our simulator is capable of executing simulations involving 100,000 agents with an average time cost of approximately 6,026 seconds per round. This performance makes large-scale social simulations practically feasible, enabling comprehensive investigations of population-level social dynamics that were previously constrained by computational limitations. Furthermore, our simulator achieves an average throughput of 49.41 events per second, processing over 294,000 interaction events in each round. These results highlight the efficiency and scalability of our distributed architecture in supporting large-scale agent-based modeling.

\begin{figure}[t]
\centering
\begin{minipage}{0.48\textwidth}
\centering
\includegraphics[width=\textwidth]{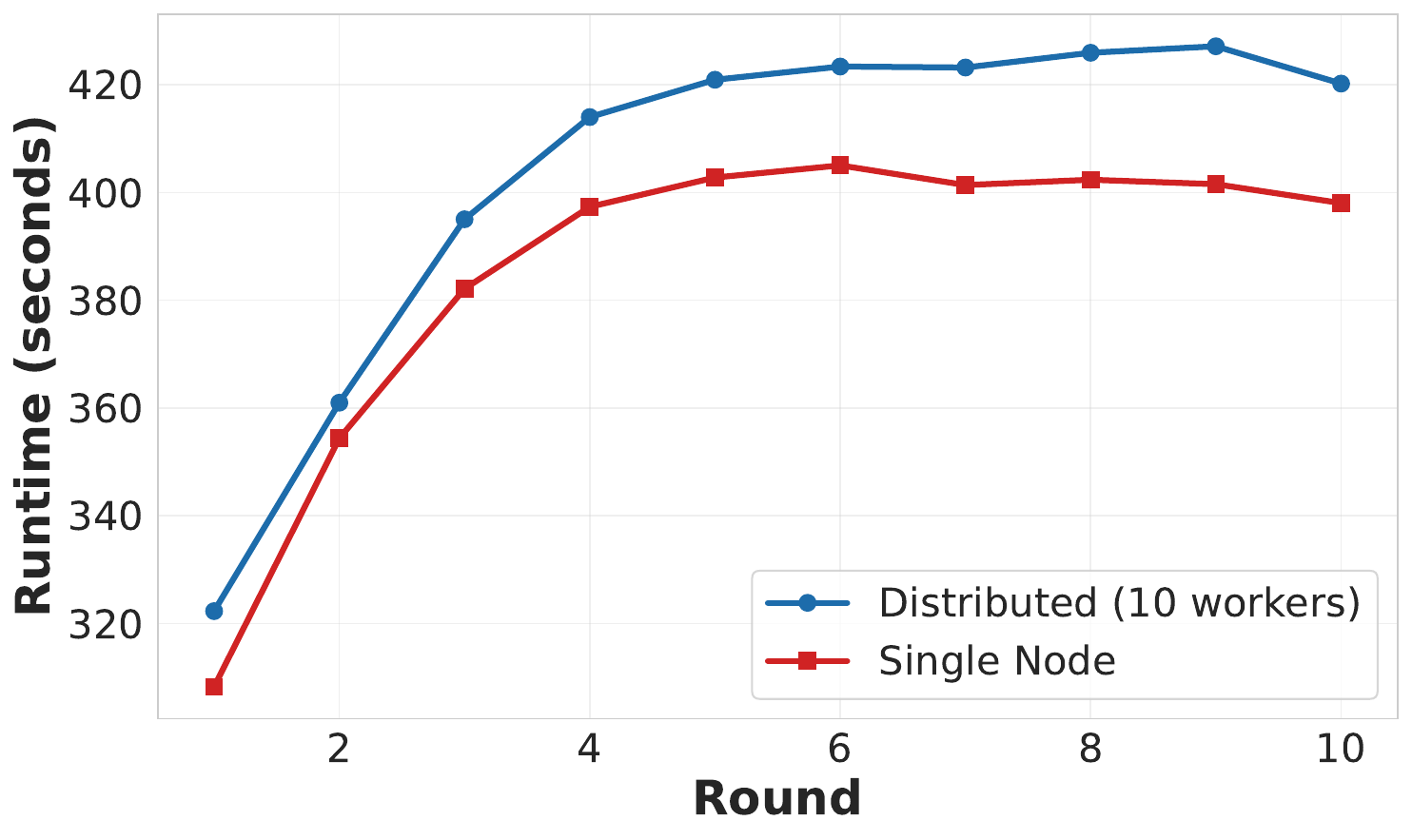}
\end{minipage}
\hfill
\begin{minipage}{0.48\textwidth}
\centering
\includegraphics[width=\textwidth]{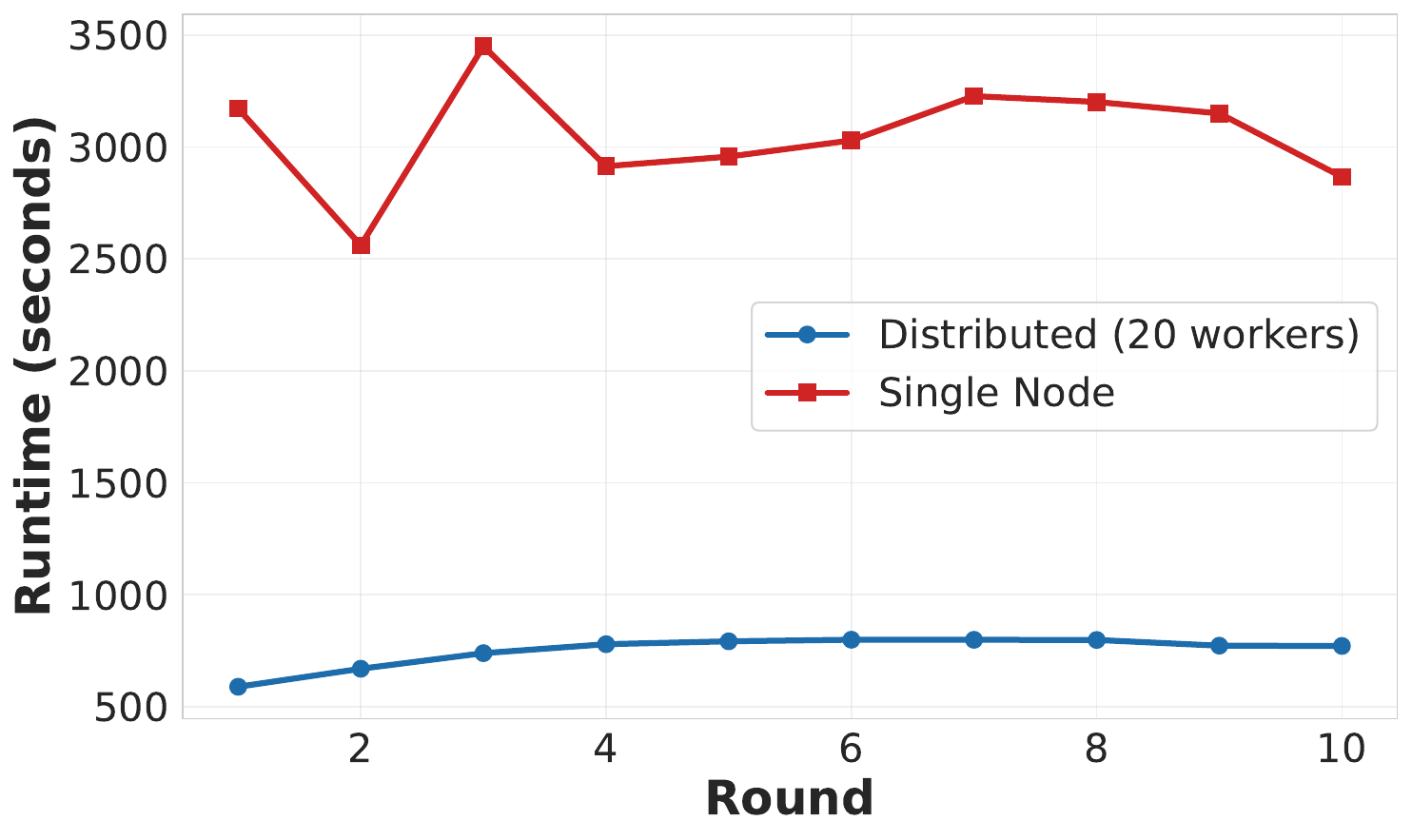}
\end{minipage}
\caption{{Performance comparison between our distributed architecture and the single-node approach.}}
\label{fig:scalability}
\vspace{-0.3cm}
\end{figure}

    To further assess the effectiveness of our distributed architecture, we compare it against the single-node approach in which agent behaviors are not distributed across multiple workers. 
    In our experiments, we find that when the number of agents exceeds 10,000, the single-node approach becomes impractical to execute.
    As a result, we set the number of agents as 5,000 and 10,000, respectively, and deploy 10 and 20 workers for each case to keep the number of agents per worker fixed.  
    From the results shown in Figure~\ref{fig:scalability}, we can see: 
    when the number of agents is 5,000, the signal-node method cost slightly less than the distributed architecture, which suggests that, if the simulation scale is not large, single-node method can be enough, and distributed architecture does not lead to lowered cost. 
    When we increase the number of agents to 10,000, the average time cost of our distributed architecture is about 750.76s, which reduces the cost of the single-node method by about 75\%.
    This result strongly validates our distributed architecture's scalability for high-concurrency, large-scale scenarios.

% \begin{figure}[t]
%     \centering
%     \includegraphics[width=0.95\textwidth]{fig/scalability_10k.pdf}
%     \caption{Performance comparison between our distributed architecture and the single-node approach.}
%     \label{fig:house_price}
% \end{figure}

\begin{table}[t]
\centering
\caption{
Evaluation results of simulation efficiency in terms of time cost and events count. For reference, we also report the throughput (i.e., events per second) and the processing rates of input and output tokens per second in each round.
}
\renewcommand\arraystretch{1.3}
\label{tab:performance}
\begin{tabular}{m{1.3cm}<{\centering}|
m{2.0cm}<{\centering}|
m{1.8cm}<{\centering}|
m{1.5cm}<{\centering}|
m{2.2cm}<{\centering}|
m{2.5cm}<{\centering}}\hline\hline
    \textbf{Round} & \textbf{Time Cost (s)} & \textbf{\#Events} & \textbf{Events/s} & \textbf{Input Tokens/s} & \textbf{Output Tokens/s}  \\
    \hline
    1 & 5,266 & 294,638 & 55.95 & 83,257 & 4,160  \\
    2 & 6,051 & 294,612 & 48.69 & 82,685 & 3,860  \\
    3 & 6,761 & 294,639 & 43.58 & 82,314 & 3,545 \\
    \hline
    {Average} & 6,026 & 294,630 & 49.41 & 82,752 & 3,855 \\
    \hline\hline
\end{tabular}
\vspace{-0.5cm}\end{table}

    \subsection{Evaluation on the AI Social Researcher}
    The AI Social Researcher is responsible for transforming user-specified research topics into valuable research questions and generating final technical reports. To evaluate its effectiveness, we assess both of these critical phases. We selected eight scenarios for our experiment, one from each domain implemented in our simulator.

    To evaluate the quality of the designed scenarios, 
    we assess four key aspects: relevance (alignment with the research topic), fidelity (accuracy in reflecting social phenomena), feasibility (practicality of implementation), and significance (research value).
    For the generated reports, we evaluate their insight (depth of analysis), structure (logical organization), content (completeness and correctness), and utility (practical applicability).
    To ensure evaluation efficiency, we employ GPT-4o~\cite{hurst2024gpt} to assess the outputs and provide an overall rating in the range of [1, 5].

\begin{table}[t]
    \centering
    \caption{Evaluation results on the scenario design quality.}
    \renewcommand\arraystretch{1.1}
    \label{tab:scenario_design}
    \begin{tabular}{m{2.3cm}|m{1.8cm}<{\centering}m{1.6cm}<{\centering}m{1.8cm}<{\centering}m{1.8cm}<{\centering}|m{1.5cm}<{\centering}}
        \hline\hline
        \textbf{Domain} & \textbf{Relevance} & \textbf{Fidelity} & \textbf{Feasibility} & \textbf{Significance} & \textbf{Overall} \\
        \hline
        Economics & 4 & 3 & 5 & 4 & 4.0 \\
        Sociology & 5 & 4 & 5 & 4 & 4.5 \\
        Politics & 3 & 3 & 4 & 4 & 3.5 \\
        Psychology & 5 & 4 & 5 & 4 & 4.5 \\
        Organization & 4 & 3 & 5 & 4 & 4.0 \\
        Demographics & 4 & 3 & 5 & 4 & 4.0 \\
        Law & 4 & 3 & 5 & 4 & 4.0 \\
        Communication & 5 & 4 & 5 & 4 & 4.5 \\
        \hline
        {Average} & 4.25 & 3.38 & 4.88 & 4.00 & 4.13 \\
        \hline\hline
    \end{tabular}
\end{table}

\begin{table}[t]
    \centering
    \caption{Evaluation results on the generated report quality.}
    \renewcommand\arraystretch{1.1}
    \label{tab:report_quality}
    \begin{tabular}{m{2.3cm}|m{1.8cm}<{\centering}m{1.6cm}<{\centering}m{1.8cm}<{\centering}m{1.8cm}<{\centering}|m{1.5cm}<{\centering}}
        \hline\hline
        \textbf{Domain} & \textbf{Insight} & \textbf{Structure} & \textbf{Content} & \textbf{Utility} & \textbf{Overall} \\
        \hline
        Economics & 4 & 5 & 4 & 4 & 4.0 \\
        Sociology & 3 & 4 & 3 & 3 & 3.0 \\
        Politics & 3 & 4 & 4 & 3 & 3.5 \\
        Psychology & 3 & 4 & 4 & 3 & 3.0 \\
        Organization & 3 & 4 & 3 & 2 & 3.0 \\
        Demographics & 3 & 3 & 3 & 3 & 3.0 \\
        Law & 4 & 4 & 4 & 3 & 4.0 \\
        Communication & 3 & 4 & 4 & 3 & 3.0 \\
        \hline
        {Average} & 3.25 & 4.00 & 3.63 & 3.00 & 3.31 \\
        \hline\hline
    \end{tabular}
    \vspace{-0.5cm}
\end{table}

    The evaluation results are presented in Table~\ref{tab:scenario_design} and Table~\ref{tab:report_quality}.
    We can see: 
    for the quality of the designed scenarios, our AI social researcher demonstrates strong performance across all metrics, with an average overall score of 4.13 out of 5. Particularly impressive was the system's ability to generate highly feasible simulation designs (average score: 4.88), indicating that the AI researcher can effectively translate conceptual research questions into implementable simulation scenarios. The system also excels in relevance (average score: 4.25), ensuring that generated scenarios accurately correspond to the intended research topics.
    
    For the quality of the generated reports, our AI social researcher performs well in structural organization (average score: 4.00) and content quality (average score: 3.63). This indicates that the system can effectively organize analytical findings into coherent, well-structured reports with reasonable depth and accuracy. The highest-rated reports were for Auction Market Dynamics (Economics) and Court Trial Simulation (Law), both scoring 4.0 overall. The relative weakness in insight (average score: 3.25) and utility (average score: 3.00) scores highlights potential areas for enhancement. While the AI researcher can competently analyze simulation data and present findings, it could be further improved to generate deeper insights and more actionable recommendations from simulation results. This is particularly evident in scenarios like Labor Market Matching Process, which receives the lowest utility score (average score: 2.0).

    Overall, the above evaluation demonstrates that our AI Social Researcher can successfully complete the full research cycle—from question formulation to report generation—with solid performance. The system excels particularly in translating research topics into feasible simulation designs, while showing room for improvement in extracting deeper insights from the simulation results.

    \section{Conclusion}\label{sec6}
    In this paper, we have introduced a novel social simulator called ``YuLan-OneSim'', which is featured in five aspects, that is, code-free scenario construction, comprehensive default scenarios, evolvable simulation, large-scale simulation and AI social researcher.
    We conduct extensive experiments to demonstrate the effectiveness and efficiency of our simulator. 
    We believe YuLan-OneSim makes a significant step towards the next generation of LLM agent based social simulator.
    In the future, we plan to incorporate more social theories to regularize agent behaviors in our simulator, and will also incorporate more spatial and multi-modal information to make the simulator more applicable.
    
    % \section*{Acknowledgments}
    % We would like to express our sincere gratitude to the AgentScope team from Alibaba Tongyi Lab for their outstanding work that has made this project possible. Their contributions have provided a solid foundation and essential tools that greatly enhanced the development process of YuLan-OneSim.
    
    % Special thanks to Xuchen Pan and Yaliang Li for their invaluable guidance and support throughout the development of this work.
\bibliographystyle{unsrtnat}
\bibliography{paper}

\clearpage

\appendix

\section{An Example of the ODD Protocol and its Generation Process}\label{odd}

\begin{tcolorbox}[
  colback=gray!4,
  colframe=gray!60!black,
  colbacktitle=gray!40!black,
  coltitle=white,
  title=ODD Protocol: Job Market Simulation,
  fonttitle=\bfseries,
  sharp corners,
  boxrule=0.5pt,
]

\textbf{Purpose:} Simulate the job seeking and recruitment process in the labor market to study information asymmetry, signaling, matching efficiency, bias, and the impact of different recruitment strategies and job-seeking behaviors on market outcomes and fairness.

\textbf{Agent Descriptions:}

\textit{Job Seeker:} Each job seeker has attributes such as ID, skills, education, experience, productivity, salary expectation, job preferences, social network, risk attitude, job search strategy, negotiation style, observable traits, and reservation wage.

\textit{Employer:} Employers are defined by industry, company size, compensation structure, hiring criteria, screening methods, potential biases, reputation, and historical hiring data. They post job openings with specific requirements (e.g., skills, education, experience) and select candidates accordingly.

\textbf{Environment:} Labor market tightness, economic conditions, regulatory environment, skill demand trends, and the rate of technological change.

\textbf{Interaction Patterns:} Job seekers and employers interact through recruitment processes. Job seekers apply to job postings; employers screen and evaluate applicants. Information is exchanged through applications, feedback, and negotiations.

\textbf{Communication Protocols:} Communication occurs via application systems or social/professional networks, involving job descriptions, resumes, screening feedback, and salary negotiations.

\textbf{Decision Mechanisms:}
\begin{itemize}[leftmargin=1.5em]
  \item Job seekers evaluate job opportunities based on skill match, attractiveness, and application cost.
  \item Employers define requirements and evaluate candidates based on fit and organizational priorities.
\end{itemize}

\textbf{Agent Behaviors:}
\begin{itemize}[leftmargin=1.5em]
  \item \textbf{Job seekers:} enter the market, set job goals, apply strategically, negotiate offers, and accept or reject jobs.
  \item \textbf{Employers:} post vacancies, filter applicants, conduct interviews, and make hiring decisions.
\end{itemize}

\textbf{Decision Algorithms:}
\begin{itemize}[leftmargin=1.5em]
  \item Job seekers use algorithms to assess job value, choose search strategies, and negotiate offers.
  \item Employers use algorithms to filter applicants and optimize hiring decisions based on constraints and priorities.
\end{itemize}

\textbf{Constraints:} Labor market conditions, regulatory frameworks, demand for skills, technological shifts, bias in screening, and influence of social networks.

\end{tcolorbox}

    \begin{tcolorbox}[
    title=ODD Protocol Gneration Process,
    colback=white,
    colframe=gray!60!black,
    colbacktitle=gray!60!black,
    coltitle=white,
    fonttitle=\bfseries,
    boxrule=0.5mm,
    arc=0mm,
     breakable,  % Allow the box to break across pages
    break at=8.8in  % Break before reaching bottom of page
]

% Stage 1
\textbf{Stage 1: Initial Research Topic}

\textbf{User:} I want to research the interaction between job seekers and employers in the job market.

\textbf{Agent:} Thank you for sharing your research topic. To build a complete simulation scenario, I need more information to develop the ODD protocol. Here's the initial framework I understand:

\begin{tcolorbox}[
    colback=gray!5,
    colframe=gray!40!black,
    boxrule=0.3mm,
    arc=1mm
]
\textbf{Research Topic:} Job Market Dynamics

\textbf{Purpose:} Study the interactions between job seekers and employers in the labor market.

\textbf{Main Agent Types:}
\begin{itemize}
    \item Job Seekers
    \item Employers
\end{itemize}
\end{tcolorbox}

Could you provide more information about:
\begin{enumerate}
    \item What specific job market phenomena or issues do you wish to study? (e.g., information asymmetry, bias, etc.)
    \item What key attributes should job seekers and employers have?
    \item What environmental factors would you like to include in the simulation? (e.g., economic conditions, skill demand trends, etc.)
\end{enumerate}

\vspace{1em}
\hrule
\vspace{1em}

% Stage 2
\textbf{Stage 2: User Provides Additional Information}

\textbf{User:} I'm primarily interested in studying how information asymmetry and screening mechanisms affect job market efficiency. Job seekers should have different skill levels, educational backgrounds, work experience, and salary expectations. Employers should have different hiring criteria, screening methods, and potential biases. Environmental factors include labor market tightness, economic conditions, and skill demand trends. I'd also like to understand how different job search strategies and recruitment approaches affect market outcomes and fairness.

\textbf{Agent:} Thank you for providing detailed information. I've expanded the ODD protocol based on your input. Please review the updated framework:

\begin{tcolorbox}[
    colback=gray!5,
    colframe=gray!40!black,
    boxrule=0.3mm,
    arc=1mm
]
\textbf{ODD Protocol: Job Market Simulation}

\textbf{Purpose:} Simulate the job seeking and recruitment process in the labor market to study information asymmetry, screening mechanisms and their impact on market efficiency, and how different job search and recruitment strategies affect market outcomes and fairness.

\textbf{Agent Descriptions:}
\begin{itemize}
    \item \textbf{Job Seekers:} Have attributes including skill levels, educational background, work experience, salary expectations, and job preferences.
    \item \textbf{Employers:} Have attributes including industry type, hiring criteria, screening methods, and potential biases.
\end{itemize}

\textbf{Environment:}
Labor market tightness, economic conditions, skill demand trends.

\textbf{Interaction Patterns:}
Job seekers apply for positions, employers screen and evaluate applicants. Information is exchanged through applications, feedback, and negotiations.
\end{tcolorbox}

Could you further elaborate on:
\begin{enumerate}
    \item What specific communication methods exist between job seekers and employers?
    \item How do job seekers evaluate and select job opportunities?
    \item What criteria do employers use to make final hiring decisions?
    \item What constraints should be considered in the simulation?
\end{enumerate}

\vspace{1em}
\hrule
\vspace{1em}

% Stage 3
\textbf{Stage 3: Complete ODD Protocol}

\textbf{User:} Communication methods include application systems and social/professional networks. Job seekers evaluate job opportunities based on skill match, attractiveness, and application cost. Employers make decisions based on candidate-job fit and organizational priorities. Constraints include labor market conditions, regulatory frameworks, skill demands, technological changes, bias in screening, and the influence of social networks.

\textbf{Agent:} The ODD protocol is now complete. Thank you for providing comprehensive information. Here is the final ODD protocol:

\begin{tcolorbox}[
    colback=gray!5,
    colframe=gray!40!black,
    boxrule=0.3mm,
    arc=1mm
]
\textbf{ODD Protocol: Job Market Simulation}

\textbf{Purpose:} Simulate the job seeking and recruitment process in the labor market to study information asymmetry, signaling, matching efficiency, bias, and the impact of different recruitment strategies and job-seeking behaviors on market outcomes and fairness.

\textbf{Agent Descriptions:}
\begin{itemize}
    \item \textbf{Job Seeker:} Each job seeker has attributes including ID, skills, education, experience, productivity, salary expectation, job preferences, social network, risk attitude, job search strategy, negotiation style, observable traits, and reservation wage.
    \item \textbf{Employer:} Employers are defined by industry, company size, compensation structure, hiring criteria, screening methods, potential biases, reputation, and historical hiring data. They post job openings with specific requirements (e.g., skills, education, experience) and select candidates accordingly.
\end{itemize}

\textbf{Environment:}
Labor market tightness, economic conditions, regulatory environment, skill demand trends, and the rate of technological change.

\textbf{Interaction Patterns:}
Job seekers and employers interact through structured recruitment processes. Job seekers apply to job postings; employers screen and evaluate applicants. Information is exchanged through applications, feedback, and negotiations.

\textbf{Communication Protocols:}
Communication occurs via application systems or social/professional networks, involving job descriptions, resumes, screening feedback, and salary negotiations.

\textbf{Decision Mechanisms:}
\begin{itemize}
    \item Job seekers evaluate job opportunities based on skill match, attractiveness, and application cost.
    \item Employers define requirements and evaluate candidates based on fit and organizational priorities.
\end{itemize}

\textbf{Agent Behaviors:}
\begin{itemize}
    \item \textbf{Job seekers:} enter the market, set job goals, apply strategically, negotiate offers, and accept or reject jobs.
    \item \textbf{Employers:} post vacancies, filter applicants, conduct interviews, and make hiring decisions.
\end{itemize}

\textbf{Constraints:}
Labor market conditions, regulatory frameworks, demand for skills, technological shifts, bias in screening, and influence of social networks.
\end{tcolorbox}

Would you like to make any adjustments or additions? We can begin building the job market simulation scenario based on this ODD protocol.

\end{tcolorbox}

    \section{Details about the 50 Default Scenarios}\label{app:detail}
    In this section, we provide a detailed overview of the scenarios implemented in our simulator. A total of 50 scenarios have been developed, covering eight distinct domains. In the following subsections, we present these scenarios domain by domain. For each scenario, we include a comprehensive description along with its corresponding ODD protocol, behavior graph, and agent profiles.

    % \subsection{Economics}

    % \subsubsection{xxx}
    % $\bullet$ \textbf{Detailed description}

    % $\bullet$ \textbf{ODD protocol}

    % $\bullet$ \textbf{Behavior graph}

    % $\bullet$ \textbf{Agent profiles}

    % \subsection{Sociology}

    % \subsection{Politics}

    % \subsection{Psychology}

    % \subsection{Organization}

    % \subsection{Demographics}

    % \subsection{Law}

    % \subsection{Communication}

    % \bibliographystyle{unsrtnat}
    % \bibliography{paper}

    \subsection{Economics}

    \subsubsection{Customer Satisfaction and Loyalty Model}
    $\bullet$ \textbf{Detailed description}
    \begin{textbox}
        \underline{\textcolor{red}{Customer Satisfaction and Loyalty Model}}\\
    
        System Goal:
      
        Observe how dynamic changes in service quality and product experience impact customer satisfaction, loyalty, and purchasing decisions.\\

        Agent Types:

        Customers are the core simulation objects, each with attributes: perceived service quality (range [0,1]), product experience score (range [0,1]), satisfaction (calculated from service and product experience), loyalty (accumulated based on historical satisfaction), purchase probability (determined by loyalty and current satisfaction). \\

        Environment Description:

        Merchant system with baseline service quality and product quality as system-level variables influencing all customers' perceptions. The system operates on discrete time steps, each representing a standardized decision cycle over 10 periods.
    
    \end{textbox}

    $\bullet$ \textbf{Behavior graph}

    \begin{figure}[h]
        \centering
        \includegraphics[width=\textwidth]{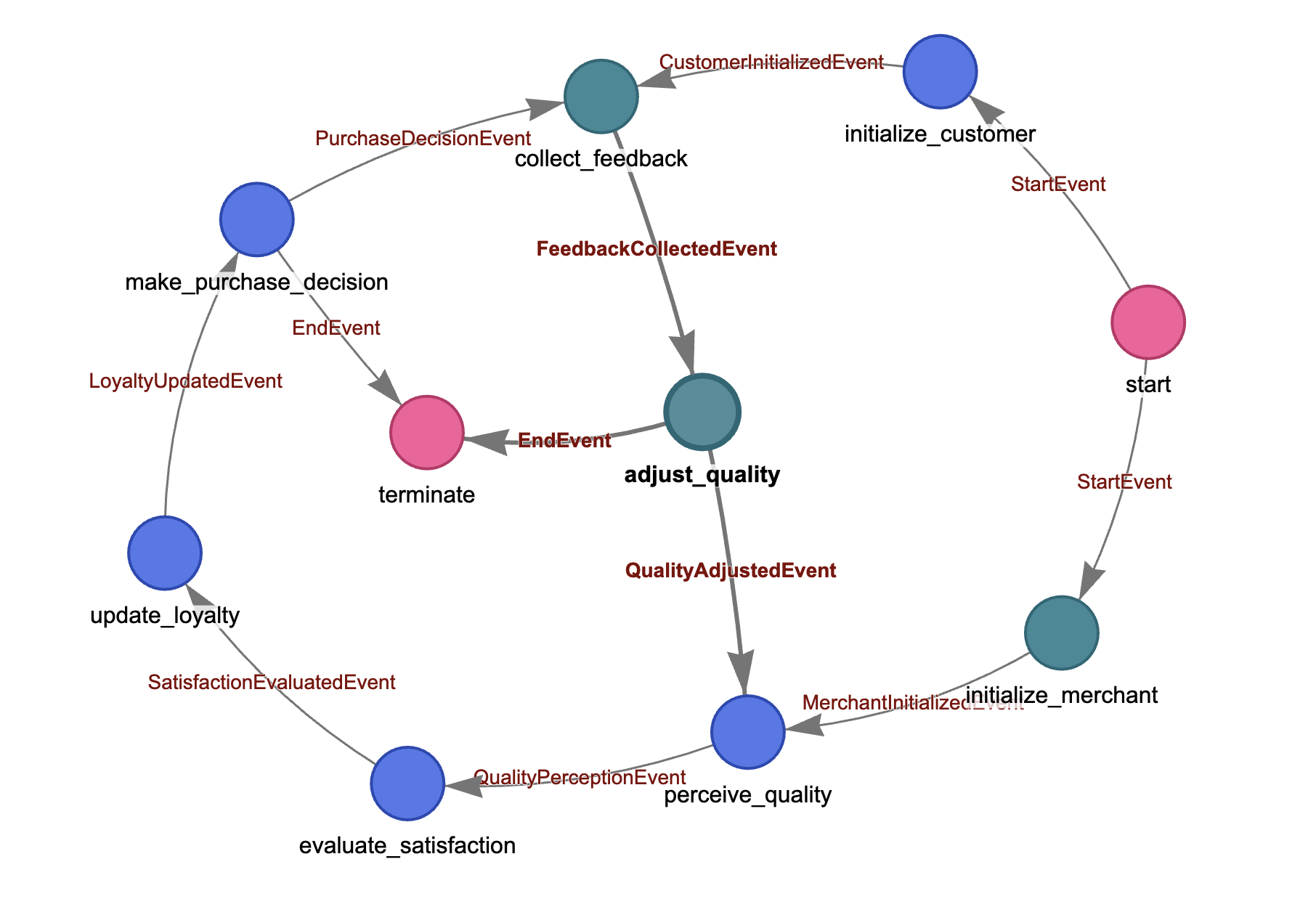}
        \caption{Behavior graph of Customer Satisfaction and Loyalty Model.}
    \end{figure}

    \subsubsection{Collective Action Problem}
    $\bullet$ \textbf{Detailed description}
    \begin{textbox}
        \underline{\textcolor{red}{Collective Action Problem}}\\
        
        System Goal:
      
        Simulate individual decision-making in collective action scenarios to analyze conflicts between individual and collective interests and explore solutions to collective action dilemmas.\\

        Agent Types:

        Individual: Each individual decides whether to cooperate in collective actions based on personal cost and benefit. 
        Group: Evaluates whether collective goals are met based on individual behaviors.\\

        Environment Description:

        The environment consists of multiple individuals and a group, with individuals having random cooperation willingness and the group having set goals and benefits.
    
    \end{textbox}

    $\bullet$ \textbf{Behavior graph}

    \begin{figure}[h]
        \centering
        \includegraphics[width=\textwidth]{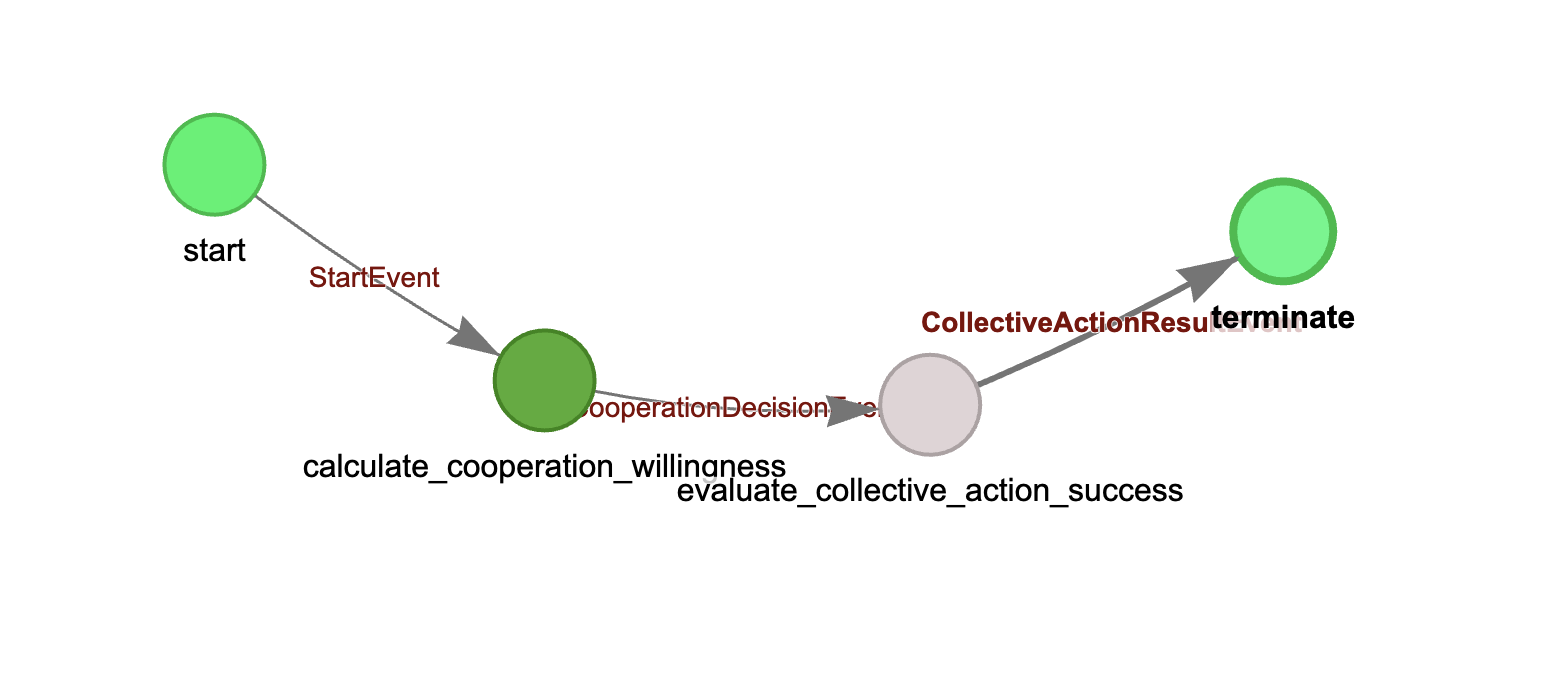}
        \caption{Behavior graph of Collective Action Problem.}
    \end{figure}

    \subsubsection{Bank Reserves}
    $\bullet$ \textbf{Detailed description}
    \begin{textbox}
        \underline{\textcolor{red}{Bank Reserves}}\\
    
        System Goal:
      
        Simulate the impact of reserve requirements on bank lending behavior and financial stability when reserves are insufficient.\\

        Agent Types:

        Banks and Customers; banks perform lending operations based on reserve levels, while customers apply for loans based on economic conditions.\\

        Environment Description:

        A banking system where banks and customers interact, influenced by economic cycles and reserve requirements.
    
    \end{textbox}

    $\bullet$ \textbf{Behavior graph}

    \begin{figure}[h]
        \centering
        \includegraphics[width=\textwidth]{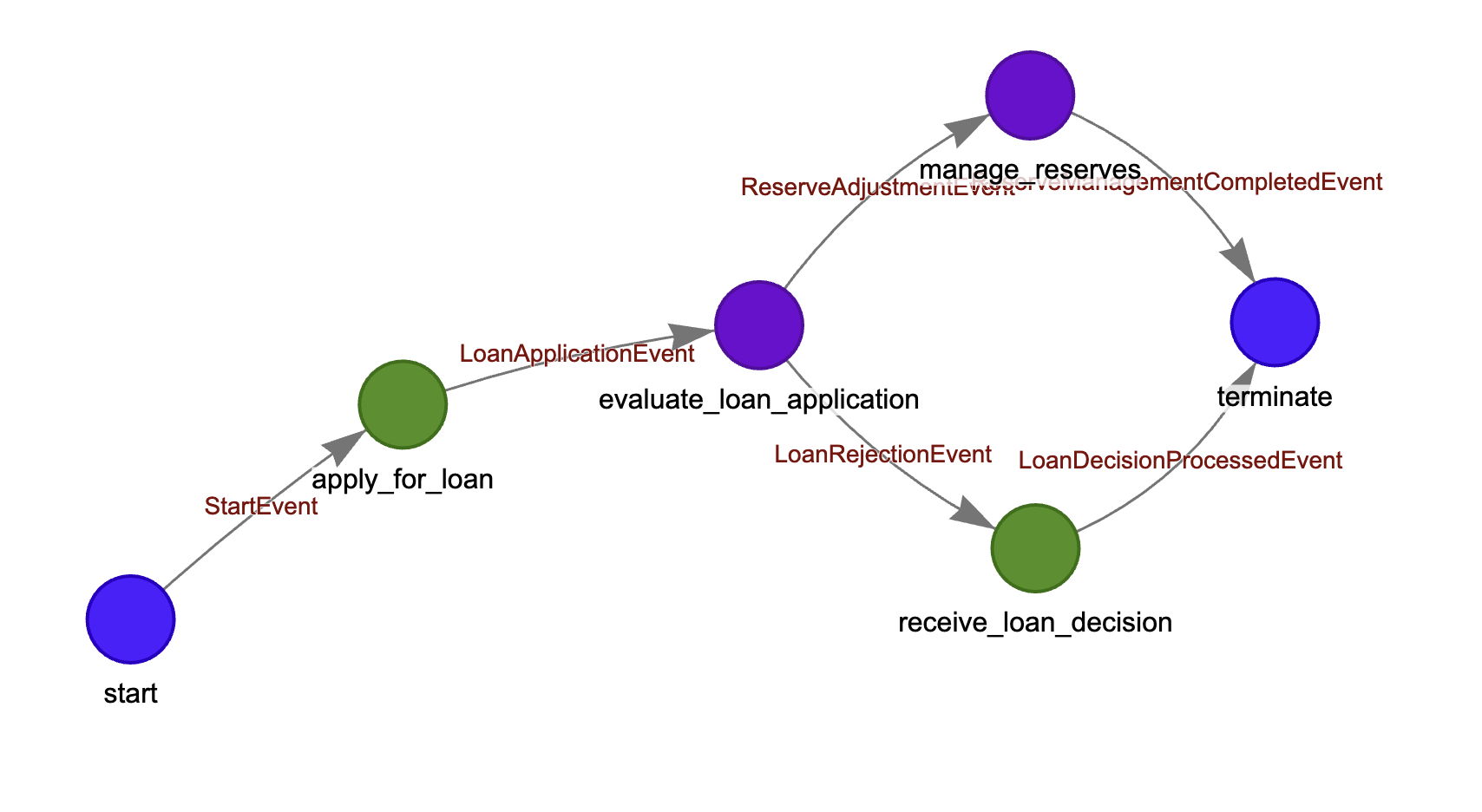}
        \caption{Behavior graph of Bank Reserves.}
    \end{figure}

    \subsubsection{Cash Flow}
    $\bullet$ \textbf{Detailed description}
    \begin{textbox}
        \underline{\textcolor{red}{Cash Flow}}\\
        
        System Goal:
      
        The goal is to simulate a simple economic system where multiple companies manage cash flow and make decisions to sustain operations, aiming to study the impact of cash flow on company decision-making, capital efficiency, and survival capability.\\

        Agent Types:

        Companies managing cash flow, Banks providing loan services, Consumers affecting company revenue.\\

        Environment Description:

        The environment includes companies with varying cash reserves, banks with loan conditions and interest rates, and consumers whose spending habits affect company revenue.
    
    \end{textbox}

    $\bullet$ \textbf{Behavior graph}
    \begin{figure}[h]
        \centering
        \includegraphics[width=\textwidth]{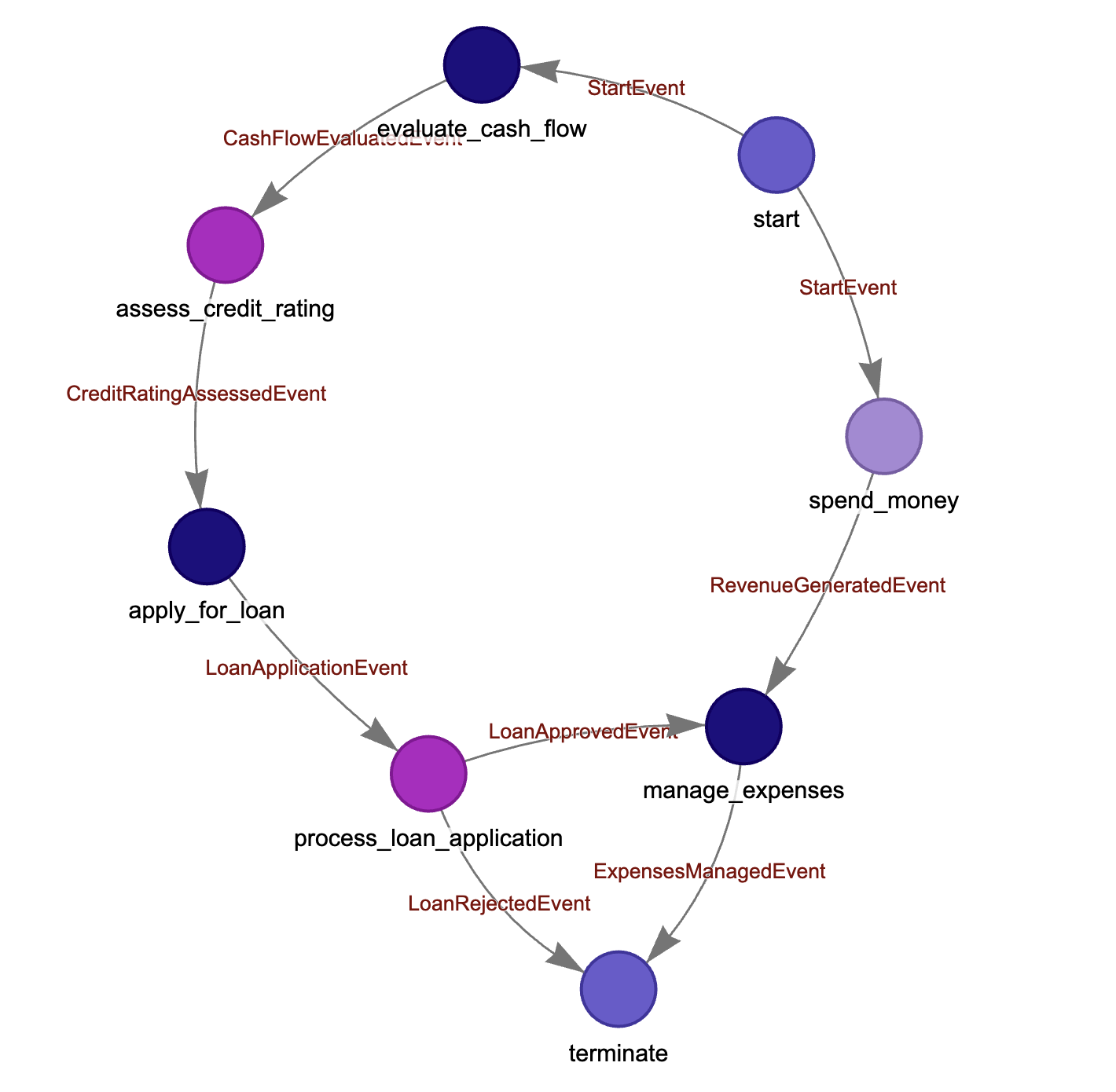}
        \caption{Behavior graph of Cash Flow.}
    \end{figure}

    \subsubsection{Rational Choice Theory}
    $\bullet$ \textbf{Detailed description}
    \begin{textbox}
        \underline{\textcolor{red}{Rational Choice Theory}}\\
        
        System Goal:
      
        Simulate how individuals make choices based on rational decision-making, validating the basic assumptions of rational choice theory, focusing on how agents evaluate the utility of options and make decisions based on utility maximization.\\
      
        Agent Types:

        Individual agents representing rational decision-making individuals, each with different resources, needs, and goals influencing their decision processes.\\

        Environment Description:

        The simulation environment includes multiple scenarios or decision options, each with different costs and benefits, requiring agents to weigh these factors to make optimal decisions.
    
    \end{textbox}

    $\bullet$ \textbf{Behavior graph}
    \begin{figure}[h]
        \centering
        \includegraphics[width=\textwidth]{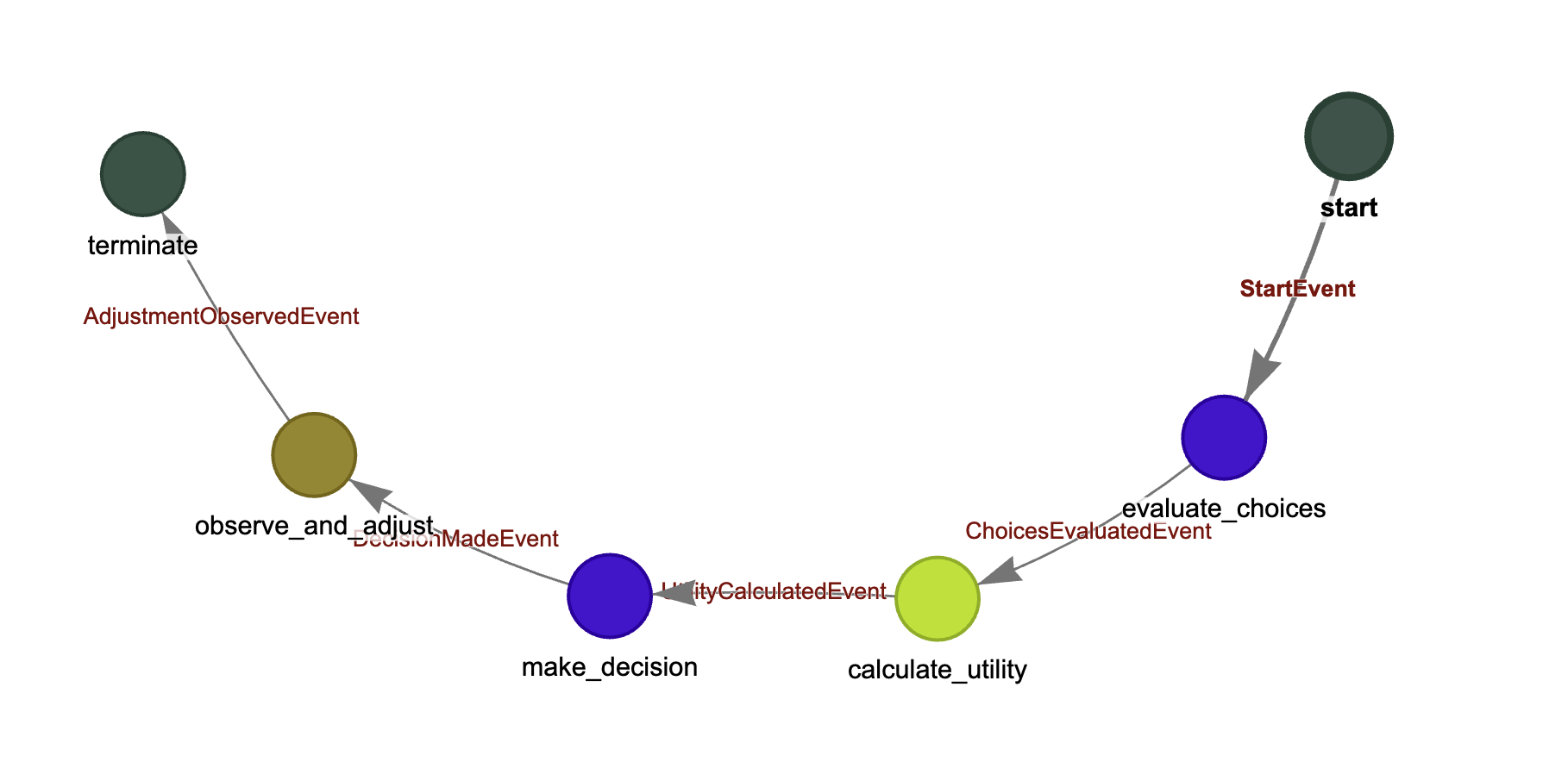}
        \caption{Behavior graph of Rational Choice Theory.}
    \end{figure}

    \subsubsection{Auction Market Dynamics}
    $\bullet$ \textbf{Detailed description}
    \begin{textbox}
     \underline{\textcolor{red}{Auction Market Dynamics}}\\
     
        System Goal:
      
        Simulate bidding behavior, information revelation processes, and market efficiency under various auction mechanisms to study strategy evolution among different types of buyers and sellers, and how market design impacts welfare distribution and efficiency.\\

        Agent Types:

        Includes Buyers characterized by id, private value, risk preference, budget, bidding strategy, learning rate, experience, and reputation; Sellers with id, reserve price, production cost, auction preference, reputation, and strategy; Auction Platform with mechanism type, fee structure, transparency level, history, and active auctions; Speculators with id, capital, strategy, information level, and risk tolerance.\\

        Environment Description:

        The environment is characterized by market liquidity, information asymmetry, market volatility, and regulation strictness, affecting agent interactions and decision-making processes.

    \end{textbox}

    $\bullet$ \textbf{Behavior graph}
    \begin{figure}[h]
        \centering
        \includegraphics[width=\textwidth]{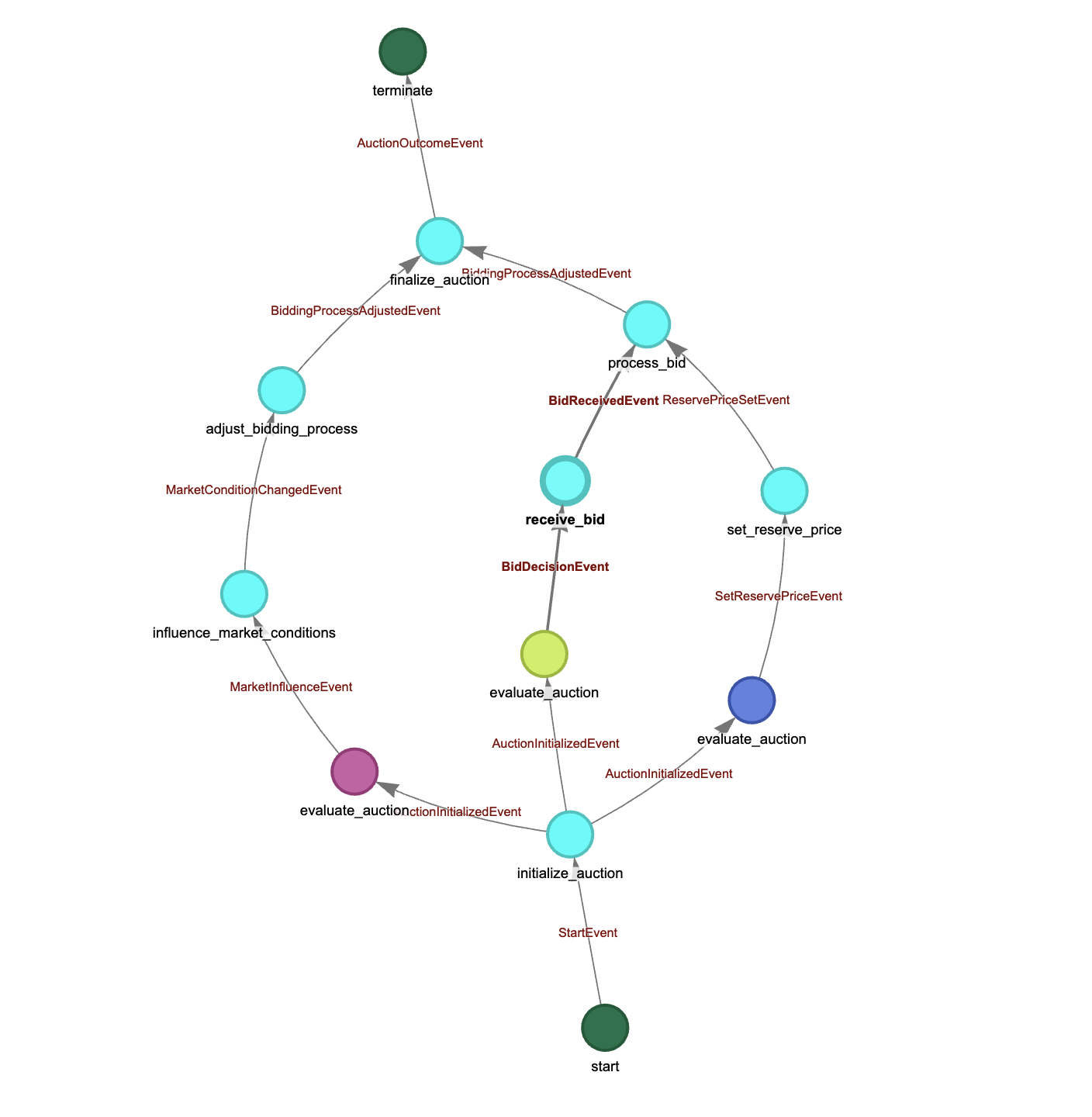}
        \caption{Behavior graph of Auction Market Dynamics.}
    \end{figure}

    \subsection{Sociology}

    \subsubsection{Cultural Capital Theory}

    $\bullet$ \textbf{Detailed description}

    \begin{textbox}

     \underline{\textcolor{red}{Cultural Capital Theory}}\\

        System Goal:
        
        To study how cultural capital influences individual social mobility and to explore the effects of education and distribution of social resources.\\

        Agent Types:

        Individuals with varying levels of cultural capital and social structures representing different social classes.\\

        Environment Description:

        A societal system with different social classes where individuals possess varying levels of cultural, economic, and social capital.\\

        Agents Initialization:

        Individuals are created within different social classes and assigned varying levels of cultural capital.\\

        Social Rules:

        High cultural capital individuals are more likely to access social resources.
    \end{textbox}

    $\bullet$ \textbf{Behavior graph}

    \begin{figure}[h]
        \centering
        \includegraphics[width=\textwidth]{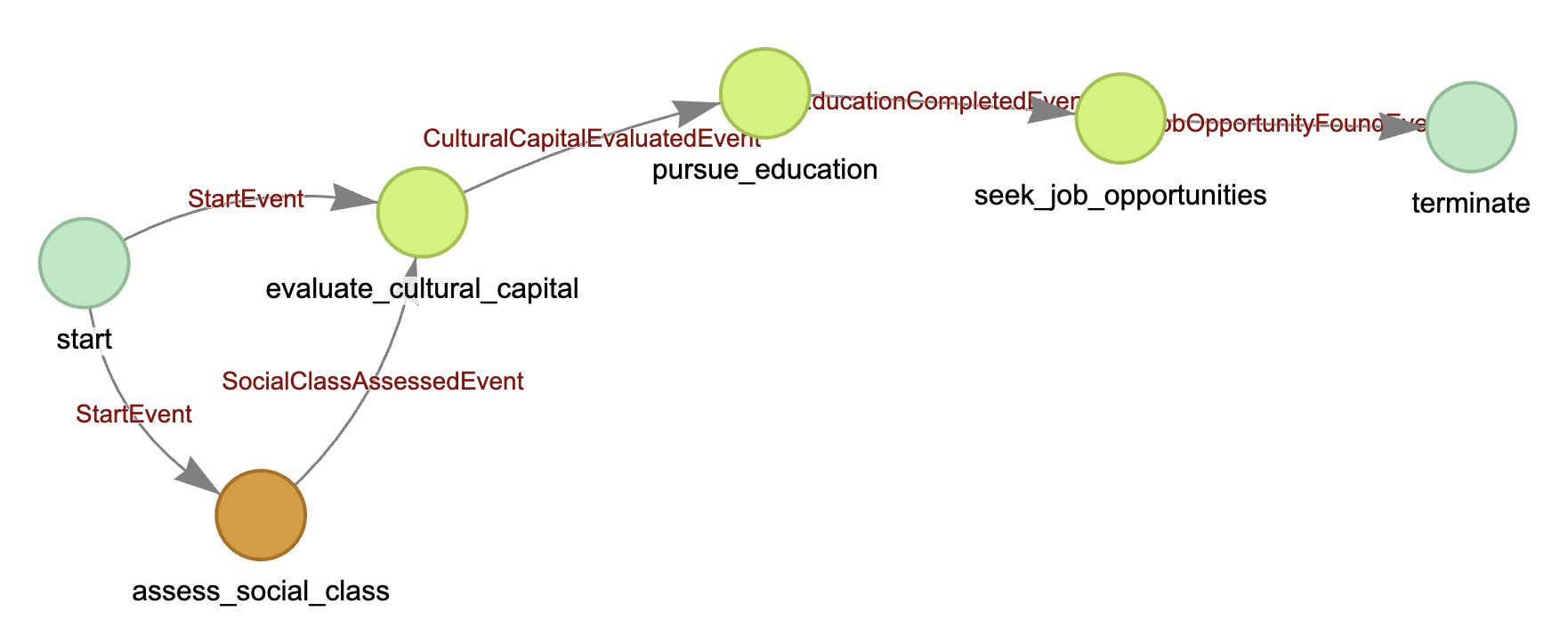}
        \caption{Behavior graph of Cultural Capital Theory.}
    \end{figure}

    \subsubsection{Social Capital Theory}

    $\bullet$ \textbf{Detailed description}

    \begin{textbox}
     \underline{\textcolor{red}{Social Capital Theory}}\\
     
        System Goal:
        
        Study how social capital flows between individuals and influences cooperation, trust, and resource access.\\

        Agent Types:

        Individuals with varying levels of social capital; Social Network representing relationships between individuals.\\

        Environment Description:

        A social network environment where individuals interact based on their social capital and relationships.\\

        Social Capital Importance:

        Social capital is crucial for accessing resources and forming new connections.\\

        Research Applications:

        The model can be applied to analyze social media, corporate management, and development in impoverished communities.
    \end{textbox}

    $\bullet$ \textbf{Behavior graph}

    \begin{figure}[h]
        \centering
        \includegraphics[width=\textwidth]{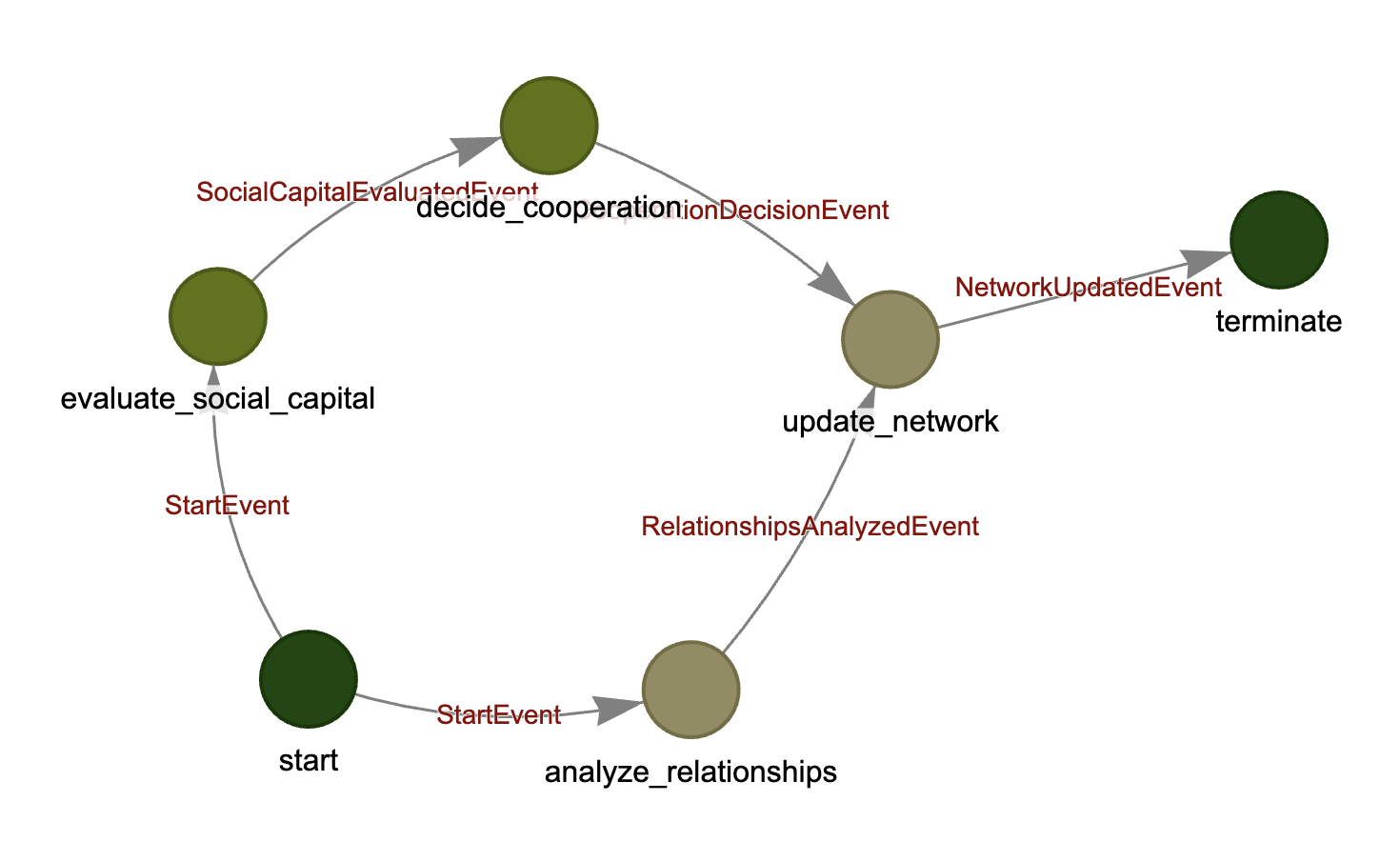}
        \caption{Behavior graph of Social Capital Theory.}
    \end{figure}

    \subsubsection{Norm Formation Theory}
    $\bullet$ \textbf{Detailed description}
    \begin{textbox}
     \underline{\textcolor{red}{Norm Formation Theory}}\\
     
        System Goal:
        
        To study how group norms form through interaction and observation, and to explore how these norms are maintained or changed.\\
        
        Agent Types:
        
        Entities (Agents) include Individuals, which are intelligent agents with different behavior tendencies, and Social Groups, which are interactive units composed of multiple individuals.\\

        Environment Description:
        
        The environment consists of multiple individuals with varying behavior tendencies randomly assigned to different social groups, where they interact and observe each other to form or adapt to group norms.
      
    \end{textbox}

    $\bullet$ \textbf{Behavior graph}
    \begin{figure}[h]
        \centering
        \includegraphics[width=\textwidth]{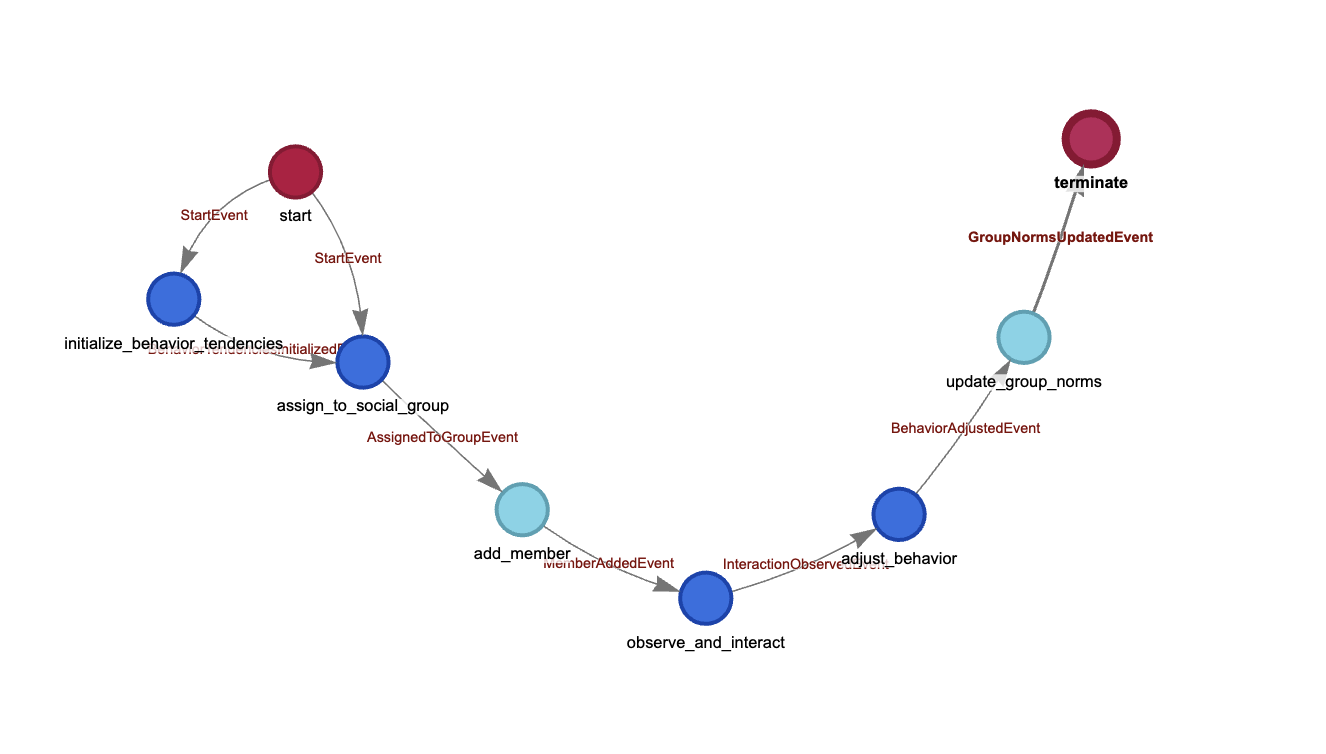}
        \caption{Behavior graph of Norm Formation Theory.}
    \end{figure}

    \subsubsection{Social Relations Theory}
    $\bullet$ \textbf{Detailed description}
    \begin{textbox}
     \underline{\textcolor{red}{Social Relations Theory}}\\
     
        System Goal:
        
        The goal of this simulation is to use LLMs to simulate how employees in an organization are influenced by social relationships, team interactions, and experiences of being noticed, impacting their work motivation, job satisfaction, and productivity. The simulation focuses on interactions between employees, leaders, and the organizational environment, exploring how social relationships, sense of belonging, and teamwork affect employee work behavior.\\
        
        Agent Types:
        
        Each agent represents an employee who can exhibit different behaviors, emotional responses, and work motivations. Agents have social and emotional needs, and the quality of their relationships with colleagues and leaders significantly impacts their job satisfaction and motivation.\\

        Environment Description:
        
        The environment simulates the organizational setting where employees interact with leaders, colleagues, and the broader organizational context. It includes elements such as social interactions, team dynamics, leadership styles, and feedback mechanisms that influence employee behavior and emotions.
      
    \end{textbox}

    $\bullet$ \textbf{Behavior graph}
    \begin{figure}[h]
        \centering
        \includegraphics[width=\textwidth]{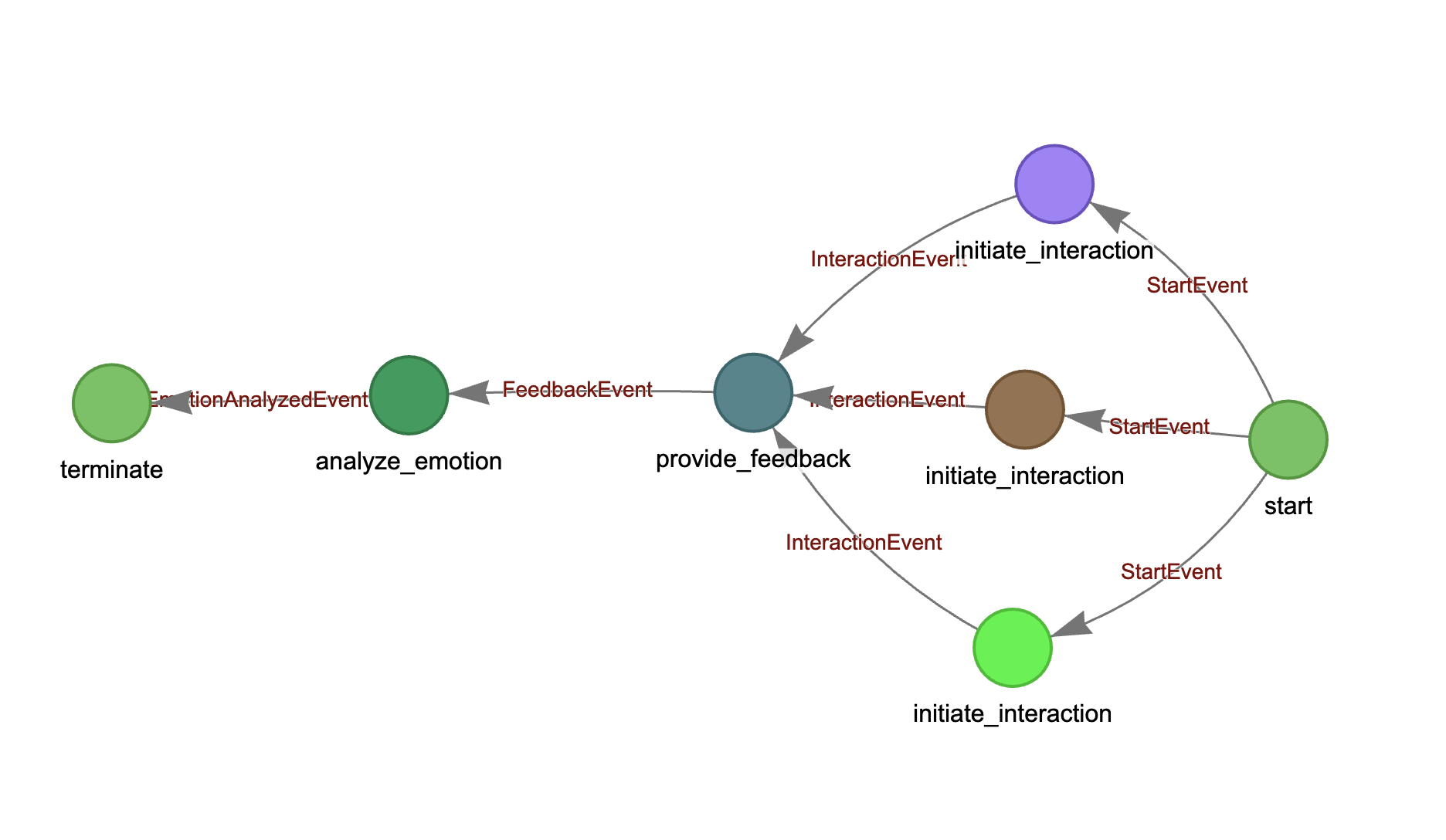}
        \caption{Behavior graph of Social Relations Theory.}
    \end{figure}

    \subsubsection{Theory of Planned Behavior}
    $\bullet$ \textbf{Detailed description}
    \begin{textbox}
     \underline{\textcolor{red}{Theory of Planned Behavior}}\\
     
        System Goal:
        
        Simulate the core mechanisms of the Theory of Planned Behavior (TPB) using a large language model (LLM) to explore the interactions between attitude, subjective norm, and perceived behavioral control in forming behavioral intentions and predicting actual behavior.\\
        
        Agent Types:
        
        Agents represent individuals with cognitive capabilities whose behavioral intentions are influenced by TPB components.\\

        Environment Description:
        
        The environment provides context with social norms such as group pressure and situational constraints like resource availability.
      
    \end{textbox}

    $\bullet$ \textbf{Behavior graph}
    \begin{figure}[h]
        \centering
        \includegraphics[width=\textwidth]{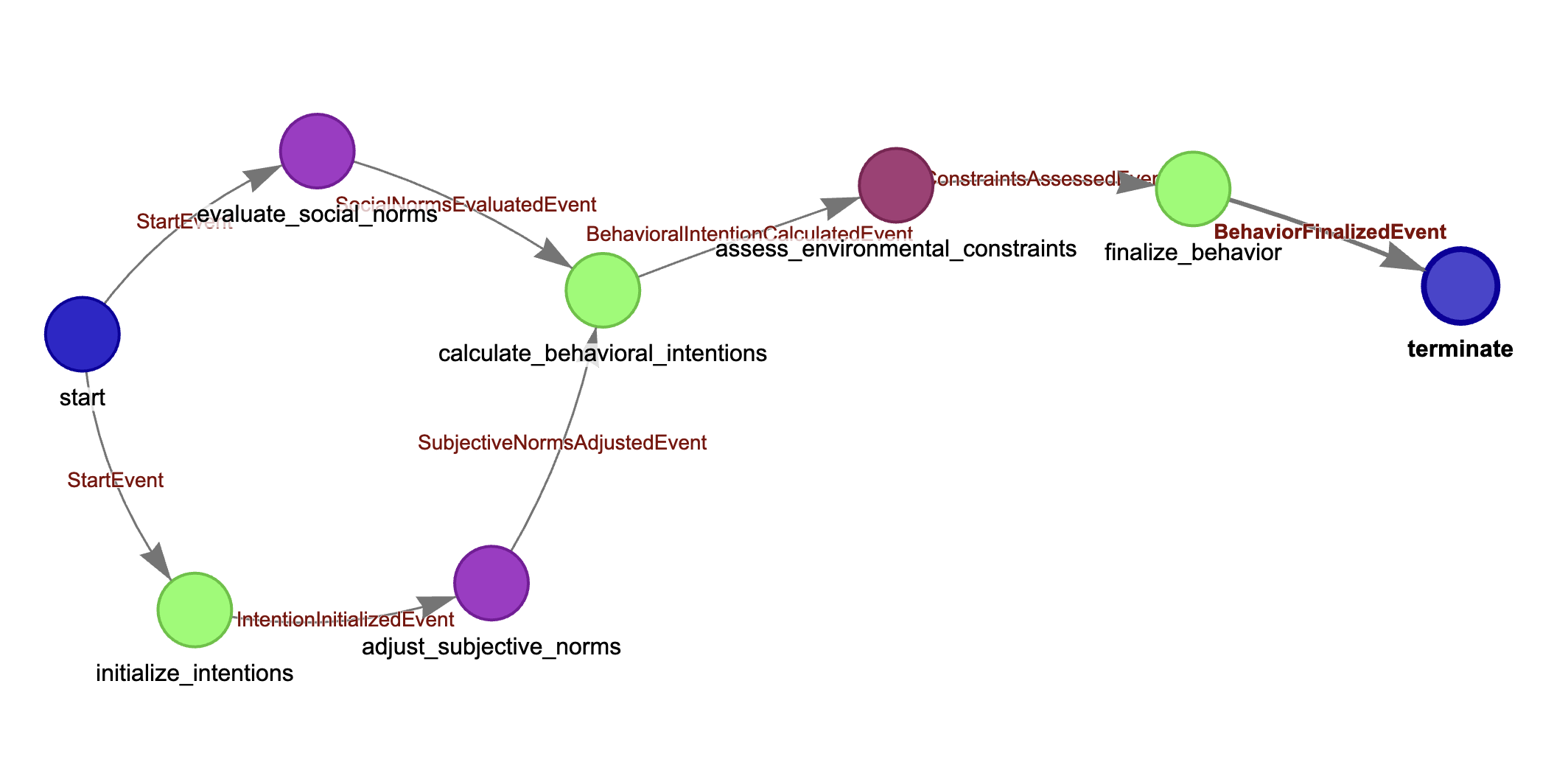}
        \caption{Behavior graph of Theory of Planned Behavior.}
    \end{figure}

    \subsubsection{Social Stratification Network}
    $\bullet$ \textbf{Detailed description}
    \begin{textbox}
     \underline{\textcolor{red}{Social Stratification Network}}\\
     
        System Goal:
        
        Simulate reciprocal behaviors and social capital accumulation in social resource exchange networks, study how different initial resource endowments and exchange rules lead to social stratification, and explore mechanisms of social mobility and intergenerational transmission.\\
        
        Agent Types:
        
        1. Individual: characterized by id, age, economic, cultural, social, and symbolic capital, occupation, income, exchange history, reciprocity expectation, trust threshold, social network, and generation. 2. Family: characterized by id, members, collective resources, inheritance rules, education investment, social status, and residential area. 3. Educational Institution: characterized by type, quality, admission criteria, tuition fees, prestige, and alumni network. 4. Employer: characterized by industry, size, hiring criteria, compensation structure, promotion paths, and status value.\\

        Environment Description:
        
        The environment includes variables such as inequality index, social mobility rate, meritocracy level, welfare system strength, and economic growth rate.
      
    \end{textbox}

    $\bullet$ \textbf{Behavior graph}
    \begin{figure}[h]
        \centering
        \includegraphics[width=\textwidth]{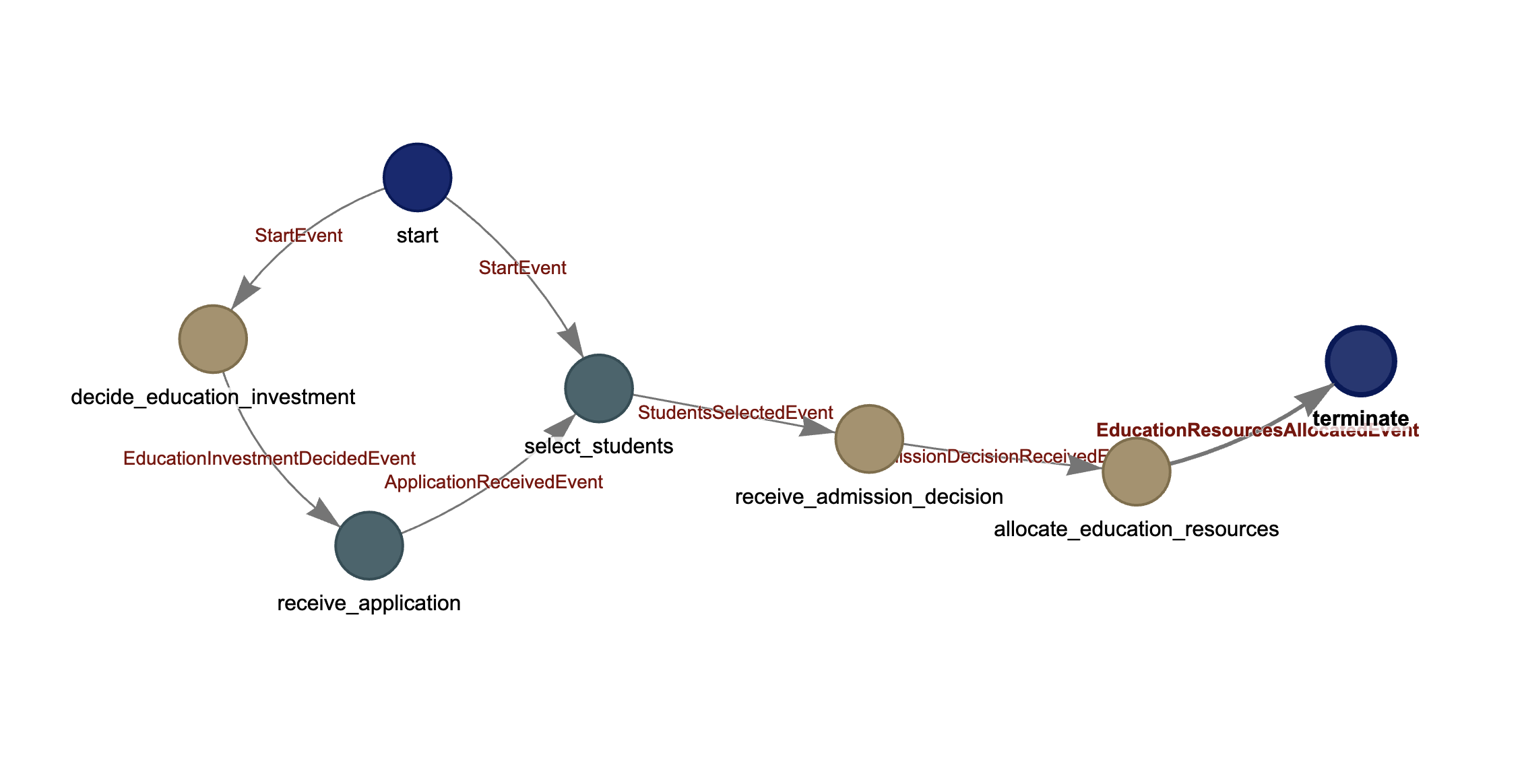}
        \caption{Behavior graph of Social Stratification Network.}
    \end{figure}

    \subsection{Politics}

    \subsubsection{Public Opinion Polling}

    $\bullet$ \textbf{Detailed description}

    \begin{textbox}
     \underline{\textcolor{red}{Public Opinion Polling}}\\
     
        System Goal:
        
        Study how public opinion polling gathers information through voter interaction, analyze dynamic changes in voter preferences, and understand how polling data influences policy choices.\\

        Agent Types:

        Voters, who are participants in opinion polls and express preferences for policies or candidates; Pollsters, who collect voter opinions and may interact with voters during polling.\\

        Environment Description:

        The environment consists of voters and pollsters interacting within a political context where voters express preferences and pollsters gather data.\\

        State Variables Description:

        Policy preference represents each voter's preference for policies, which may change based on polling interaction; Polling data reflects the popularity of different candidates or policies; Trust level indicates the degree of trust voters have in pollsters when providing information.
    \end{textbox}

    $\bullet$ \textbf{Behavior graph}

    \begin{figure}[h]
        \centering
        \includegraphics[width=0.5\textwidth]{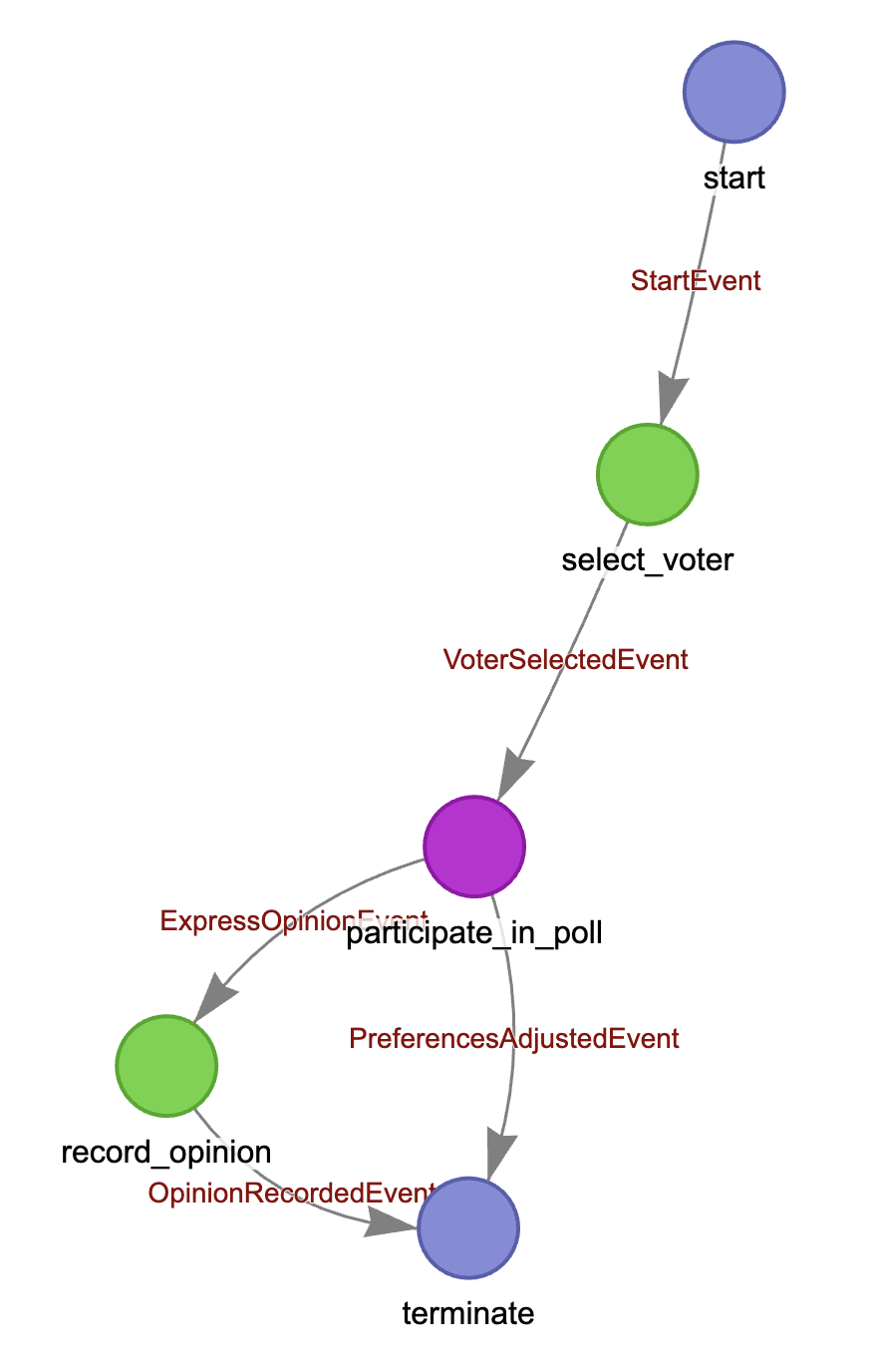}
        \caption{Behavior graph of Public Opinion Polling.}
    \end{figure}

    \subsubsection{Simple Policy Implementation Model}
    $\bullet$ \textbf{Detailed description}
    \begin{textbox}
     \underline{\textcolor{red}{Simple Policy Implementation Model}}\\
     
        System Goal:
      
        Simulate the execution process of government policies and study the fundamental mechanisms of policy implementation.\\

        Agent Types:

        Government, Citizens\\

        Environment Description:

        The environment consists of government entities executing policies and citizens reacting to these policies.

    \end{textbox}

    $\bullet$ \textbf{Behavior graph}
    \begin{figure}[h]
        \centering
        \includegraphics[width=\textwidth]{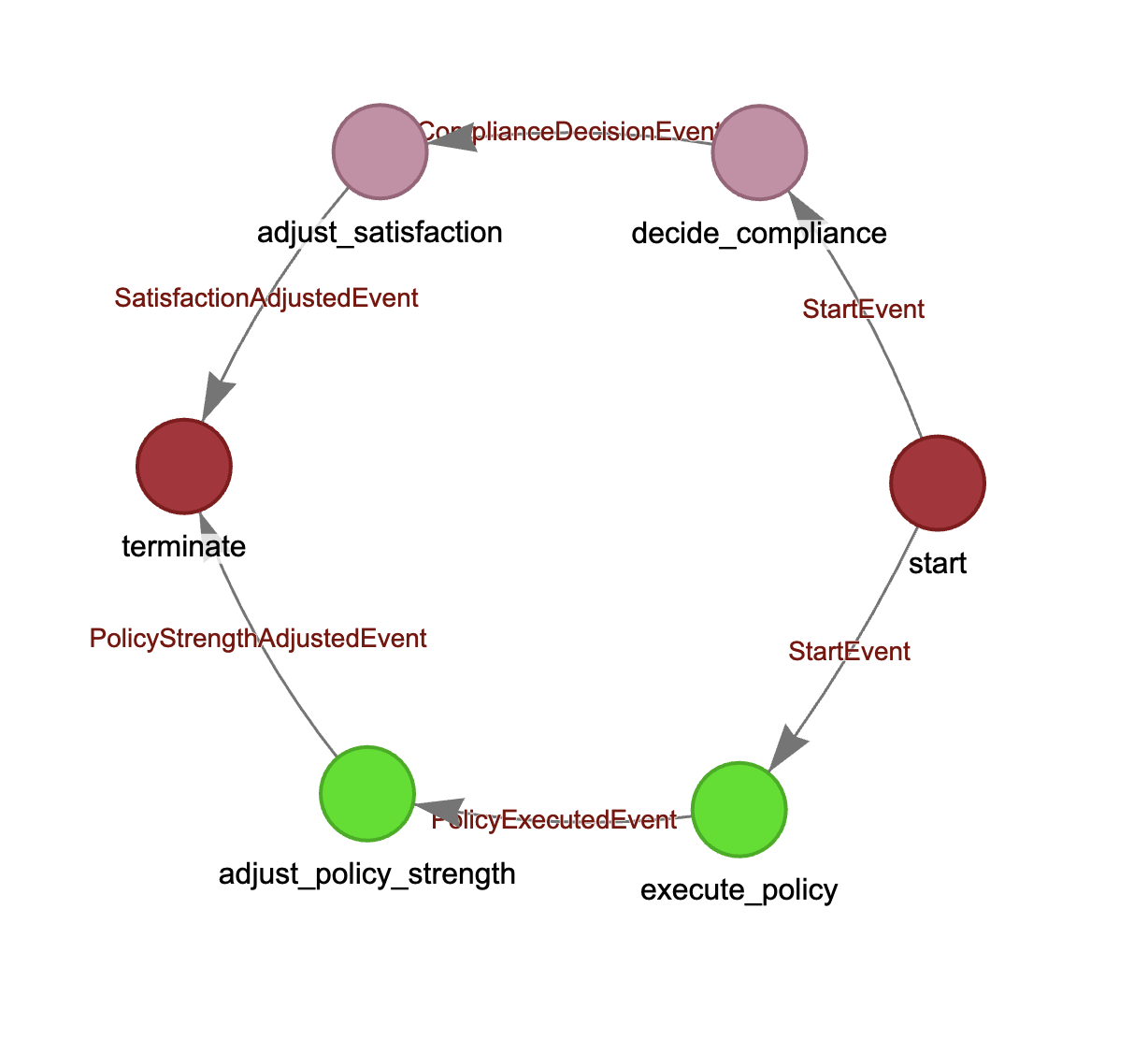}
        \caption{Behavior graph of Simple Policy Implementation Model.}
    \end{figure}

    \subsubsection{Voting}
    $\bullet$ \textbf{Detailed description}
    \begin{textbox}
     \underline{\textcolor{red}{Voting}}\\
     
        System Goal:
      
        Analyze how collective voting decisions produce results and how different voting rules affect fairness and efficiency in social choices.\\

        Agent Types:
        
        Voters who are agents deciding whether to vote and whom to vote for.\\

        Environment Description:

        A voting environment where voters have preferences for candidates or policies and decide their participation based on certain rules.

    \end{textbox}

    $\bullet$ \textbf{Behavior graph}
    \begin{figure}[h]
        \centering
        \includegraphics[width=\textwidth]{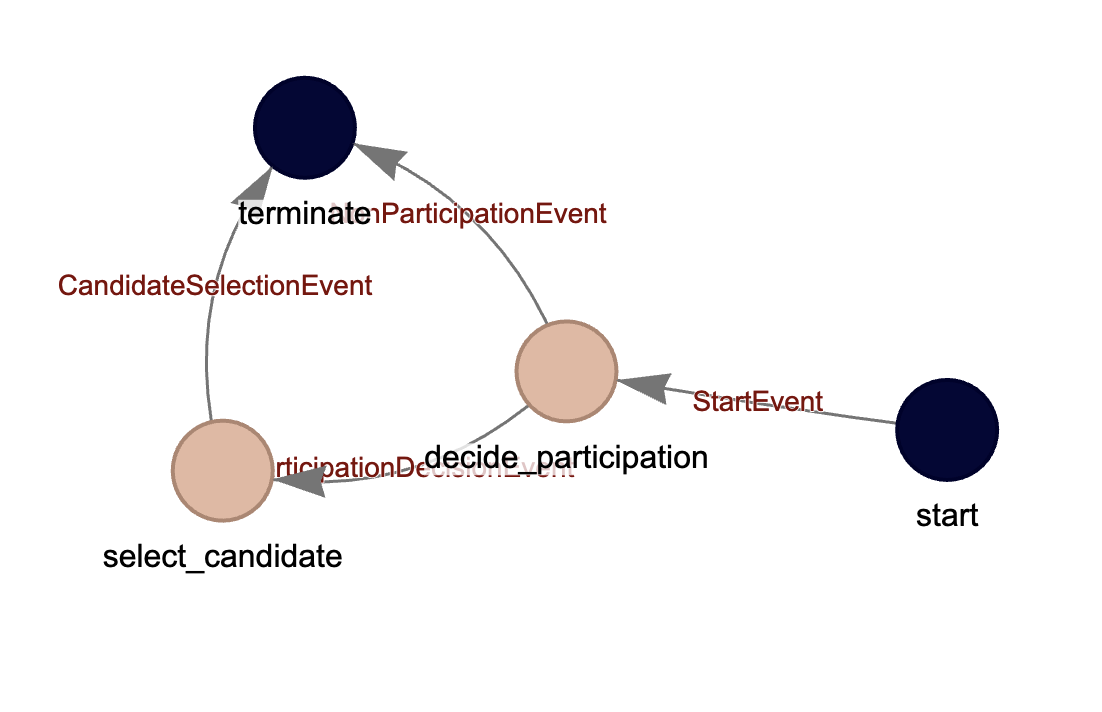}
        \caption{Behavior graph of Voting.}
    \end{figure}

    \subsubsection{Selective Exposure Theory}
    $\bullet$ \textbf{Detailed description}
    \begin{textbox}
     \underline{\textcolor{red}{Selective Exposure Theory}}\\
     
        System Goal:
      
        Simulate voter behavior in choosing specific news sources based on personal preferences to study how selective exposure leads to political polarization among groups, analyzing information bubbles and echo chamber effects.\\

        Agent Types:
        
        Voters who select media sources and adjust political preferences; Media entities providing political content with potential biases.\\

        Environment Description:

        The environment consists of voters and various media sources with different political biases, where voters can choose which media to consume.

    \end{textbox}

    $\bullet$ \textbf{Behavior graph}
    \begin{figure}[h]
        \centering
        \includegraphics[width=\textwidth]{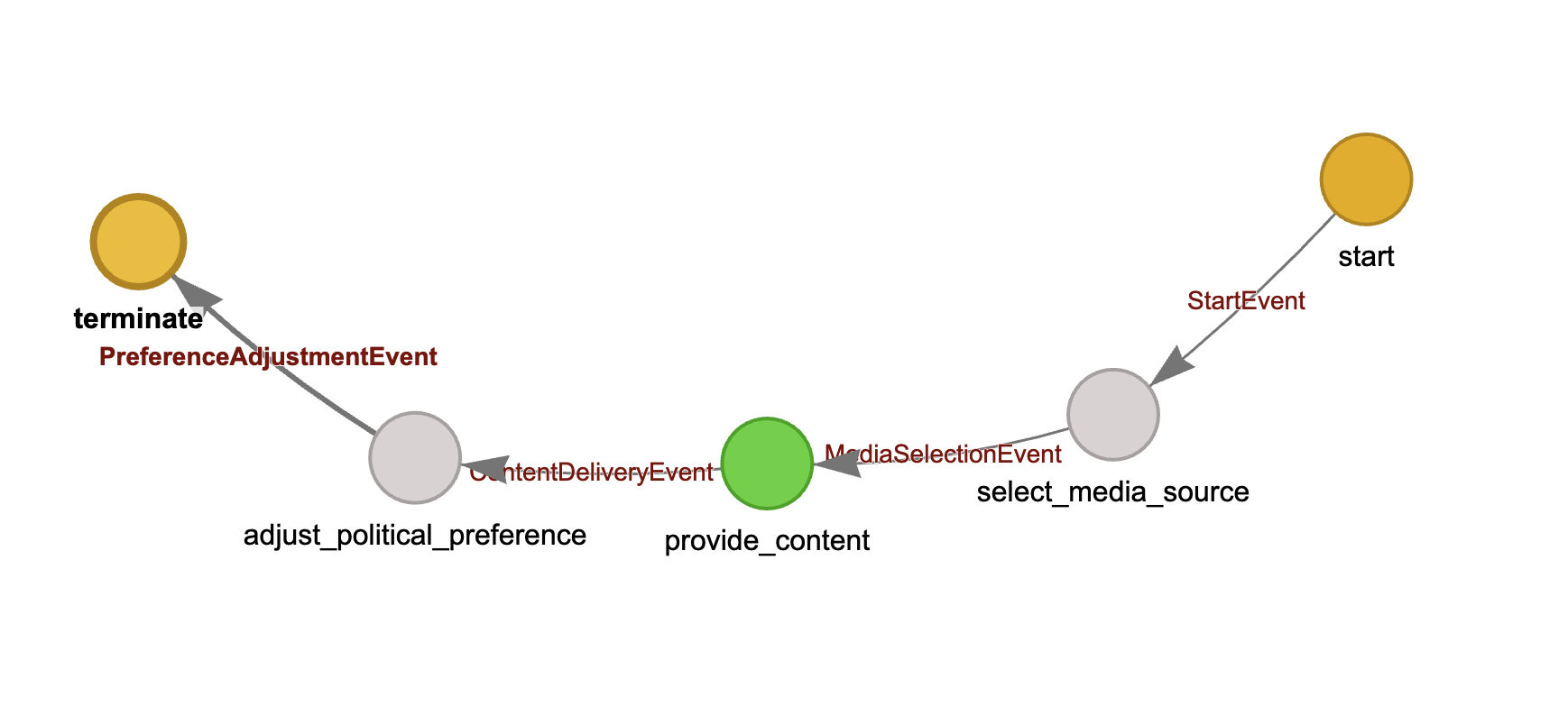}
        \caption{Behavior graph of Selective Exposure Theory.}
    \end{figure}

    \subsubsection{Rebellion}
    $\bullet$ \textbf{Detailed description}
    \begin{textbox}
     \underline{\textcolor{red}{Rebellion}}\\
     
        System Goal:
        
        The model simulates individual rebellion behavior against unjust regimes to study how different repression strategies affect social stability.\\

        Agent Types:
        
        Citizens who may participate in rebellion and the Government responsible for maintaining order and deciding repression strategies.\\

        Environment Description:
        
        A societal setting where citizens and government interact, with citizens evaluating their grievances and the government implementing repression.
      
    \end{textbox}

    $\bullet$ \textbf{Behavior graph}
    \begin{figure}[h]
        \centering
        \includegraphics[width=\textwidth]{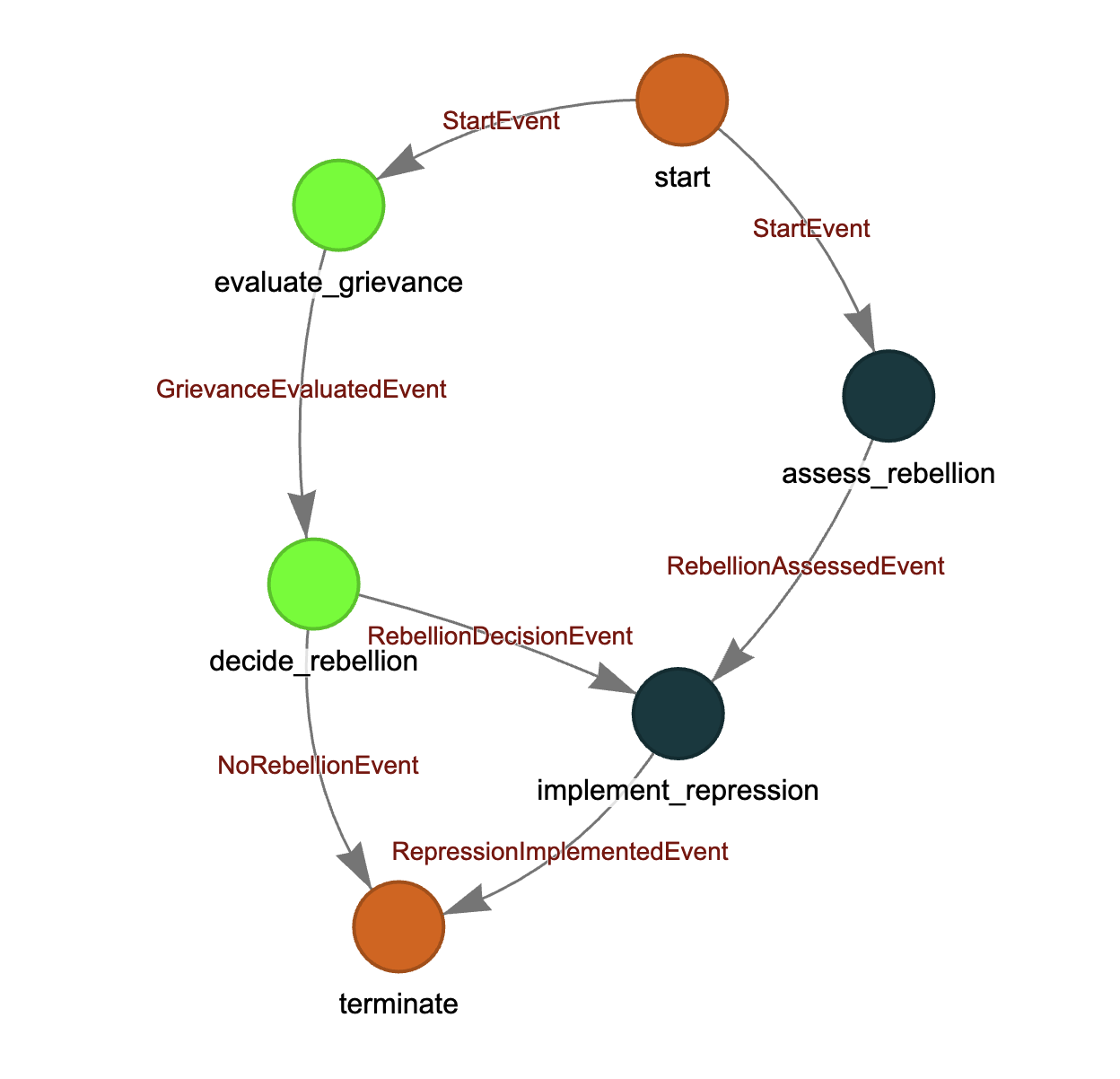}
        \caption{Behavior graph of Rebellion.}
    \end{figure}

    \subsubsection{Electoral Polarization System}
    $\bullet$ \textbf{Detailed description}
    \begin{textbox}
     \underline{\textcolor{red}{Electoral Polarization System}}\\
     
        System Goal:
        
        Simulate the dynamics of party competition, media influence, and voter behavior during elections to understand the impact of different electoral systems, media environments, and social network structures on political polarization and stability.\\

        Agent Types:
        
        1. Voter: Attributes include id, ideology, party\_identification, issue\_positions, issue\_importance, information\_level, media\_consumption, social\_network, voting\_history, susceptibility, economic\_status, demographic\_group. 2. Party: Attributes include id, ideology, platform, resources, strategy, base\_voters, target\_voters, message\_framing, leadership, history. 3. Media: Attributes include id, type, ideology\_bias, audience\_reach, credibility, framing\_strategy, agenda\_setting\_power, content\_selection\_bias, audience\_demographics. 4. Interest\_Group: Attributes include id, interests, resources, influence\_strategy, connections, public\_support, lobbying\_effectiveness.\\

        Environment Description:
        
        Includes electoral\_system, polarization\_index, economic\_conditions, salient\_issues, media\_environment\_diversity, and external\_shocks. The environment simulates the electoral system type, economic conditions, and external shocks impacting the political landscape.
      
    \end{textbox}

    $\bullet$ \textbf{Behavior graph}
    \begin{figure}[h]
        \centering
        \includegraphics[width=\textwidth]{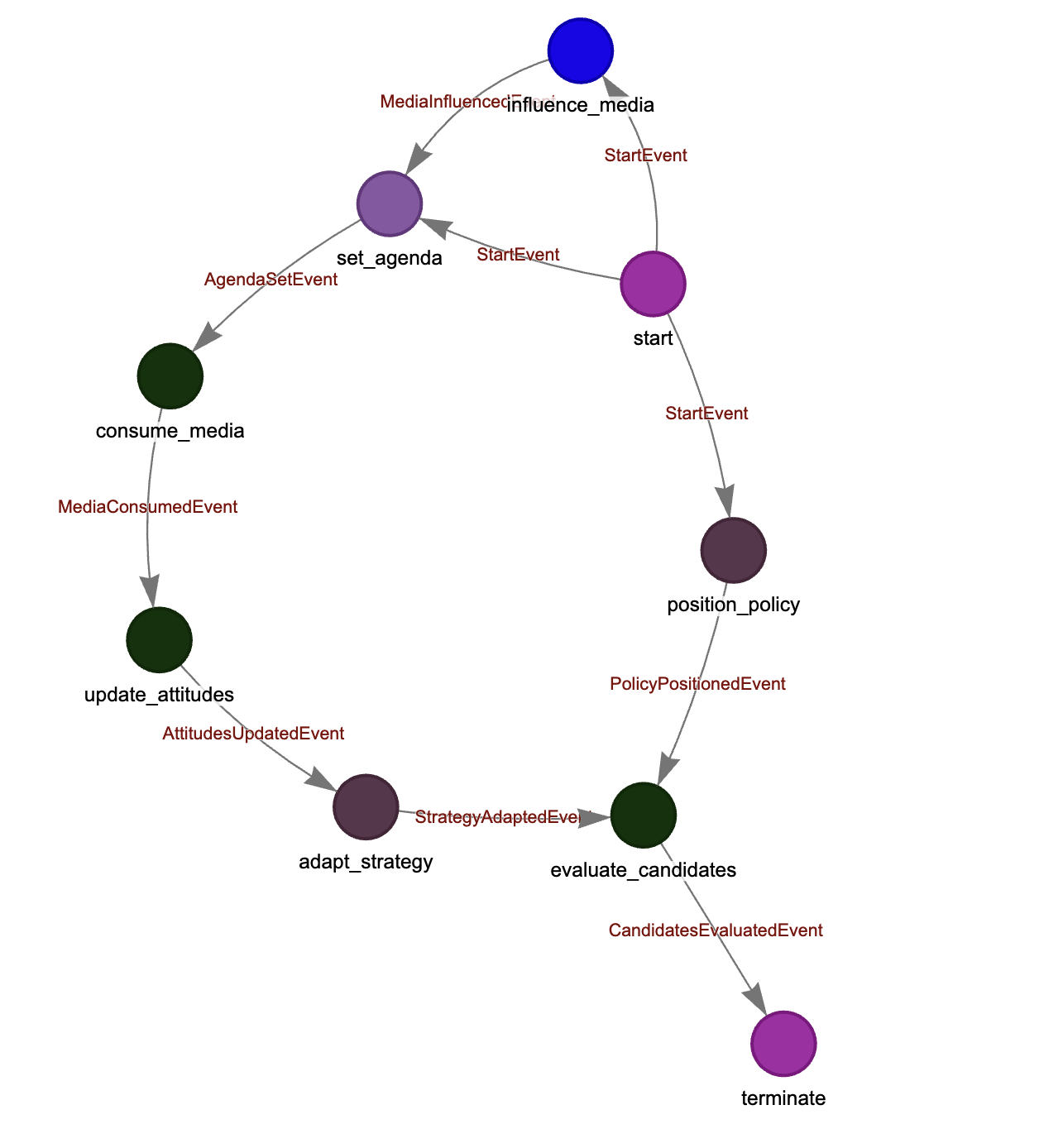}
        \caption{Behavior graph of Electoral Polarization System.}
    \end{figure}

    \subsection{Psychology}

    \subsubsection{Cognitive Dissonance Theory}

    $\bullet$ \textbf{Detailed description}

    \begin{textbox}

     \underline{\textcolor{red}{Cognitive Dissonance Theory}}\\
     
        System Goal:
        
        Simulate cognitive dissonance theory using three LLM agents to study the conflict between behavior and belief.\\

        Agent Types:

        Three agents: Actor A, Observer B, and Feedbacker C.\\

        Environment Description:

        A simulated environment where agents interact based on cognitive dissonance scenarios.\\

        Agent Roles:

        Actor A experiences cognitive dissonance, Observer B provides feedback, Feedbacker C evaluates responses.\\

        System Scope:

        Focuses on how agents manage cognitive dissonance through belief change, behavior change, or rationalization.
    \end{textbox}

    $\bullet$ \textbf{Behavior graph}

    \begin{figure}[h]
        \centering
        \includegraphics[width=\textwidth]{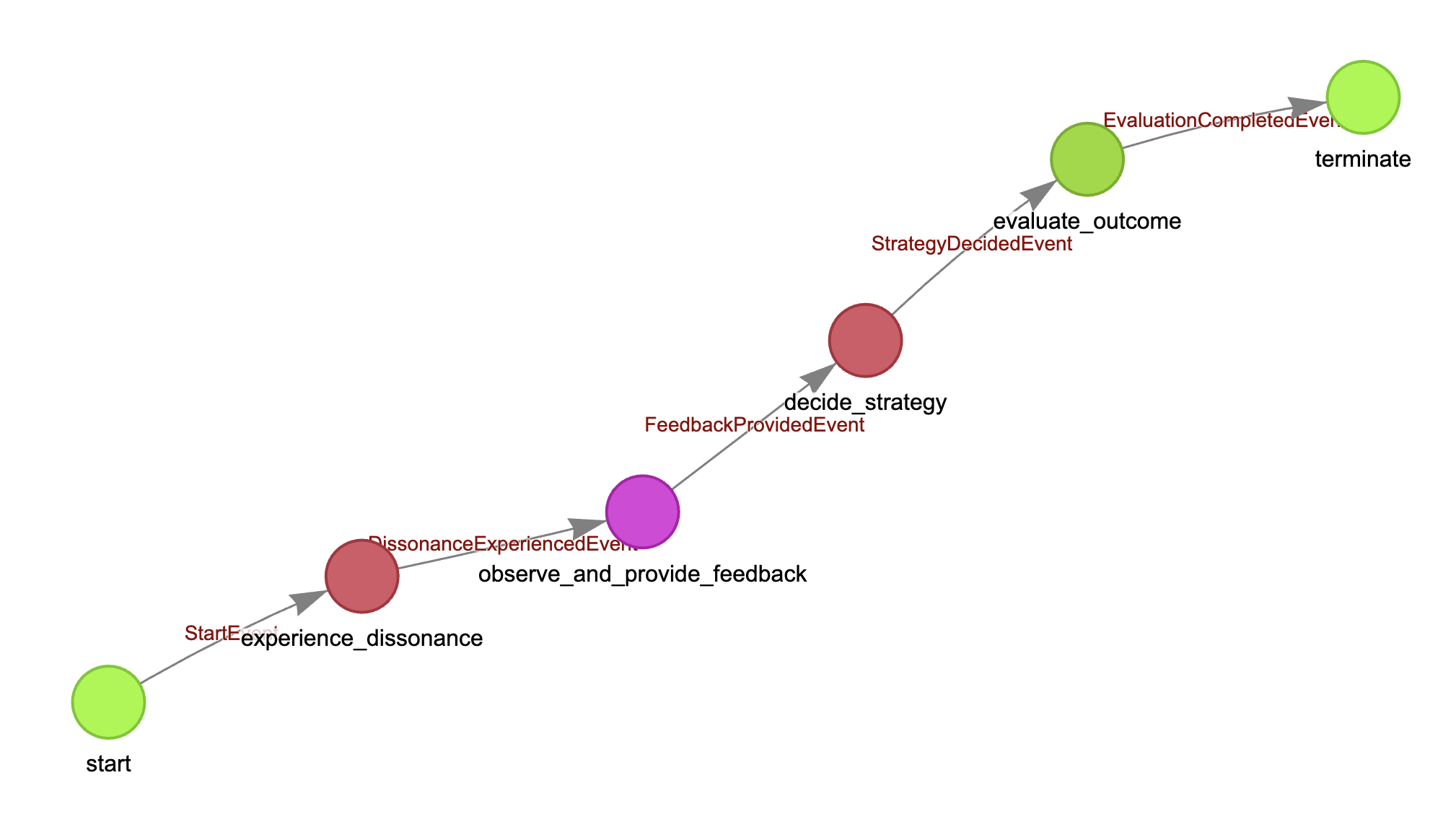}
        \caption{Behavior graph of Cognitive Dissonance Theory.}
    \end{figure}

    \subsubsection{Emotional Contagion Model}

    $\bullet$ \textbf{Detailed description}

    \begin{textbox}

     \underline{\textcolor{red}{Emotional Contagion Model}}\\

        System Goal:
        
        Simulate the process of emotional contagion to study how emotions spread among individuals and affect the overall emotional atmosphere of a group.\\

        Agent Types:

        Individual agents representing group members, each with an emotional state such as joy, sadness, or anger.\\

        Environment Description:

        A social network structure where agents interact through various forms of communication, influencing each other's emotional states.\\

        Emotional Dynamics:

        Focus on observing how emotions spread during social interactions and analyzing the dynamic process of emotional contagion.\\

        System Assumptions:

        Emotions can spread between individuals through social interactions, influencing behavior, speech, and decisions, thus affecting others' emotions.
    \end{textbox}

    $\bullet$ \textbf{Behavior graph}

    \begin{figure}[h]
        \centering
        \includegraphics[width=\textwidth]{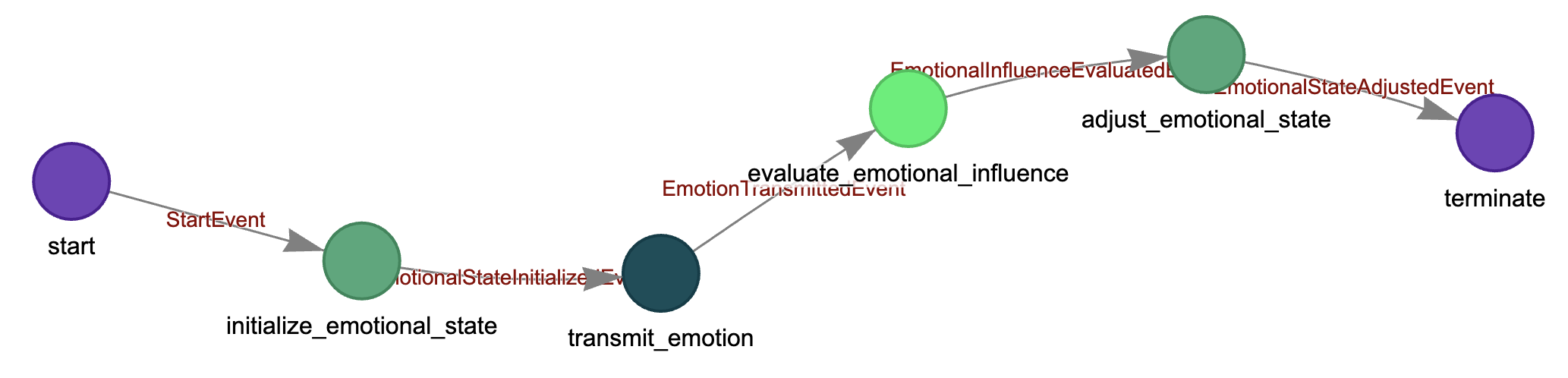}
        \caption{Behavior graph of Emotional Contagion Model.}
    \end{figure}

    \subsubsection{Conformity Behavior Model}
    $\bullet$ \textbf{Detailed description}
    \begin{textbox}
     \underline{\textcolor{red}{Conformity Behavior Model}}\\
     
        System Goal:
        
        Simulate conformity behavior and social influence processes in group environments, studying information cascades, collective decision biases, minority influence mechanisms, and how individual traits and environmental factors modulate the degree of conformity.\\
        
        Agent Types:
        
        Includes Individuals with attributes like id, conformity tendency, self-confidence, status, belief certainty, opinion visibility, decision history, social ties, openness, and threshold; Groups with attributes such as id, members, cohesion, norms, majority threshold, communication structure, homogeneity, and decision method; Opinion Leaders characterized by charisma, credibility, visibility, consistency, strategic thinking, and followers; Decision Contexts encompassing type, ambiguity, stakes, time pressure, feedback immediacy, and objective truth.\\

        Environment Description:
        
        Encompasses social pressure, information availability, uncertainty level, cultural setting, and anonymity level, providing a comprehensive backdrop for agent interactions and decision-making."

    \end{textbox}

    $\bullet$ \textbf{Behavior graph}
    \begin{figure}[h]
        \centering
        \includegraphics[width=\textwidth]{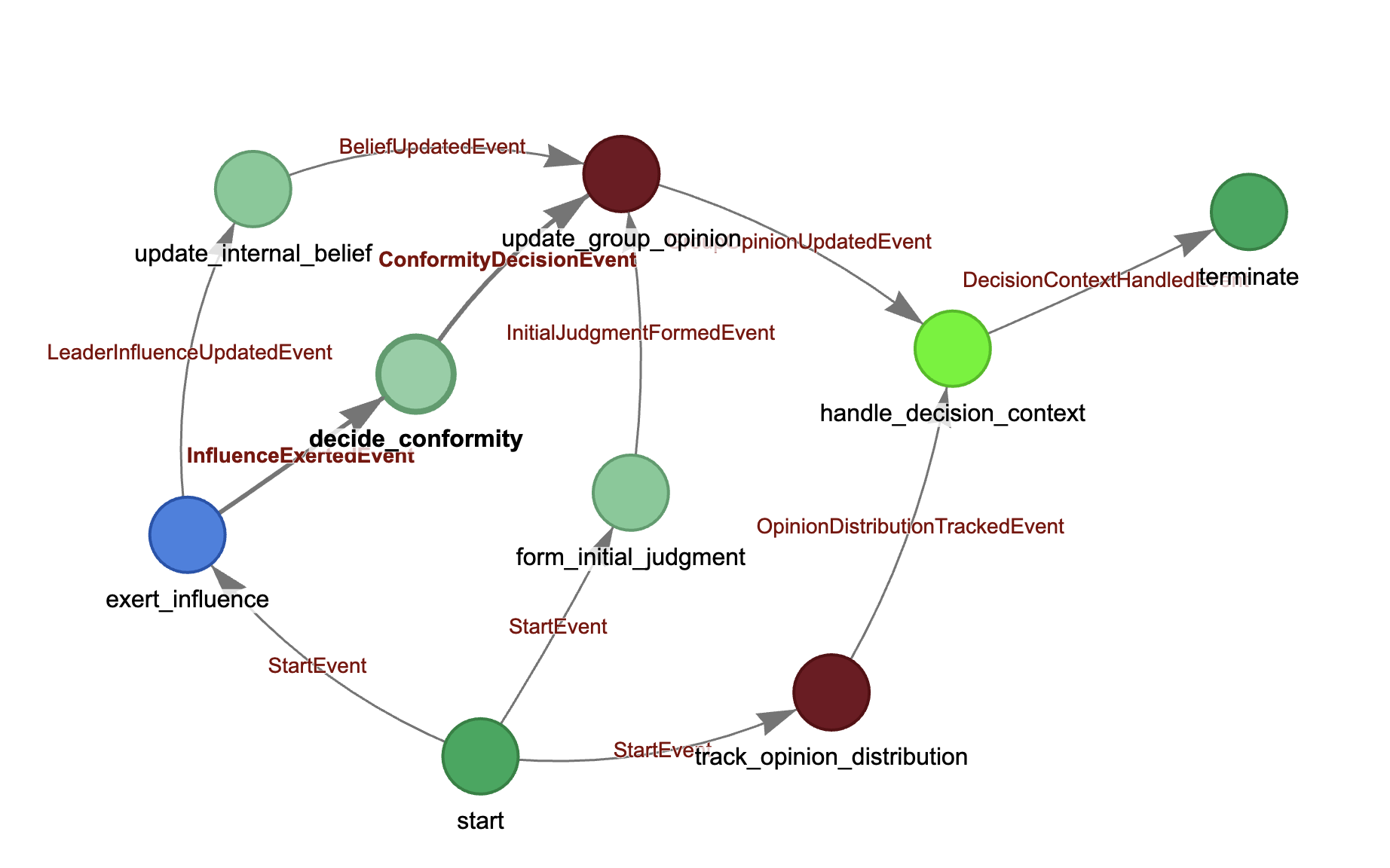}
        \caption{Behavior graph of Conformity Behavior Model.}
    \end{figure}

    \subsubsection{Antisocial Personality Theory}
    $\bullet$ \textbf{Detailed description}
    \begin{textbox}
     \underline{\textcolor{red}{Antisocial Personality Theory}}\\
     
        System Goal:
        
        Simulate how individuals with antisocial personality traits exhibit specific behavioral patterns in social interactions, such as lack of empathy, manipulation, and violation of social norms.\\
        
        Agent Types:
        
        Individuals exhibiting antisocial personality traits engaging in social interactions.\\

        Environment Description:
        
        A social environment where individuals interact, potentially leading to conflict, manipulation, or violent behavior.
      
    \end{textbox}

    $\bullet$ \textbf{Behavior graph}
    \begin{figure}[h]
        \centering
        \includegraphics[width=\textwidth]{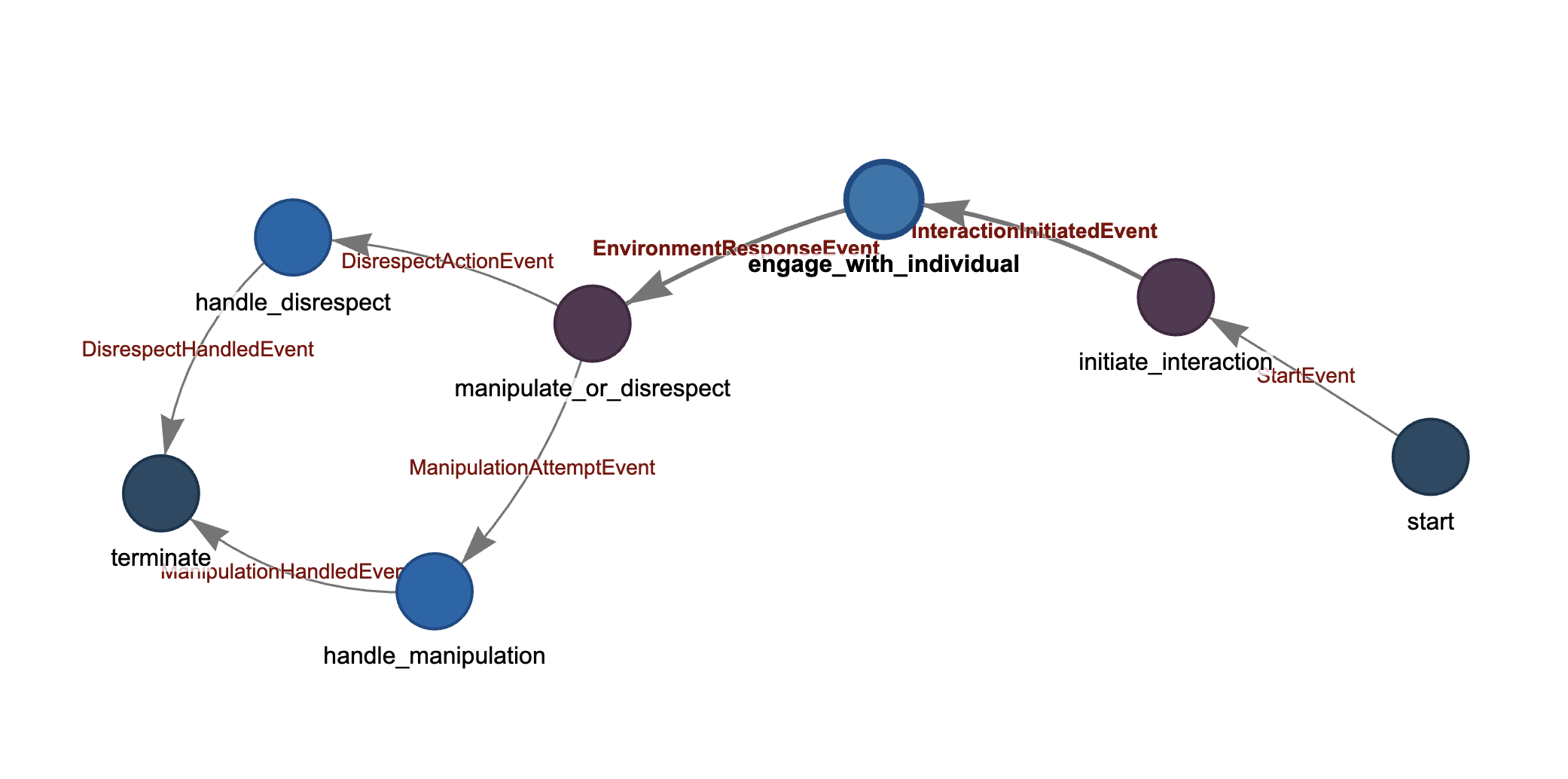}
        \caption{Behavior graph of Antisocial Personality Theory.}
    \end{figure}

    \subsubsection{Metacognition Theory}
    $\bullet$ \textbf{Detailed description}
    \begin{textbox}
     \underline{\textcolor{red}{Metacognition Theory}}\\
     
        System Goal:
        
        Simulate multiple LLM agents using metacognition to monitor and regulate their cognitive processes while executing tasks.\\
        
        Agent Types:
        
        Three agents: Task Executor, Monitor, Evaluator.\\

        Environment Description:
        
        The environment involves executing tasks that require reflection on cognitive processes, such as solving complex math problems or writing essays.
      
    \end{textbox}

    $\bullet$ \textbf{Behavior graph}
    \begin{figure}[h]
        \centering
        \includegraphics[width=\textwidth]{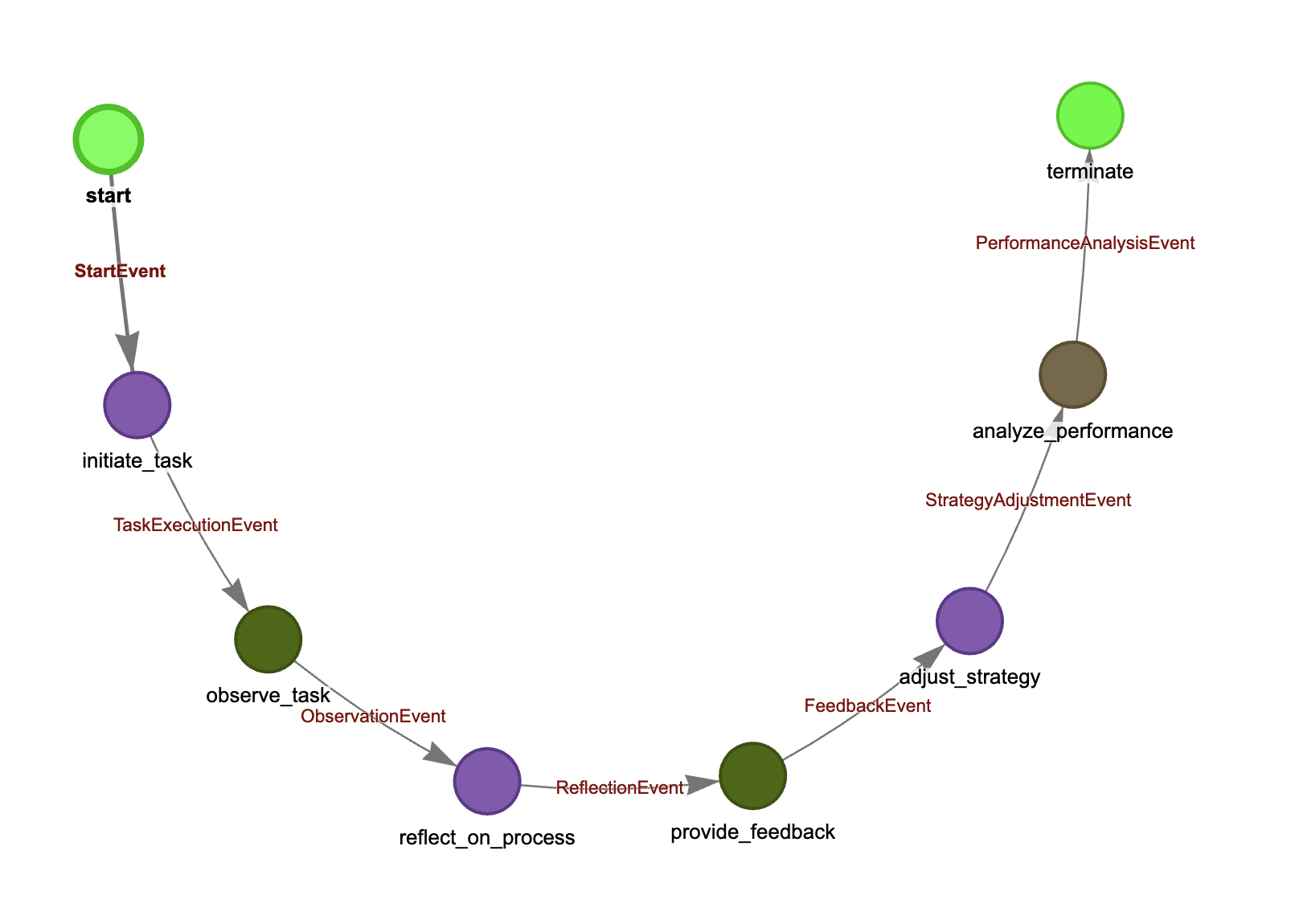}
        \caption{Behavior graph of Metacognition Theory.}
    \end{figure}

    \subsubsection{Attribution Theory}
    $\bullet$ \textbf{Detailed description}
    \begin{textbox}
     \underline{\textcolor{red}{Attribution Theory}}\\
     
        System Goal:
        
        Explore the application of attribution theory in social interactions through simulation of three LLM agents: Participant A, Participant B, and Feedbacker C.\\
        
        Agent Types:
        
        Three agents: Participant A, Participant B, Feedbacker C. Participant A and B engage in interactions; Feedbacker C analyzes attribution bias.\\

        Environment Description:
        
        The environment consists of interaction scenarios such as task collaboration and social dialogue where agents attribute behaviors to internal or external causes.
      
    \end{textbox}

    $\bullet$ \textbf{Behavior graph}
    \begin{figure}[h]
        \centering
        \includegraphics[width=\textwidth]{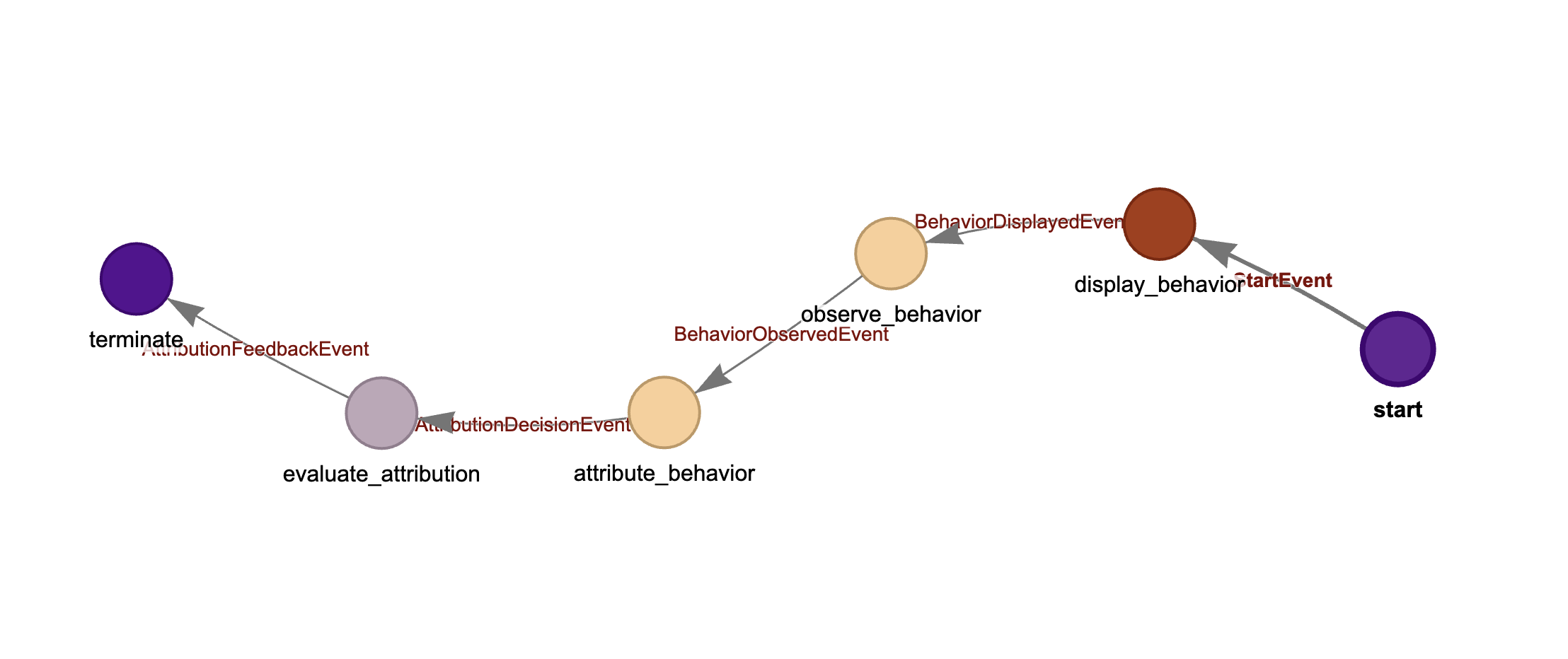}
        \caption{Behavior graph of Attribution Theory.}
    \end{figure}

    \subsection{Organization}

    \subsubsection{Scientific Management Theory}

    $\bullet$ \textbf{Detailed description}

    \begin{textbox}
     \underline{\textcolor{red}{Scientific Management Theory}}\\
     
        System Goal:
      
        The goal of the simulation is to explore Taylor's scientific management principles by simulating task allocation, standardized operations, and incentive measures to enhance work efficiency and productivity. The simulation includes interactions between workers and managers, analyzing how managers optimize workflows through scientific analysis and decision-making, and motivate employees for higher productivity.\\

        Agent Types:

        Agents represent employees or managers within an organization. Worker Agents execute specific tasks influenced by task allocation, standardized workflows, and incentive mechanisms. Manager Agents design workflows, allocate tasks, set incentives, and supervise workers to enhance efficiency.\\

        Environment Description:

        The environment models a workplace where interactions between workers and managers occur, focusing on task distribution, workflow standardization, and incentive implementation to improve productivity.
    \end{textbox}

    $\bullet$ \textbf{Behavior graph}

    \begin{figure}[h]
        \centering
        \includegraphics[width=0.7\textwidth]{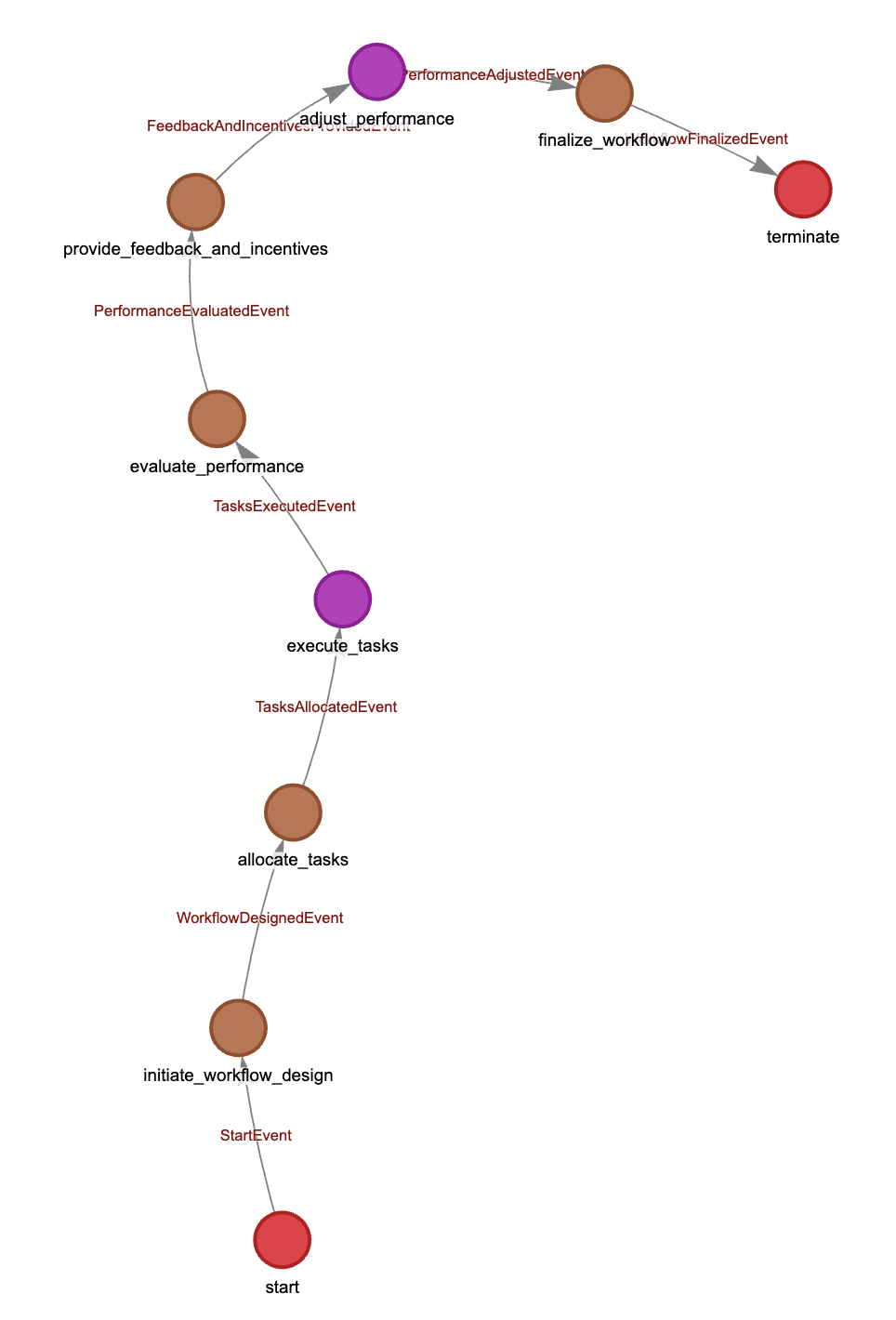}
        \caption{Behavior graph of Scientific Management Theory.}
    \end{figure}

    \subsubsection{Labor Market Matching Process}

    $\bullet$ \textbf{Detailed description}

    \begin{textbox}
     \underline{\textcolor{red}{Labor Market Matching Process}}\\
     
        System Goal:

        Simulate the job seeking and recruitment process in the labor market to study information asymmetry, signaling, matching efficiency, bias, and the impact of different recruitment strategies and job-seeking behaviors on market outcomes and fairness.\\

        Agent Types:

        Job\_Seeker: Attributes include id, skills, education, experience, productivity, salary\_expectation, job\_preferences, network, risk\_attitude, job\_search\_strategy, negotiation\_style, observable\_traits, reservation\_wage. 
        Employer: Attributes include id, industry, size, jobs, compensation\_structure, culture, hiring\_criteria, screening\_methods, bias, reputation, historical\_hires, turnover\_rate. 
        Job: Attributes include id, employer, required\_skills, education\_requirement, experience\_requirement, salary\_range, job\_description, actual\_job\_quality, location, employment\_type. 
        Recruitment\_Channel: Attributes include type, reach, cost, information\_richness, filtering\_mechanism, matching\_algorithm.\\

        Environment Description:

        Labor market tightness, economic conditions, regulatory environment, skill demand trends, technological change rate.
    \end{textbox}

    $\bullet$ \textbf{Behavior graph}

    \begin{figure}[h]
        \centering
        \includegraphics[width=0.9\textwidth]{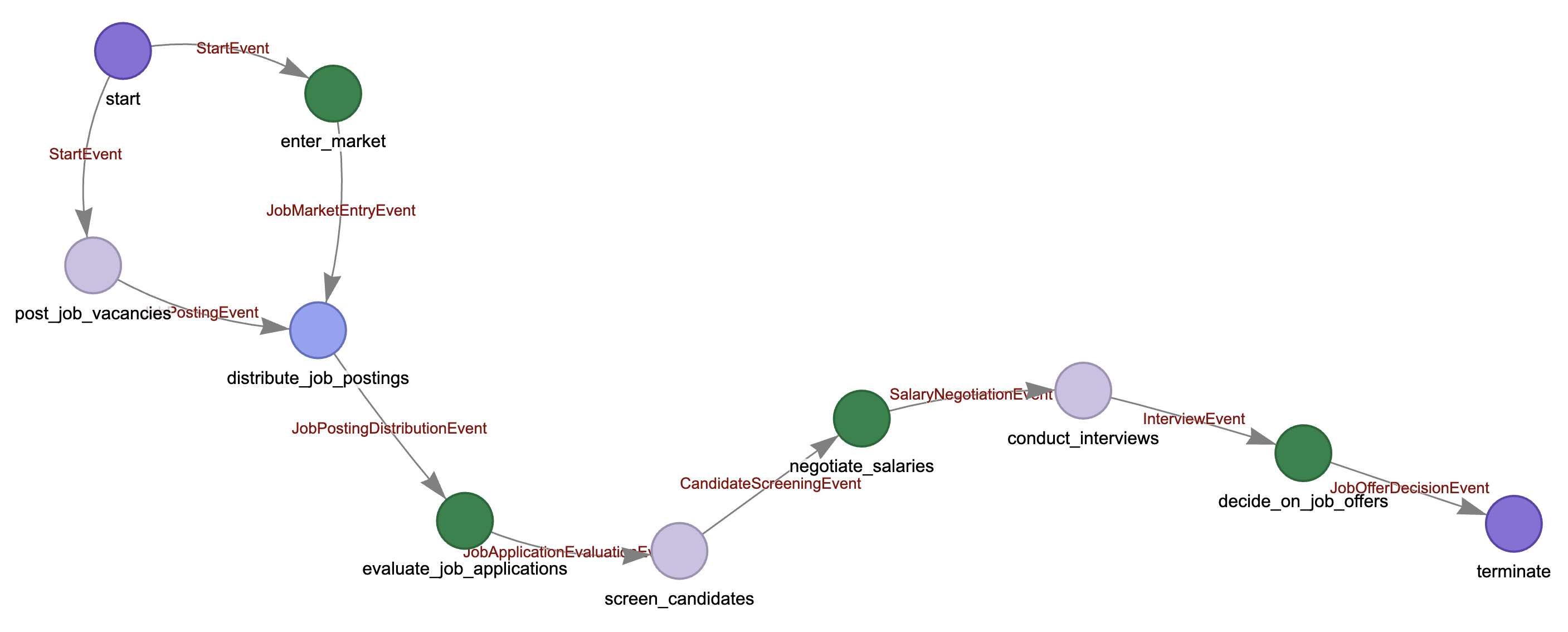}
        \caption{Behavior graph of Labor Market Matching Process.}
    \end{figure}

    \subsubsection{Hawthorne Studies}

    $\bullet$ \textbf{Detailed description}

    \begin{textbox}
     \underline{\textcolor{red}{Hawthorne Studies}}\\
     
        System Goal:
        
        Simulate the Hawthorne experiments using LLM to study how employee productivity, job satisfaction, and motivation change under different work environments and social interaction conditions. The simulation includes employee reactions to being observed, the impact of leader attention on productivity, and how team interactions affect overall performance. The goal is to understand the role of social factors in work performance through emotional responses and social interactions.\\

        Agent Types:

        Worker Agents: Represent employees whose behavior and performance are influenced by leadership attention, team atmosphere, and social interactions.
        
        Leader Agents: 
        Manage and guide employees, enhancing productivity through care and attention.
        
        Team Agents: Represent different team members, simulating how interaction, cooperation, and emotional support affect work performance.\\

        Environment Description:

        Physical Environment: Simulates various physical environment changes (e.g., lighting, work hours) from the Hawthorne experiments and analyzes how these changes interact with social factors like leadership attention. 
        
        Leadership Attention: Simulates how leaders motivate employees through attention, feedback, and communication to improve productivity. 
        
        Team Interaction: Simulates interaction and cooperation among team members to study how social relations and team atmosphere enhance overall work efficiency.
    \end{textbox}

    $\bullet$ \textbf{Behavior graph}

    \begin{figure}[h]
        \centering
        \includegraphics[width=0.8\textwidth]{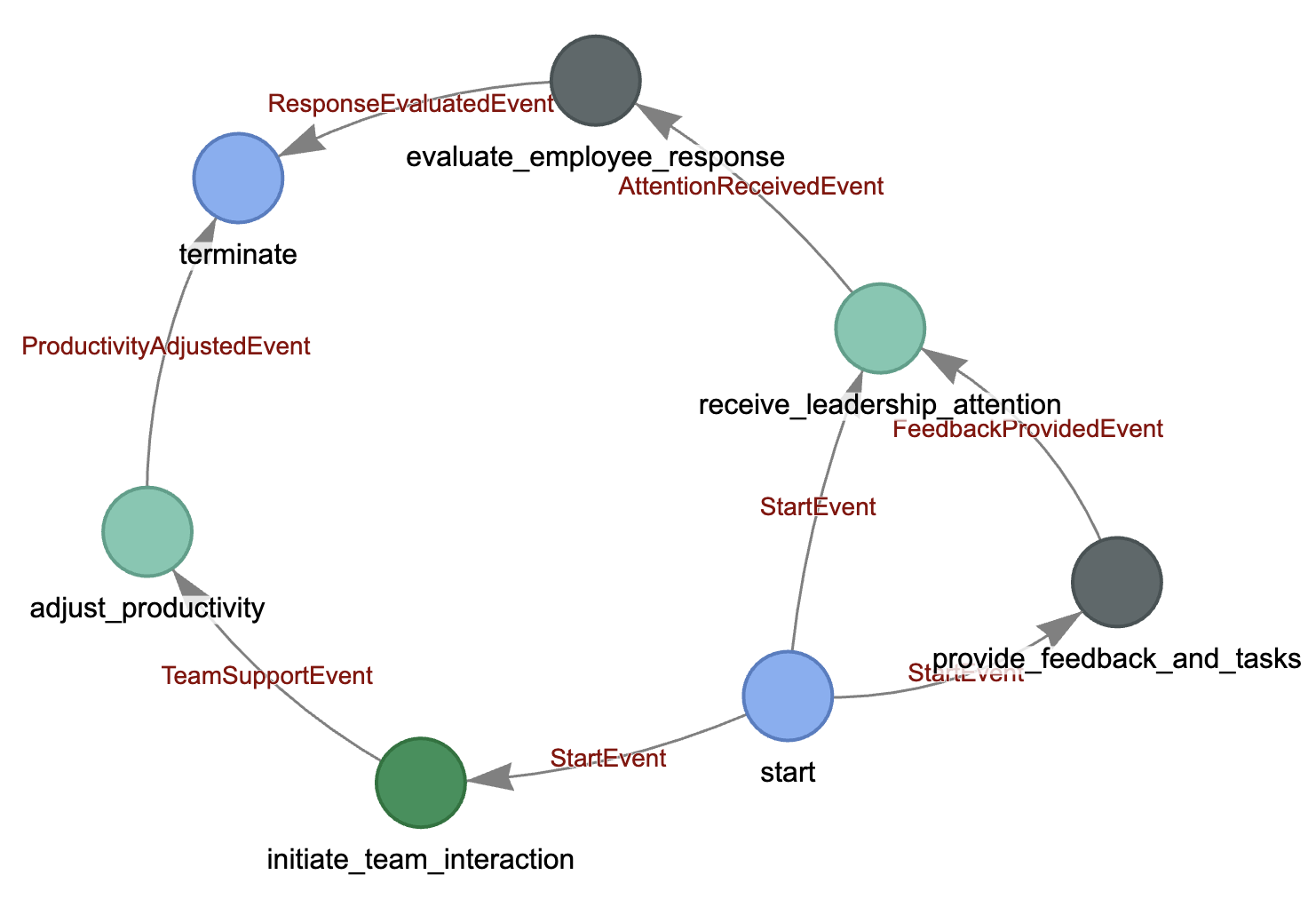}
        \caption{Behavior graph of Hawthorne Studies.}
    \end{figure}

    \subsubsection{Decision Theory}

    $\bullet$ \textbf{Detailed description}

    \begin{textbox}
     \underline{\textcolor{red}{Decision Theory}}\\
     
        System Goal:
        
        The simulation aims to model different decision-making processes within decision theory using LLMs, studying how choices are made in uncertain and complex environments. It analyzes decision-makers' behaviors when faced with limited information and multiple options, focusing on decision processes in the presence of risks, costs, utilities, and how to make optimal decisions based on goals and constraints.\\

        Agent Types:

        Decision Maker Agents representing individuals or teams making choices based on different criteria such as cost, utility, and risk. Environment Agents representing external factors affecting decisions, simulating uncertainty, risks, and resource constraints. Alternative Options Agents representing different choices faced by decision-makers, each with distinct costs, utilities, and risks.\\

        Environment Description:

        The environment includes elements of uncertainty, risk, and resource constraints that impact decision-making. It simulates various factors influencing decisions, such as external conditions, available information, and constraints that decision-makers must navigate.
    \end{textbox}

    $\bullet$ \textbf{Behavior graph}

    \begin{figure}[h]
        \centering
        \includegraphics[width=0.7\textwidth]{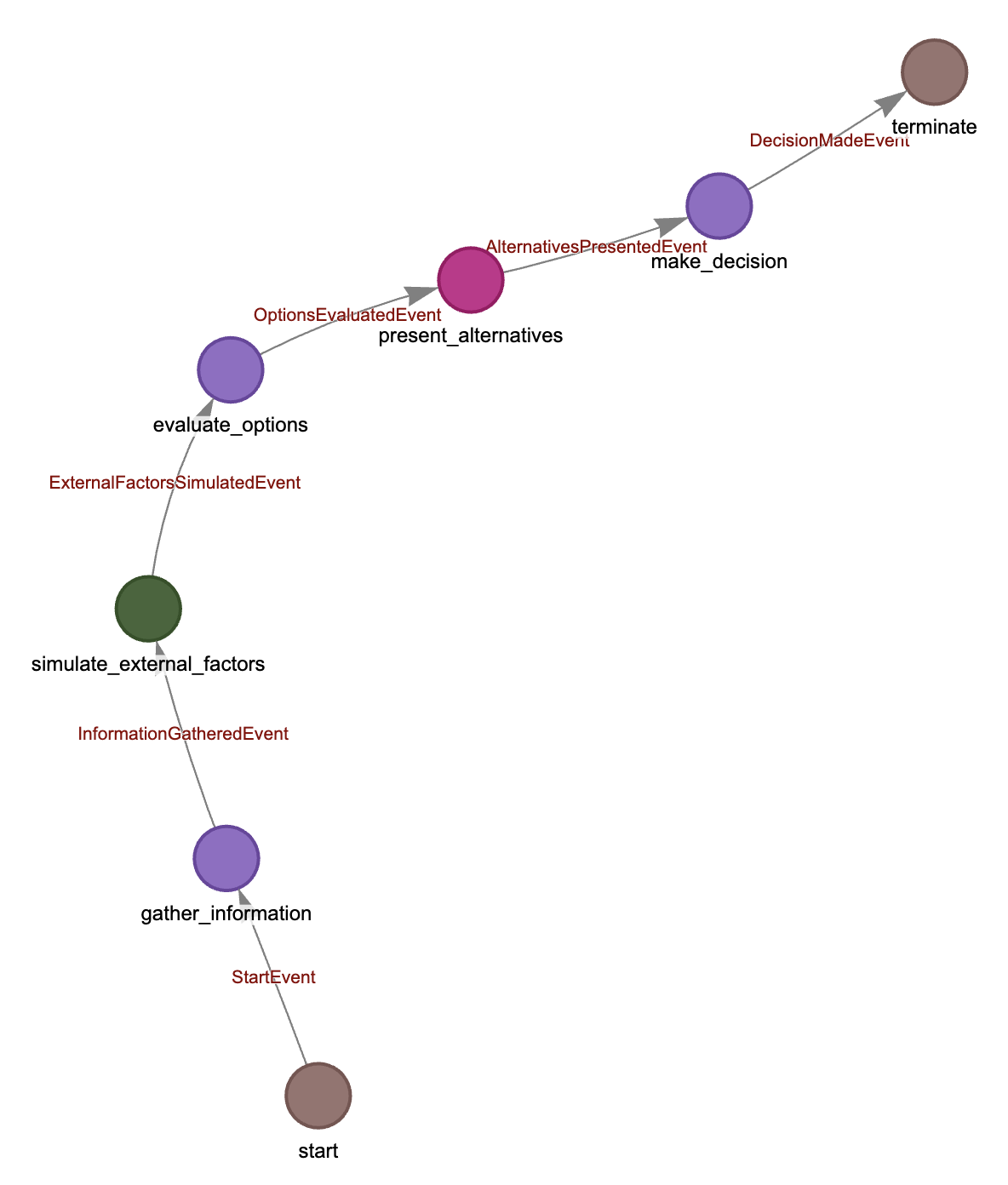}
        \caption{Behavior graph of Decision Theory.}
    \end{figure}

    \subsubsection{Organizational Change and Adaptation Theory}

    $\bullet$ \textbf{Detailed description}

    \begin{textbox}
     \underline{\textcolor{red}{Organizational Change and Adaptation Theory}}\\
     
        System Goal:
        
        The goal of this simulation is to use large language models (LLMs) to simulate how organizations undergo change and adaptation in response to internal and external shifts. The simulation explores how employees, leaders, and managers within an organization respond to change, focusing on communication, feedback, and adjustment mechanisms during the change process. By simulating adaptation to various types of change, the LLM helps us understand how change implementation affects employee behavior, work performance, and organizational culture transformation.\\

        Agent Types:

        Employee Agents: Represent employees within the organization, showing varied responses to change such as support, resistance, or adaptation. Their behavior and emotional responses are influenced by the type of change and leadership style. Leader Agents: Guide the change process, making crucial decisions, communicating, and providing feedback. They help employees adapt through encouragement, motivation, and coordination. Manager Agents: Oversee the specific execution of changes, ensuring the process proceeds as planned and assisting employees in overcoming resistance.\\

        Environment Description:
        
        The environment includes a simulated organizational setting where different types of changes occur, such as incremental improvements, rapid adjustments due to market shifts, and strategic realignments. The environment is dynamic, with interactions among various agents representing different organizational roles and levels.
    \end{textbox}

    $\bullet$ \textbf{Behavior graph}

    \begin{figure}[h]
        \centering
        \includegraphics[width=0.9\textwidth]{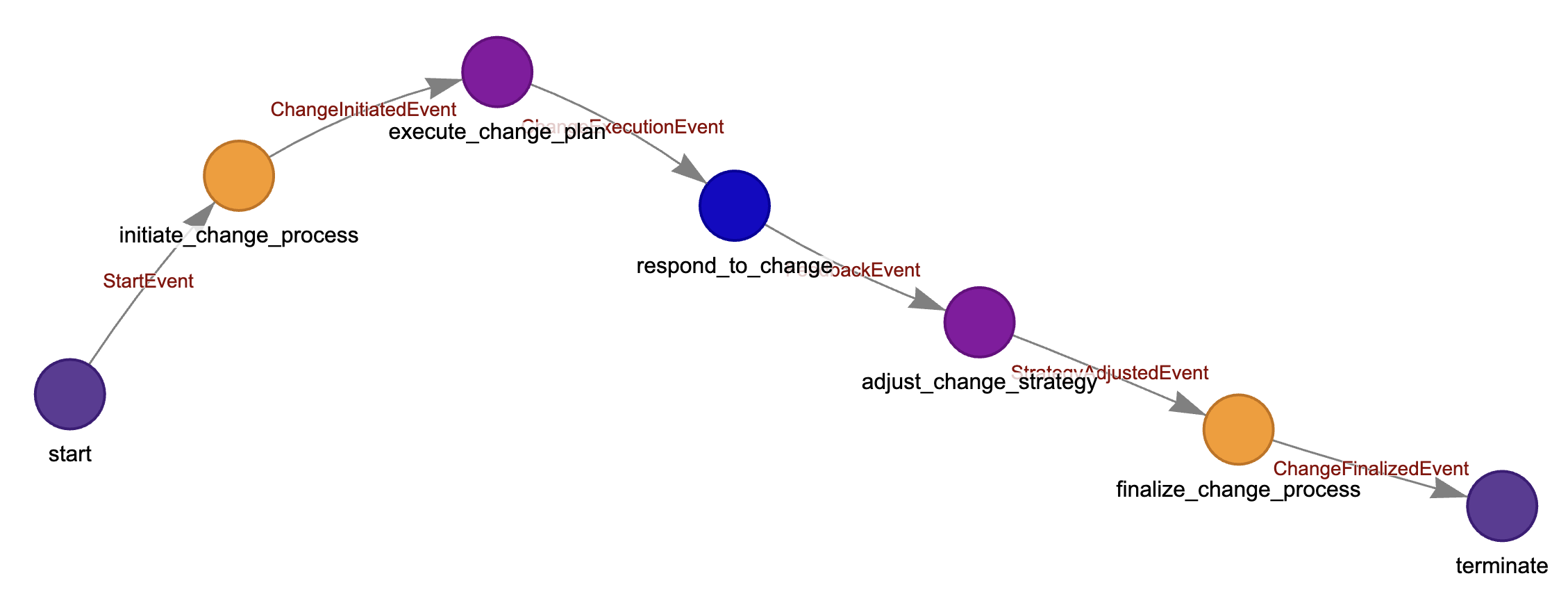}
        \caption{Behavior graph of Organizational Change and Adaptation Theory.}
    \end{figure}

    \subsubsection{Hierarchy of Needs}

    $\bullet$ \textbf{Detailed description}

    \begin{textbox}
     \underline{\textcolor{red}{Hierarchy of Needs}}\\
     
        System Goal:
        
        To use a Large Language Model (LLM) to simulate how an individual progresses from basic physiological needs to self-actualization needs within an organizational or team environment. The simulation explores how satisfying different levels of needs influences individual behavior, motivation, and decision-making through interactions with other individuals or systems. Scenarios include virtual companies or social environments where individuals interact based on their need levels, affecting work performance, collaboration, and personal growth.\\

        Agent Types:
        
        Agents are individuals with varying levels of needs, ranging from physiological to self-actualization. Each agent can exhibit focus or motivation toward a particular level of need at different times.\\

        Environment Description:

        The environment includes organizational culture, work pressure, social networks, leadership styles, and other factors that impact the process of need satisfaction for agents.
    \end{textbox}

    $\bullet$ \textbf{Behavior graph}

    \begin{figure}[h]
        \centering
        \includegraphics[width=0.8\textwidth]{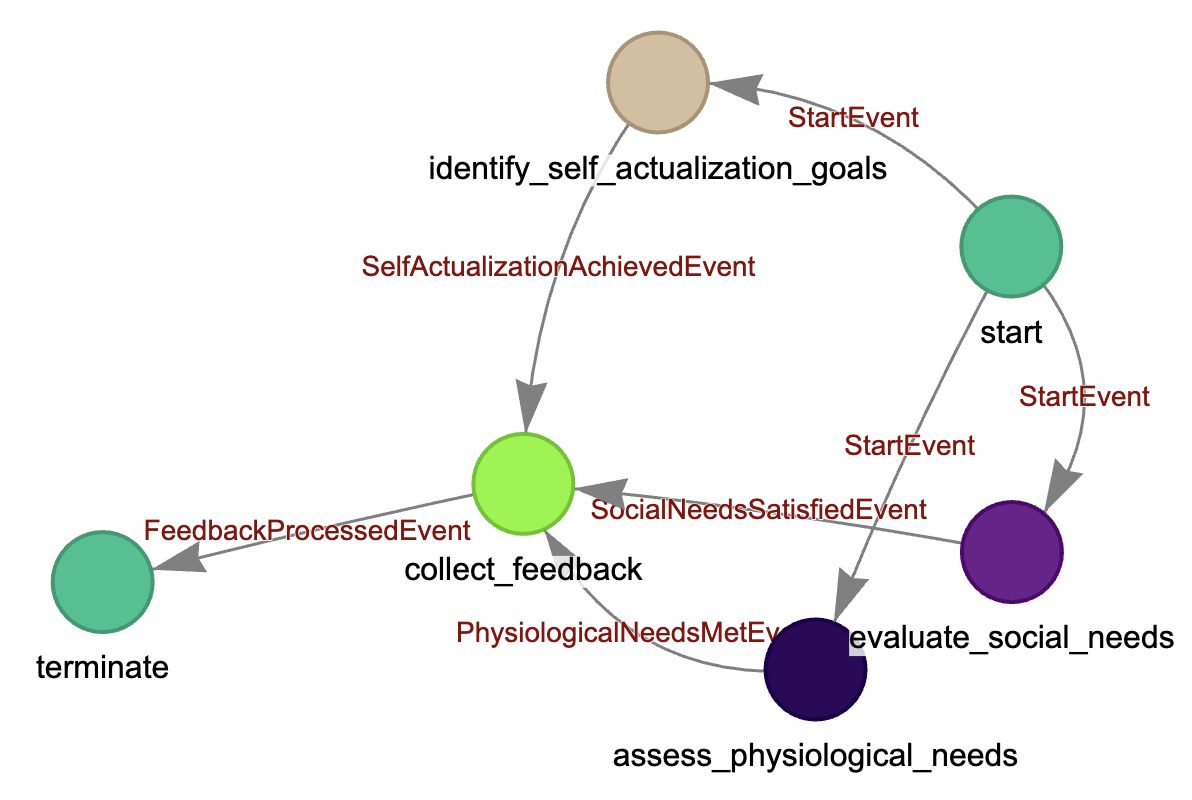}
        \caption{Behavior graph of Hierarchy of Needs.}
    \end{figure}

    \subsection{Demographics}

    \subsubsection{Epidemic Transmission Network}

    $\bullet$ \textbf{Detailed description}

    \begin{textbox}
     \underline{\textcolor{red}{Epidemic Transmission Network}}\\
     
        System Goal:

        Simulate the spread of disease within a population to study how individual behavior, demographic characteristics, interventions, and information dissemination affect epidemic dynamics, as well as evaluate the effectiveness of healthcare resource allocation and public health policies.\\

        Agent Types:

        Individuals with attributes such as id, age, health\_status, infection\_time, symptoms\_severity, vaccination\_status, risk\_perception, compliance\_tendency, mobility\_pattern, social\_contacts, demographic\_group, information\_sources, trust\_in\_authorities; 
        Households with attributes like id, members, location, size, quarantine\_status, infection\_risk; 
        Public Health Authorities with attributes including testing\_capacity, contact\_tracing\_efficiency, policy\_options, communication\_strategy, credibility, resource\_allocation; 
        Healthcare Facilities with attributes such as location, capacity, current\_occupancy, staff\_level, treatment\_effectiveness, resource\_level.\\

        Environment Description:

        Includes disease\_parameters like basic reproduction number R0, incubation period, infectious period; 
        information\_environment with accuracy and coverage; 
        economic\_pressure affecting policy implementation and individual behavior; 
        vaccination\_availability; and population\_density distribution.
    \end{textbox}

    $\bullet$ \textbf{Behavior graph}

    \begin{figure}[h]
        \centering
        \includegraphics[width=0.7\textwidth]{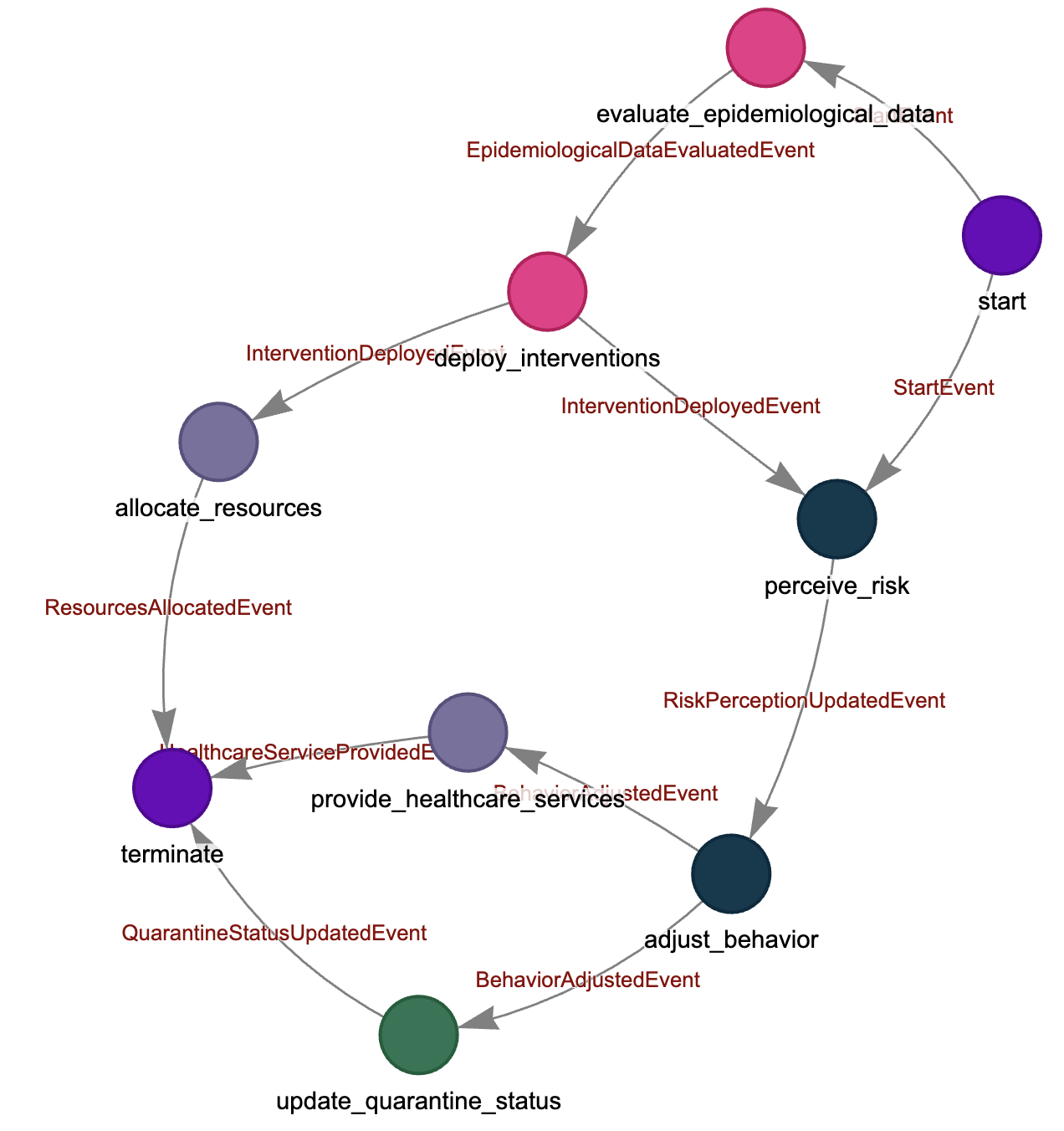}
        \caption{Behavior graph of Epidemic Transmission Network.}
    \end{figure}

    \subsubsection{Health Inequality}

    $\bullet$ \textbf{Detailed description}

    \begin{textbox}
     \underline{\textcolor{red}{Health Inequality}}\\
     
        System Goal:
        
        Simulate and understand the causes and mechanisms behind health inequalities by modeling complex interactions between individuals, families, communities, and societal institutions. Use simulations to predict how health policies, education, and healthcare services impact health inequalities among different groups such as low-income populations, minority groups, rural and urban residents. Provide effective policy recommendations to reduce health inequalities by analyzing the combined effects of social determinants.\\

        Agent Types:

        The agents in the model represent different individuals and groups in society, making health-related decisions based on socioeconomic conditions, education levels, and health behaviors. Individual agents simulate people of different genders, ages, socioeconomic backgrounds, cultures, and ethnicities, with their health influenced by living conditions, education, income, and healthcare resource availability. Family agents play a crucial role in health decisions, affecting family members' health through decisions on nutrition, education, and health behaviors. Community agents, as basic social structure units, influence residents' health levels through social capital, public resources, and medical facilities. Government agents determine the distribution of health resources and opportunities through health policies, medical security systems, and educational policies. Healthcare system agents simulate how healthcare systems in different regions affect individual health, including service accessibility, quality, and insurance.\\

        Environment Description:

        The environment includes various socioeconomic conditions and healthcare systems that influence health inequalities. It involves different groups such as low-income populations, minority groups, rural and urban residents, and factors like income level, education level, social capital, health behaviors, and healthcare accessibility. The simulation environment evolves through stages: initial state with significant health inequalities, social intervention phase with policy introductions, and long-term effects phase to evaluate the impact of policies on health inequalities.
    \end{textbox}

    $\bullet$ \textbf{Behavior graph}

    \begin{figure}[h]
        \centering
        \includegraphics[width=\textwidth]{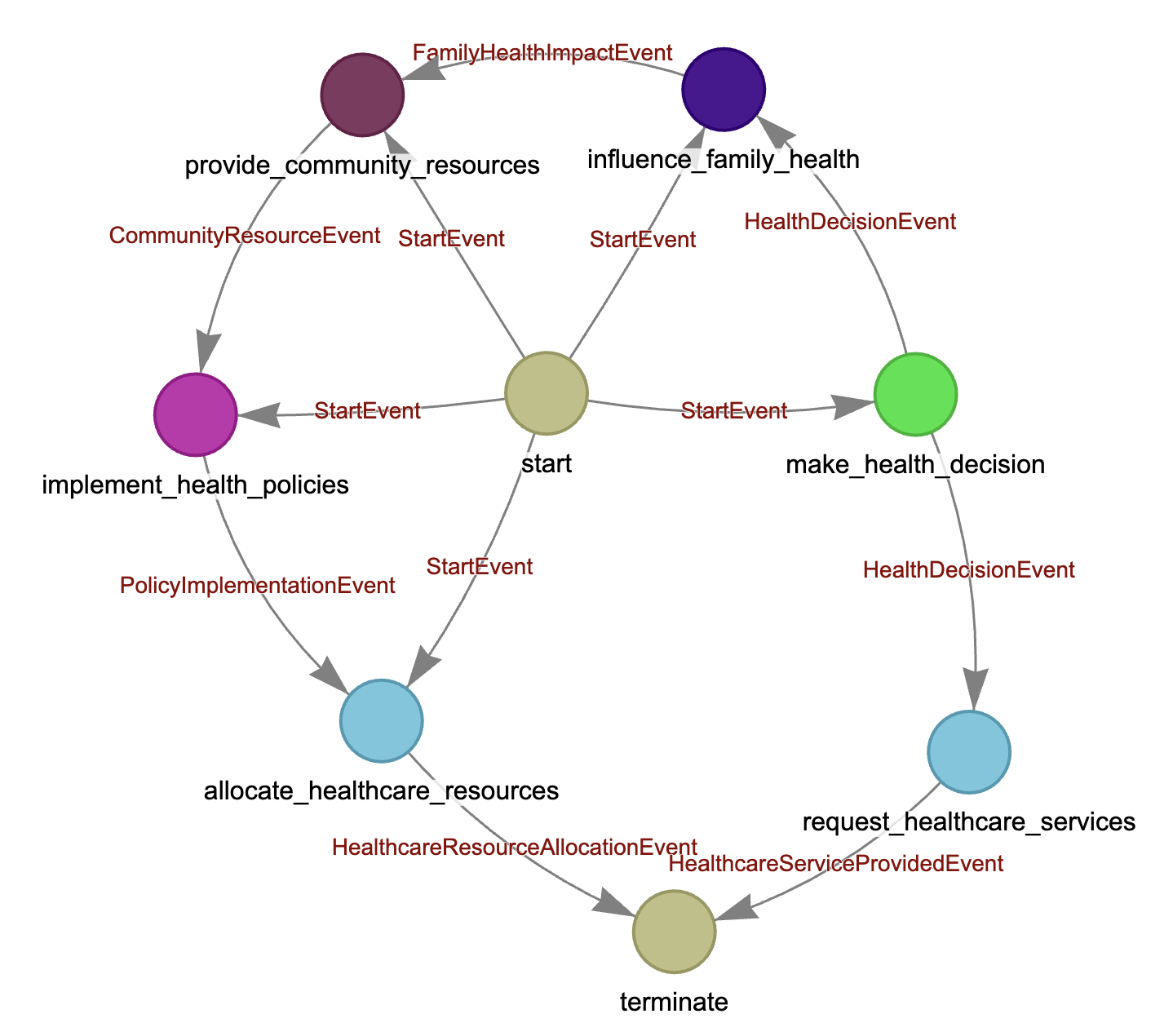}
        \caption{Behavior graph of Health Inequality.}
    \end{figure}

    \subsubsection{SIR Model}

    $\bullet$ \textbf{Detailed description}

    \begin{textbox}
     \underline{\textcolor{red}{SIR Model}}\\
     
        System Goal:
        
        To extend the SIR model using large language models (LLM) to simulate complex disease transmission processes, incorporating social factors such as government interventions, public health policies, and social network structures to better describe disease spread within populations.\\

        Agent Types:

        Individual agents representing people with health states of susceptible (S), infected (I), or recovered (R); group agents representing social groups like families or communities; government agents implementing policies; healthcare system agents managing treatment and recovery.\\

        Environment Description:

        The environment includes social interactions within and between groups, government policy impacts, healthcare infrastructure, and public health behaviors affecting disease transmission.\\

        Policy Evaluation:

        Researchers can evaluate the impact of different preventive policies like isolation, vaccination, and social distancing on disease spread and explore strategies for epidemic control in dynamic social environments.
    \end{textbox}

    $\bullet$ \textbf{Behavior graph}

    \begin{figure}[h]
        \centering
        \includegraphics[width=\textwidth]{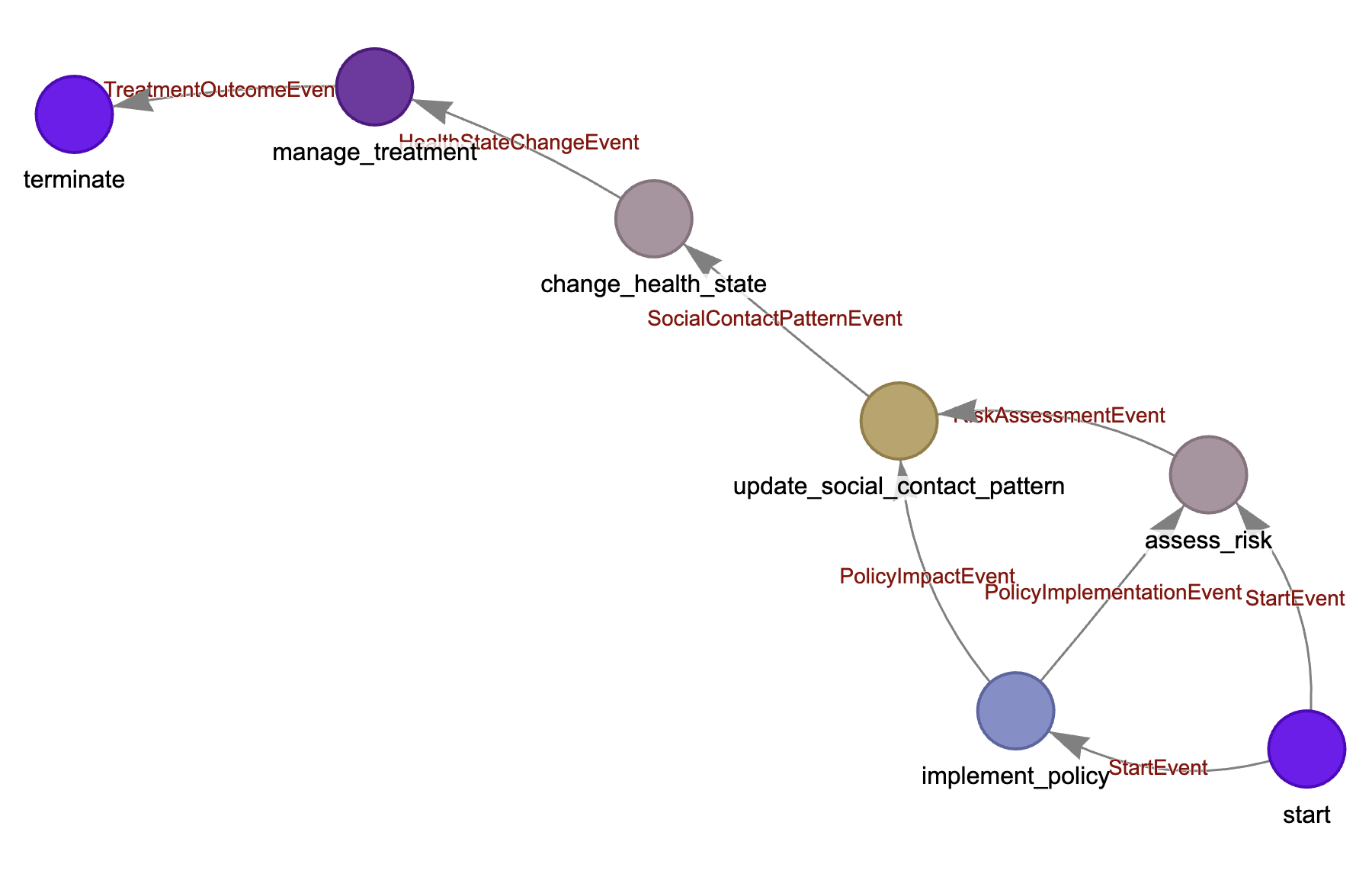}
        \caption{Behavior graph of SIR Model.}
    \end{figure}

    \subsubsection{Life Course Theory}

    $\bullet$ \textbf{Detailed description}

    \begin{textbox}
     \underline{\textcolor{red}{Life Course Theory}}\\
     
        System Goal:
        
        Simulate individual behaviors and decisions across different life stages to analyze the impact of various social, economic, and environmental factors on health and social behavior.\\

        Agent Types:

        Individual agents representing different life stages, family agents, education system agents, government agents, healthcare system agents.\\

        Environment Description:

        The environment includes socioeconomic conditions, educational systems, healthcare systems, and government policies affecting individuals throughout their life course.

        Life Course Impact:

        Explore how early education, nutrition, family structure, and other factors influence health, educational achievements, social adaptability, and adult behavior and health.

        Comprehensive Evaluation:
        
        Assess the impact of early life experiences on later development by integrating health, education, social support, and career choices.
    \end{textbox}

    $\bullet$ \textbf{Behavior graph}

    \begin{figure}[h]
        \centering
        \includegraphics[width=0.8\textwidth]{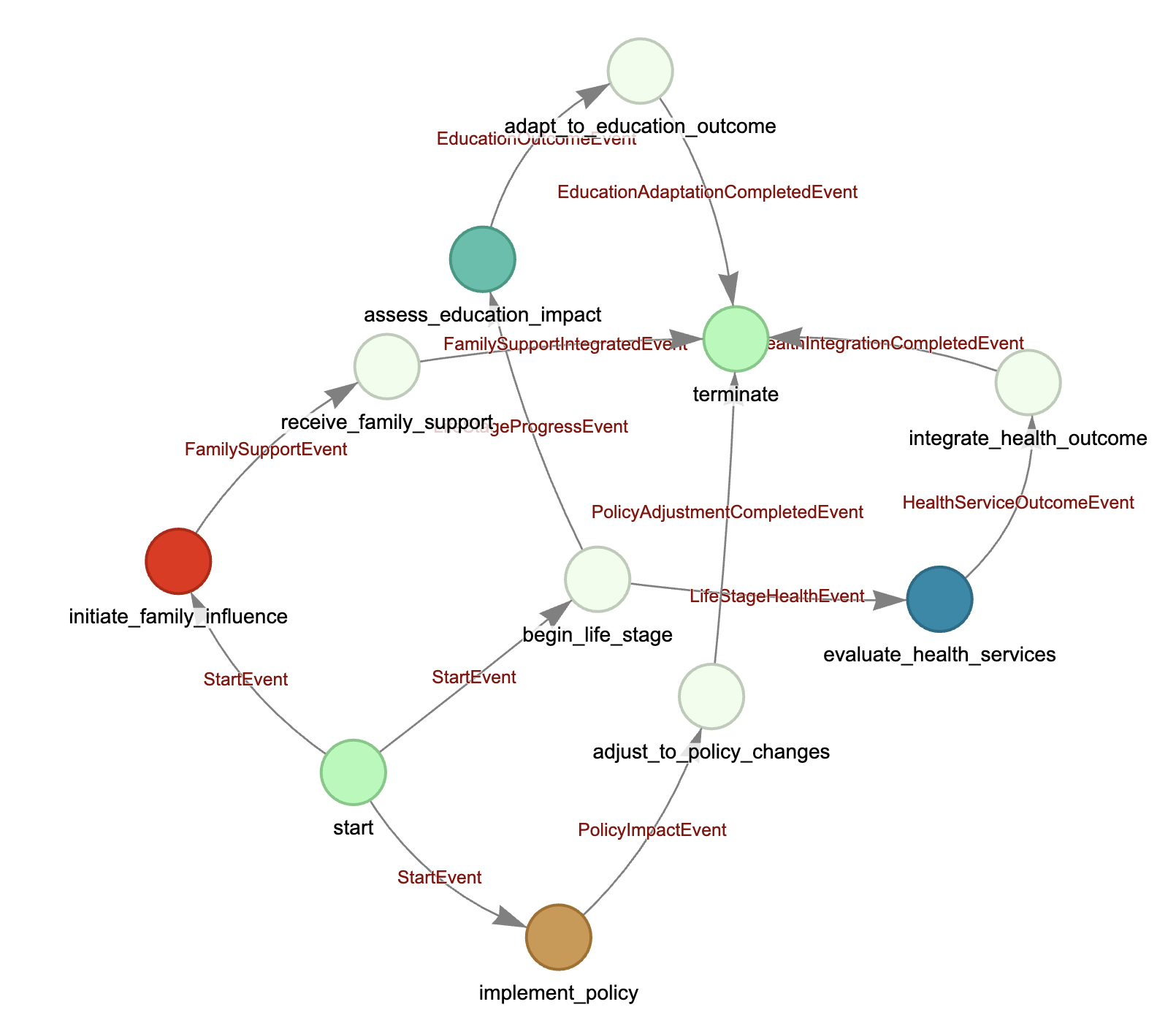}
        \caption{Behavior graph of Life Course Theory.}
    \end{figure}

    \subsubsection{Health Belief Model}

    $\bullet$ \textbf{Detailed description}

    \begin{textbox}
     \underline{\textcolor{red}{Health Belief Model}}\\
     
        System Goal:
        
        Simulate individual decision-making in the face of health threats using the Health Belief Model (HBM) and explore how perception of threat, benefits, barriers, and self-efficacy influence health behavior adoption.\\

        Agent Types:

        The system uses multiple agents representing individuals, families, communities, and government entities.\\

        Environment Description:

        The environment includes social settings where individuals, families, communities, and government interact, influenced by health policies, social support, and available resources.

        Health Decision Process:

        The simulation explores health decision processes in different contexts, including the impact of social support, information dissemination, and interventions.
    \end{textbox}

    $\bullet$ \textbf{Behavior graph}

    \begin{figure}[h]
        \centering
        \includegraphics[width=0.8\textwidth]{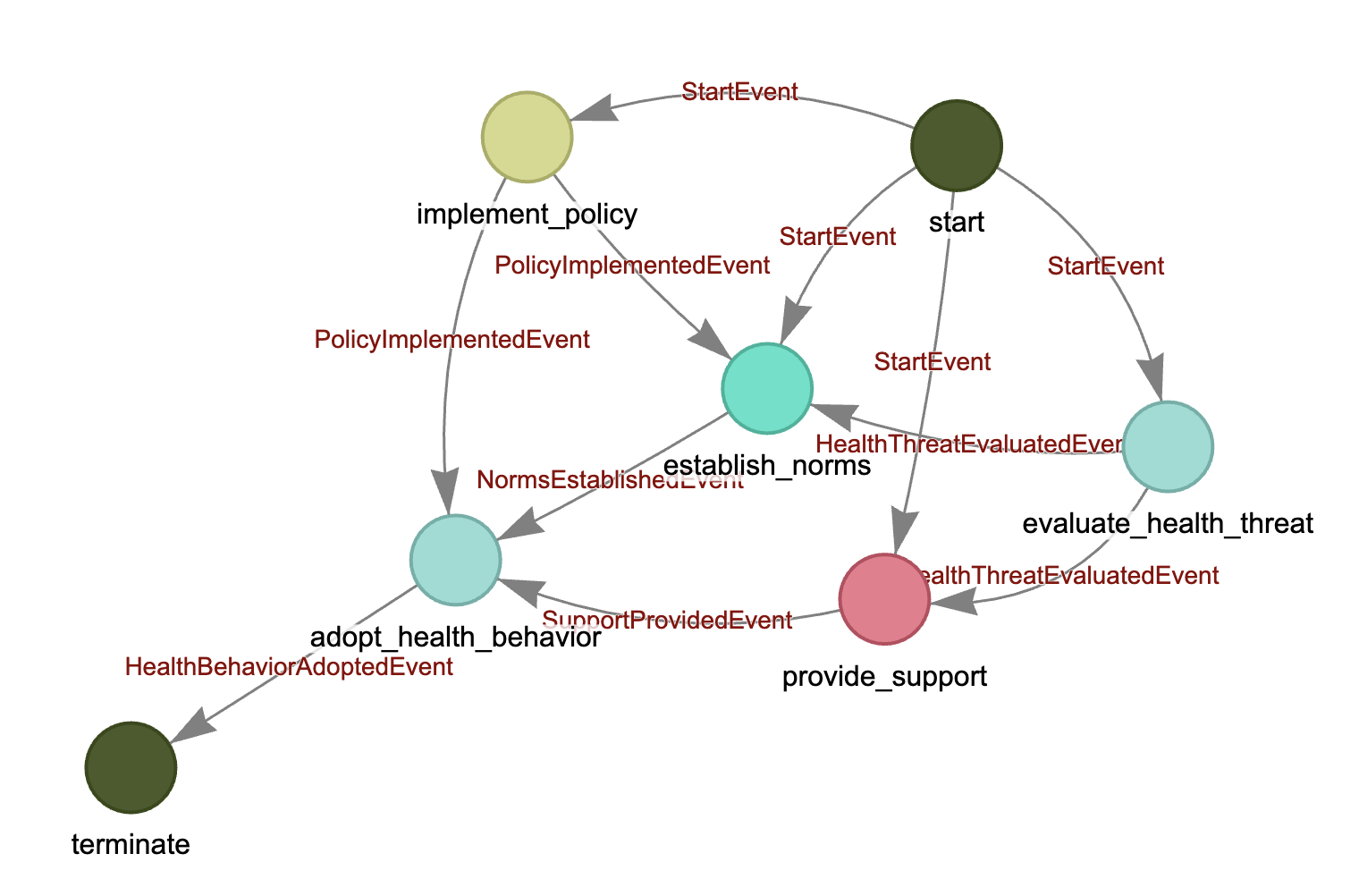}
        \caption{Behavior graph of Health Belief Model.}
    \end{figure}

    \subsubsection{Community Health Mobilization Theory}

    $\bullet$ \textbf{Detailed description}

    \begin{textbox}
     \underline{\textcolor{red}{Community Health Mobilization Theory}}\\
     
        System Goal:
        
        Simulate the process of community health mobilization focusing on interactions between community members, leaders, and public health experts to drive health improvements through collective action.\\

        Agent Types:

        Community Leader, Community Member, Public Health Expert\\

        Environment Description:

        Community background including socioeconomic status, cultural context, health issues, and available resources affecting mobilization outcomes.

        Health Challenges:

        Focus on specific health challenges such as vaccine uptake, chronic disease management, and infectious disease prevention.
    \end{textbox}

    $\bullet$ \textbf{Behavior graph}

    \begin{figure}[h]
        \centering
        \includegraphics[width=0.8\textwidth]{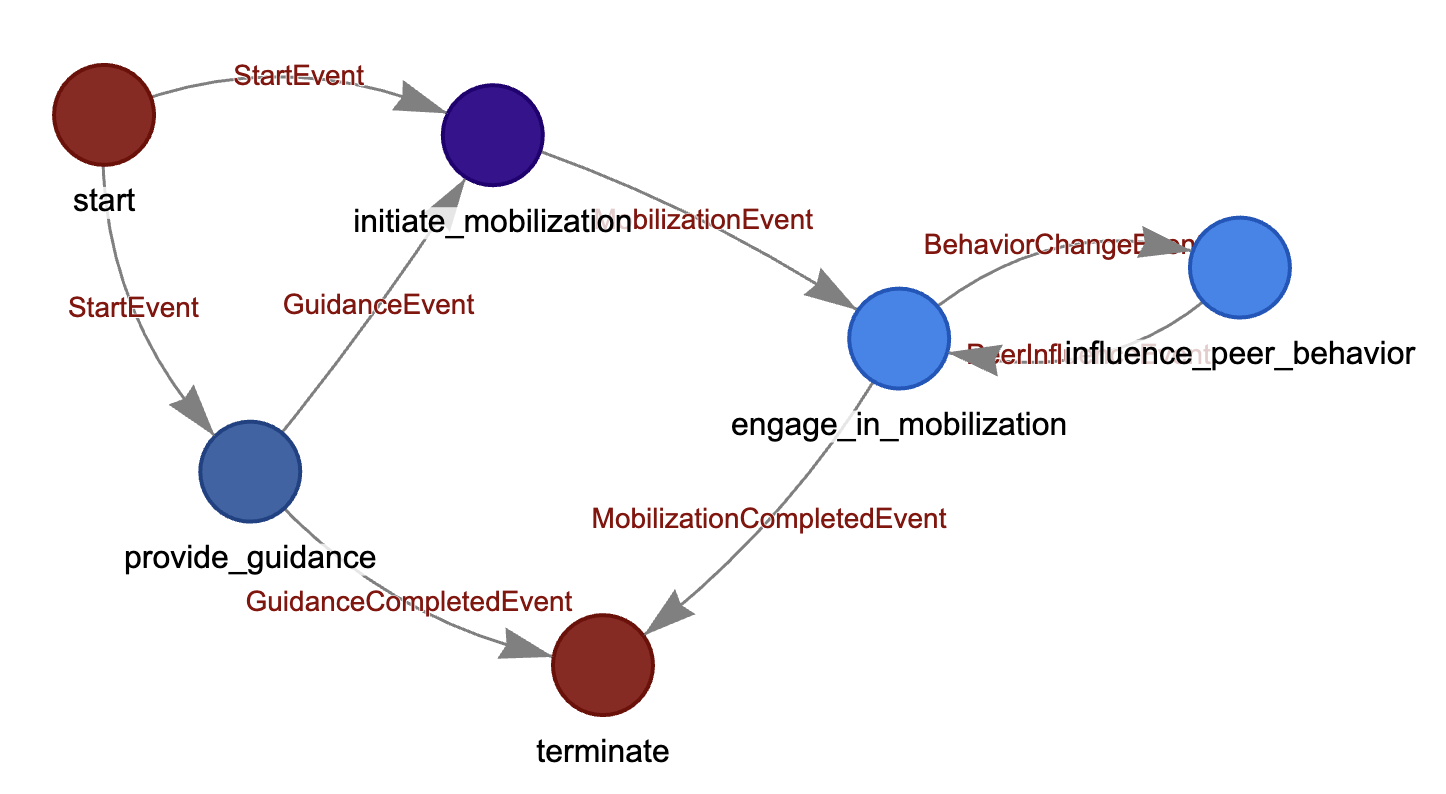}
        \caption{Behavior graph of Community Health Mobilization Theory.}
    \end{figure}

    \subsubsection{Reciprocal Altruism Theory}

    $\bullet$ \textbf{Detailed description}

    \begin{textbox}
     \underline{\textcolor{red}{Reciprocal Altruism Theory}}\\
     
        System Goal:
        
        The goal of the simulation is to model how reciprocal behaviors in a social network can promote the spread of healthy behaviors.\\

        Agent Types:
        
        Individual A as a health behavior advocate, Individual B as a recipient of health behavior change, Community Network as a facilitator of behavior spread.\\

        Environment Description:
        
        The environment is a social network representing a community where reciprocal behaviors and health information are exchanged.

        Health Behavior Spread:

        Simulation demonstrates how individuals influence others through reciprocal behaviors to promote health changes like smoking cessation and increased physical activity.

        Social Chain Reaction:

        Understanding how health behaviors spread through social interaction chains and evaluating the effects in various scenarios.
    \end{textbox}

    $\bullet$ \textbf{Behavior graph}

    \begin{figure}[h]
        \centering
        \includegraphics[width=0.9\textwidth]{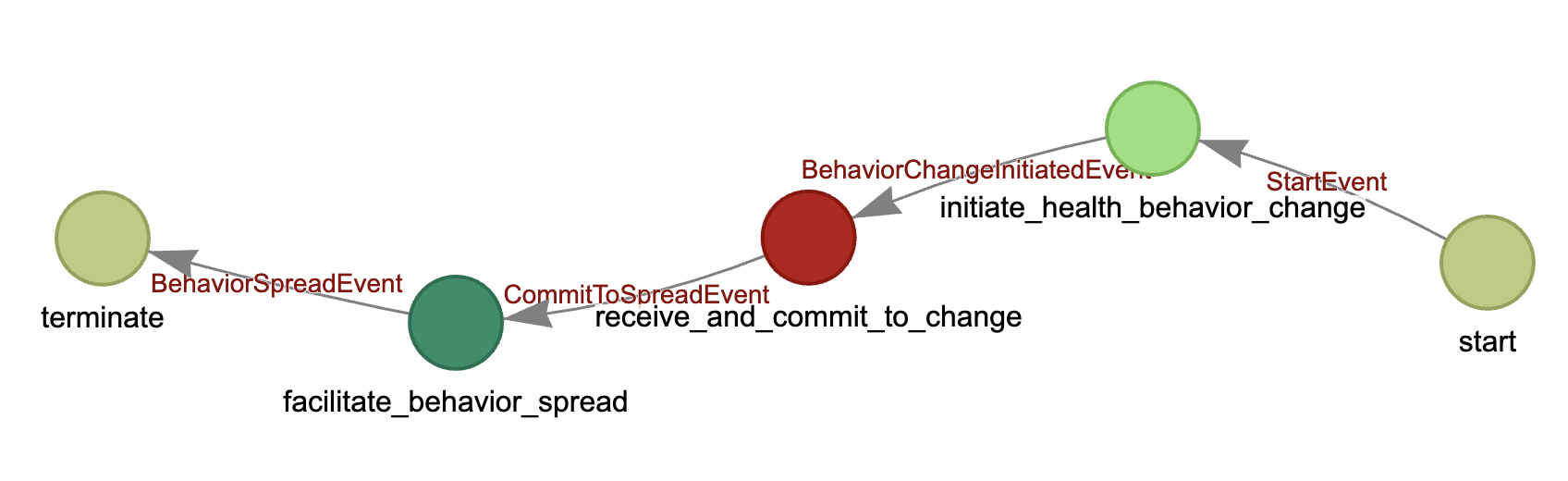}
        \caption{Behavior graph of Reciprocal Altruism Theory.}
    \end{figure}

    \subsection{Law}

    \subsubsection{Court Trial Simulation}

    $\bullet$ \textbf{Detailed description}

    \begin{textbox}
     \underline{\textcolor{red}{Court Trial Simulation}}\\
     
        System Goal:
        
        Simulate decision-making and interactions in court trials to explore how evidence presentation, persuasion strategies, cognitive biases, and procedural factors affect judicial outcomes, while assessing fairness and efficiency across different legal systems. Achieve quantifiable analysis of judicial processes through standardized models of judicial procedures and agent behaviors.\\

        Agent Types:

        Defendant, Prosecutor, Defense Lawyer, Judge, Jury, Witness. Each agent type has specific state variables and behavioral capabilities that influence their role in the trial process.\\

        Environment Description:

        Includes legal system type (common law with jury/civil law with judge ruling), court backlog affecting trial time and resource allocation, public attention impacting behavior pressure and media coverage, strength of relevant precedents affecting legal arguments and judgment basis, and procedural rule strictness influencing evidence admission standards and trial flow.
    \end{textbox}

    $\bullet$ \textbf{Behavior graph}

    \begin{figure}[h]
        \centering
        \includegraphics[width=0.8\textwidth]{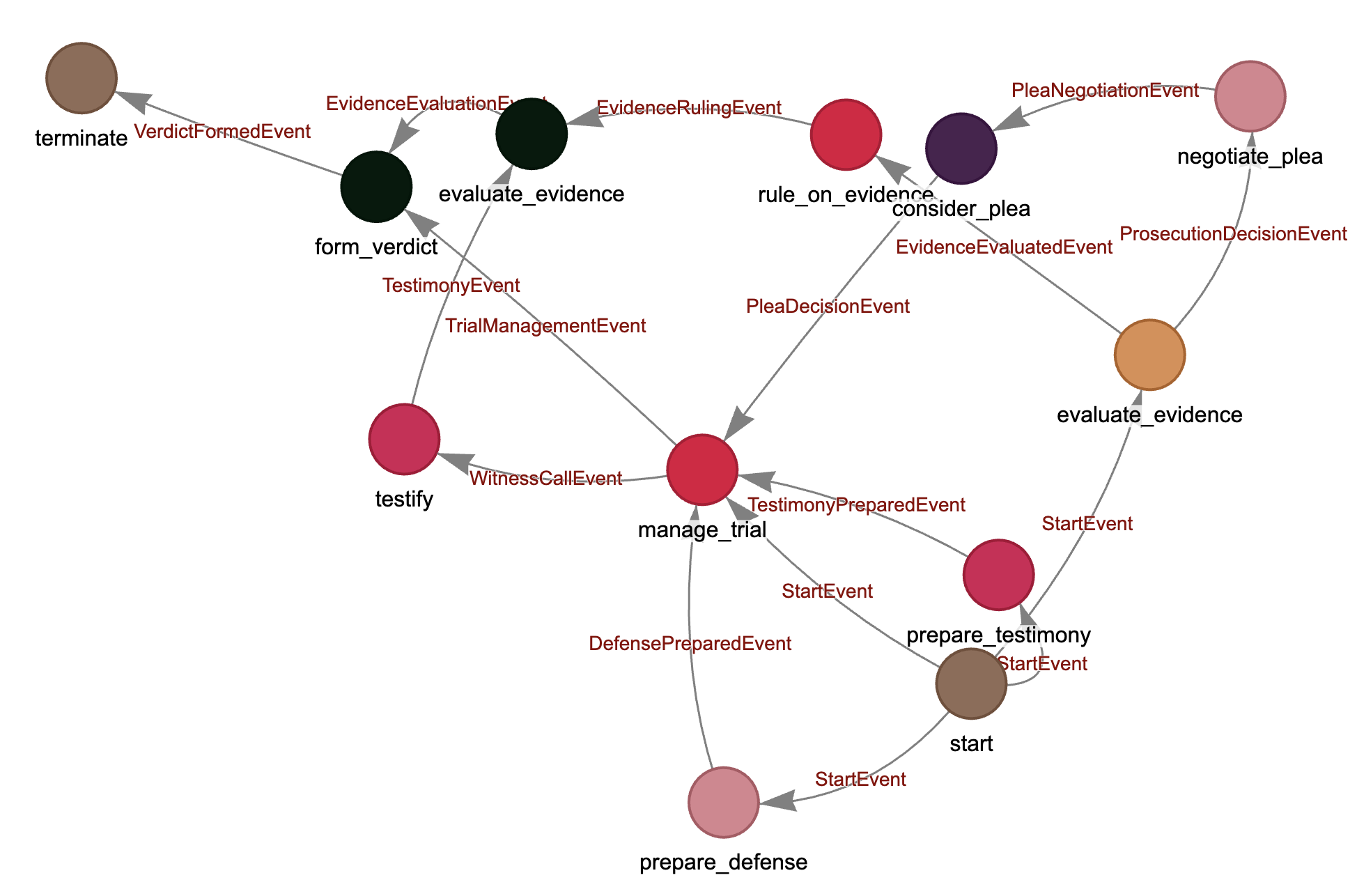}
        \caption{Behavior graph of Court Trial Simulation.}
    \end{figure}

    \subsubsection{Social Contract Theory}

    $\bullet$ \textbf{Detailed description}

    \begin{textbox}
     \underline{\textcolor{red}{Social Contract Theory}}\\
     
        System Goal:
        
        Use large language models to simulate social contract theory and explore its effects on individual freedom, government power, and social order.\\

        Agent Types:

        Individual agents, government agents, social group agents, contract breakdown agents, public policy agents.\\

        Environment Description:

        Simulated environments based on different social contract theories such as Hobbes, Locke, and Rousseau, exploring individual and government interactions.

        Focus on Conflict Resolution:

        Examine how social contracts address social conflicts, public welfare, rights protection, and government governance.

        Impact Analysis:

        Analyze how social contracts affect law, government governance, and social stability across different political systems and social structures.
    \end{textbox}

    $\bullet$ \textbf{Behavior graph}

    \begin{figure}[h]
        \centering
        \includegraphics[width=0.6\textwidth]{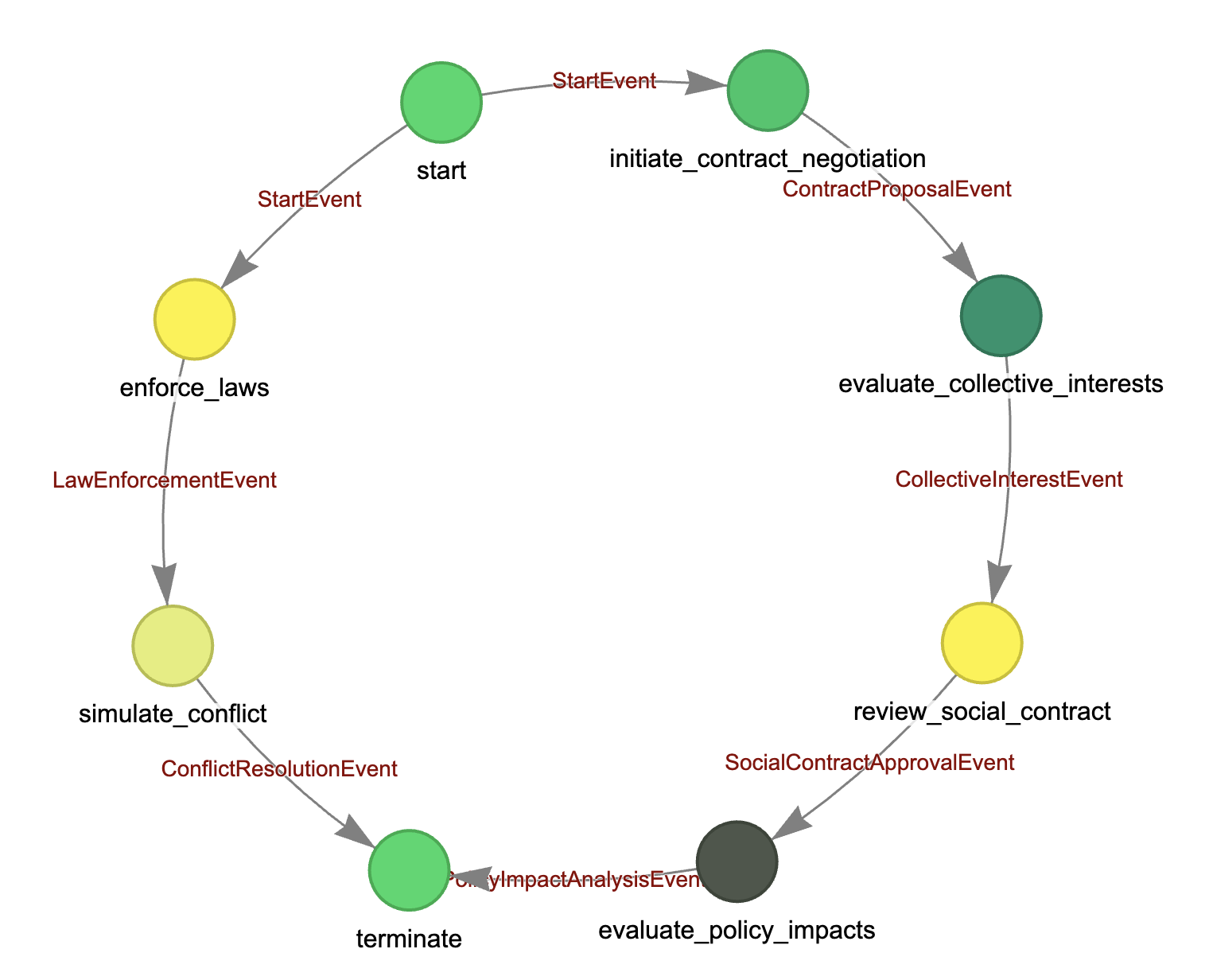}
        \caption{Behavior graph of Social Contract Theory.}
    \end{figure}

    \subsubsection{Case Law Model}

    $\bullet$ \textbf{Detailed description}

    \begin{textbox}
     \underline{\textcolor{red}{Case Law Model}}\\
     
        System Goal:
        
        Simulate case law using large language models (LLM) to analyze the evolution of court rulings and their impact on legal interpretation and precedent principles.\\

        Agent Types:

        Judge agents, case agents, precedent agents, socio-political environment agents, legal expert agents\\

        Environment Description:

        The environment includes various legal fields like criminal law, civil law, contract law, administrative law, influenced by social, political, and economic factors.

        Legal Evolution Analysis:

        Understand how legal rules evolve with court decisions and adapt to changes in society, politics, and economy.

        Precedent Application:

        Examine how precedents are applied or rejected in similar cases to maintain legal coherence and consistency.
    \end{textbox}

    $\bullet$ \textbf{Behavior graph}

    \begin{figure}[h]
        \centering
        \includegraphics[width=0.7\textwidth]{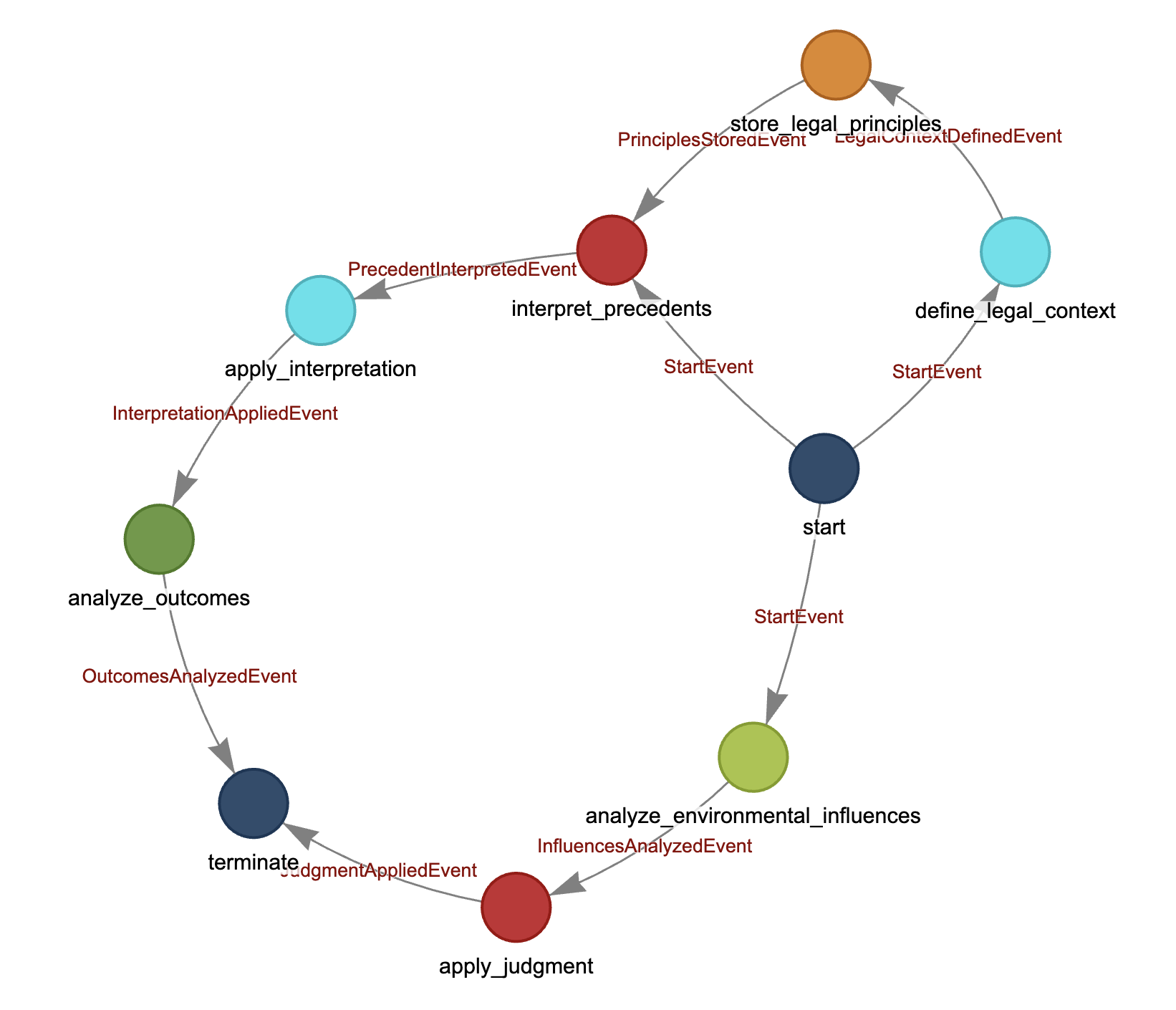}
        \caption{Behavior graph of Case Law Model.}
    \end{figure}

    \subsubsection{Tort Law and Compensation}

    $\bullet$ \textbf{Detailed description}

    \begin{textbox}
     \underline{\textcolor{red}{Tort Law and Compensation}}\\
     
        System Goal:
        
        Simulate compensation mechanisms in tort cases to explore how different compensation types affect case decisions and behaviors of parties involved.\\

        Agent Types:

        Plaintiff (victim), Defendant (tortfeasor), Judge (decision-maker)\\

        Environment Description:

        A simulated legal environment where agents interact based on roles in a tort case scenario.

        Compensation Types:

        Restitution compensation, actual loss compensation, punitive compensation

        Focus of Simulation:

        Modeling decisions on compensation amounts and types, analyzing their impact on behaviors and case outcomes
    \end{textbox}

    $\bullet$ \textbf{Behavior graph}

    \begin{figure}[h]
        \centering
        \includegraphics[width=0.5\textwidth]{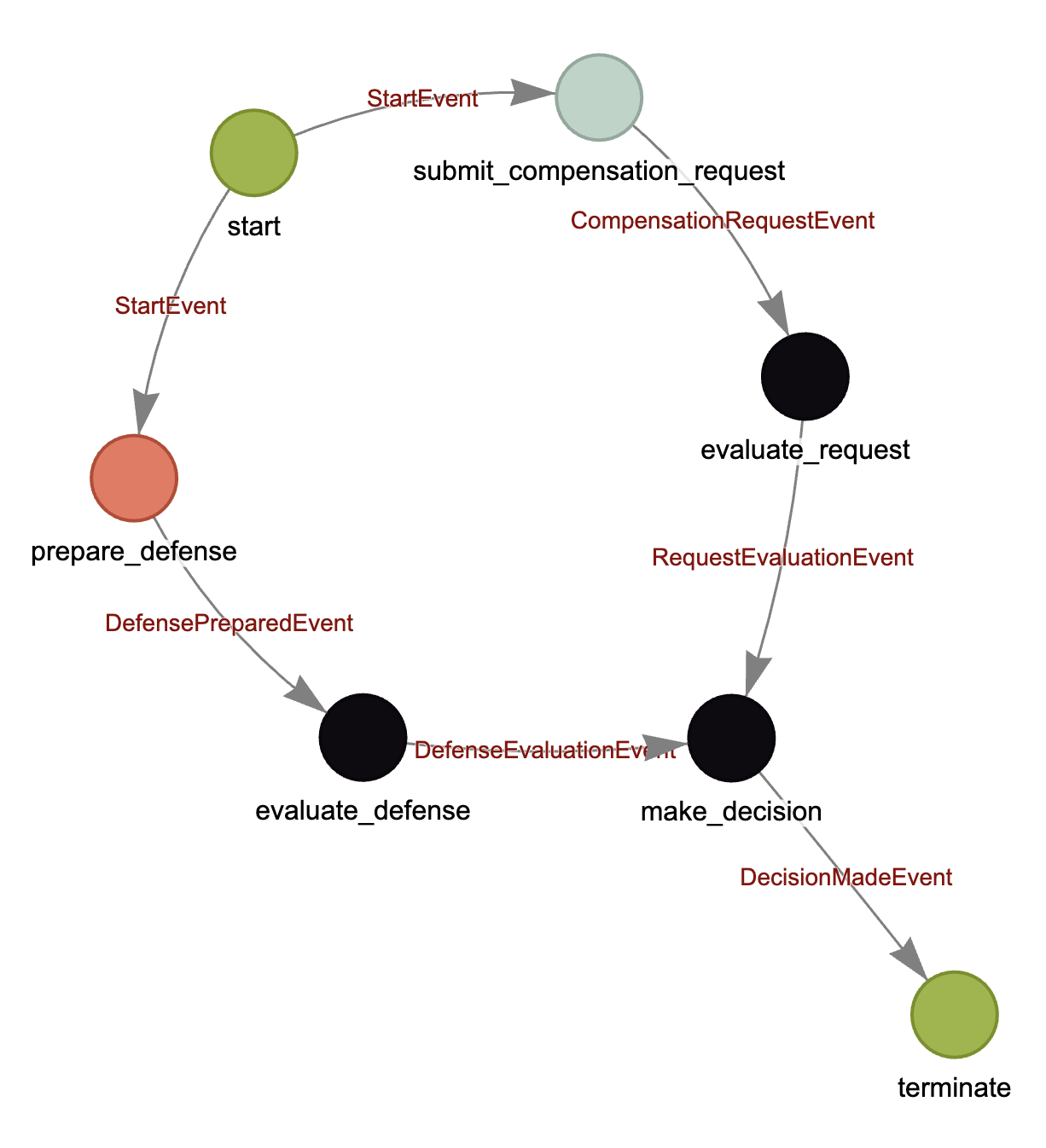}
        \caption{Behavior graph of Tort Law and Compensation.}
    \end{figure}

    \subsubsection{Unjust Enrichment}

    $\bullet$ \textbf{Detailed description}

    \begin{textbox}
     \underline{\textcolor{red}{Unjust Enrichment}}\\
     
        System Goal:
        
        To simulate scenarios of unjust enrichment and explore legal rules for resolution without contractual relationships.\\

        Agent Types:

        Plaintiff (victim seeking return of unjust enrichment), Defendant (party benefiting from unjust enrichment), Judge (adjudicates the case).\\

        Environment Description:

        A legal simulation environment involving cases of unjust enrichment without contractual agreements.

        Focus:

        The simulation focuses on identifying unjust enrichment and the legal mechanisms for restitution.

        Case Scenario:

        Involves a case where one party gains property or benefits through illegal means, requiring judicial decision on restitution.
    \end{textbox}

    $\bullet$ \textbf{Behavior graph}

    \begin{figure}[h]
        \centering
        \includegraphics[width=\textwidth]{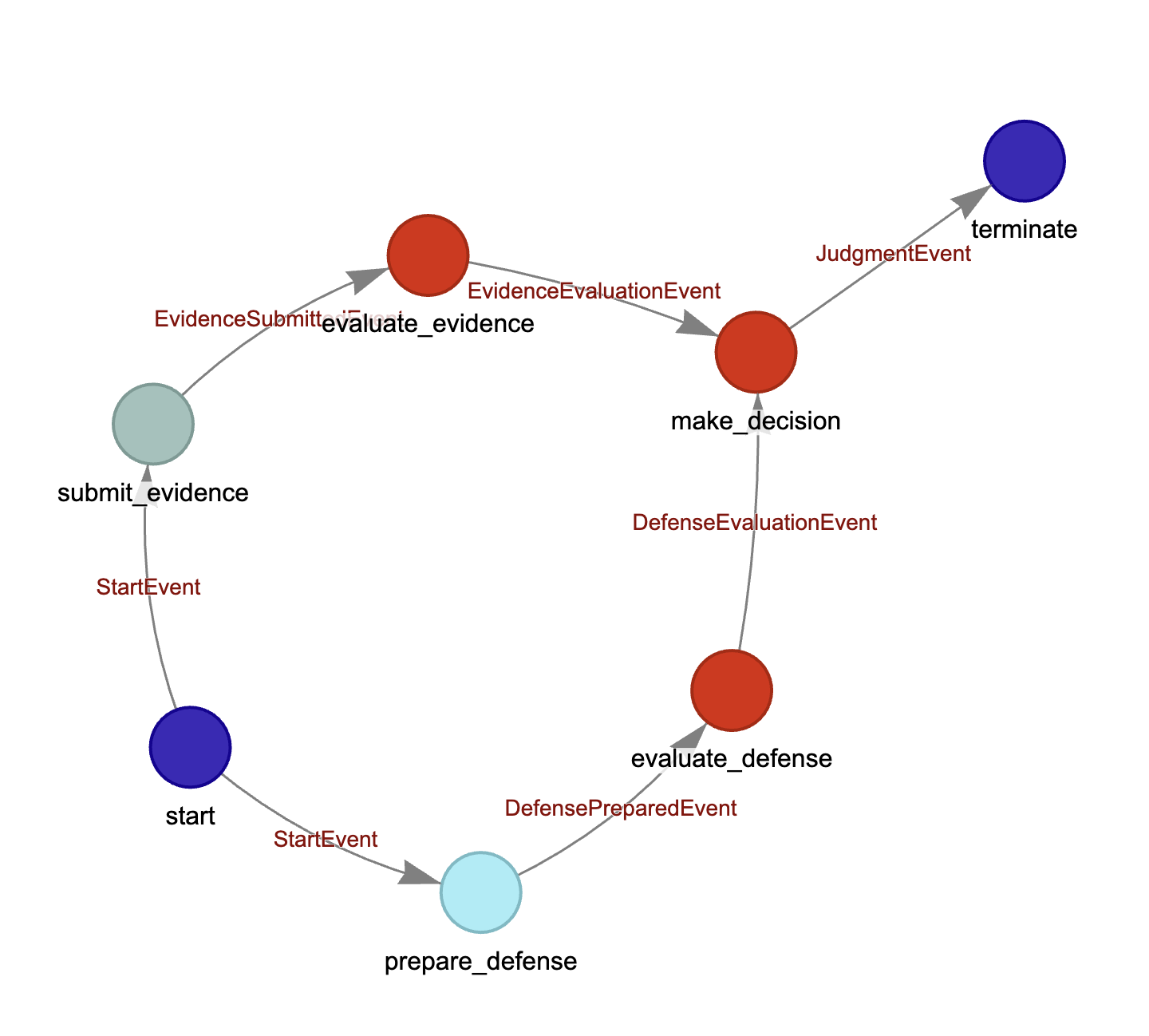}
        \caption{Behavior graph of Unjust Enrichment.}
    \end{figure}

    \subsubsection{Self-Defense and Excessive Defense}

    $\bullet$ \textbf{Detailed description}

    \begin{textbox}
     \underline{\textcolor{red}{Self-Defense and Excessive Defense}}\\
     
        System Goal:
        
        The goal is to simulate a simple violent conflict scenario to analyze how to judge if self-defense is reasonable within a legal framework and if excessive defense exists.\\

        Agent Types:

        Three agents: aggressor (attacker), defender (defendee), and judge (adjudicator).\\

        Environment Description:

        A simulated environment involving a violent conflict where the defender faces a threat from the aggressor, and a judge evaluates the situation.

        Simulation Focus:

        Focus on the necessity and proportionality of defense actions and how to make judgments under unclear circumstances based on legal rules.
    \end{textbox}

    $\bullet$ \textbf{Behavior graph}

    \begin{figure}[h]
        \centering
        \includegraphics[width=\textwidth]{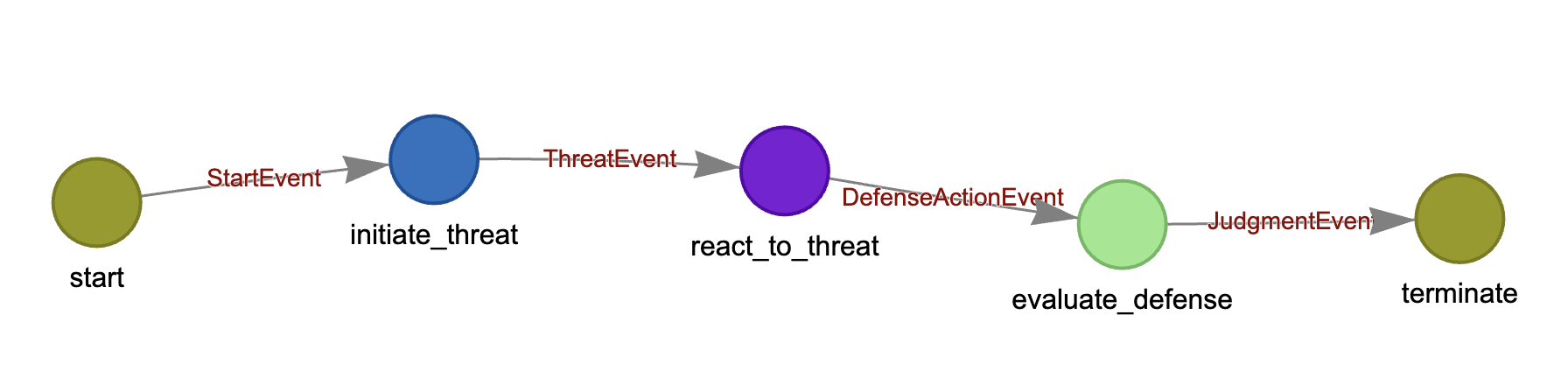}
        \caption{Behavior graph of Self-Defense and Excessive Defense.}
    \end{figure}

    \subsubsection{Work Hours and Overtime in Labor Law}

    $\bullet$ \textbf{Detailed description}

    \begin{textbox}
     \underline{\textcolor{red}{Work Hours and Overtime in Labor Law}}\\
     
        System Goal:
        
        Simulate a labor dispute case to explore the balance of rights between employees and employers regarding work hours and overtime pay.\\

        Agent Types:

        Employee, Employer, Judge\\

        Environment Description:

        A simulated legal environment where labor law is applied to resolve disputes concerning work hours and overtime pay.

        Dispute Focus:

        Focus on the recognition of overtime hours and rest periods within labor law.
    \end{textbox}

    $\bullet$ \textbf{Behavior graph}

    \begin{figure}[h]
        \centering
        \includegraphics[width=\textwidth]{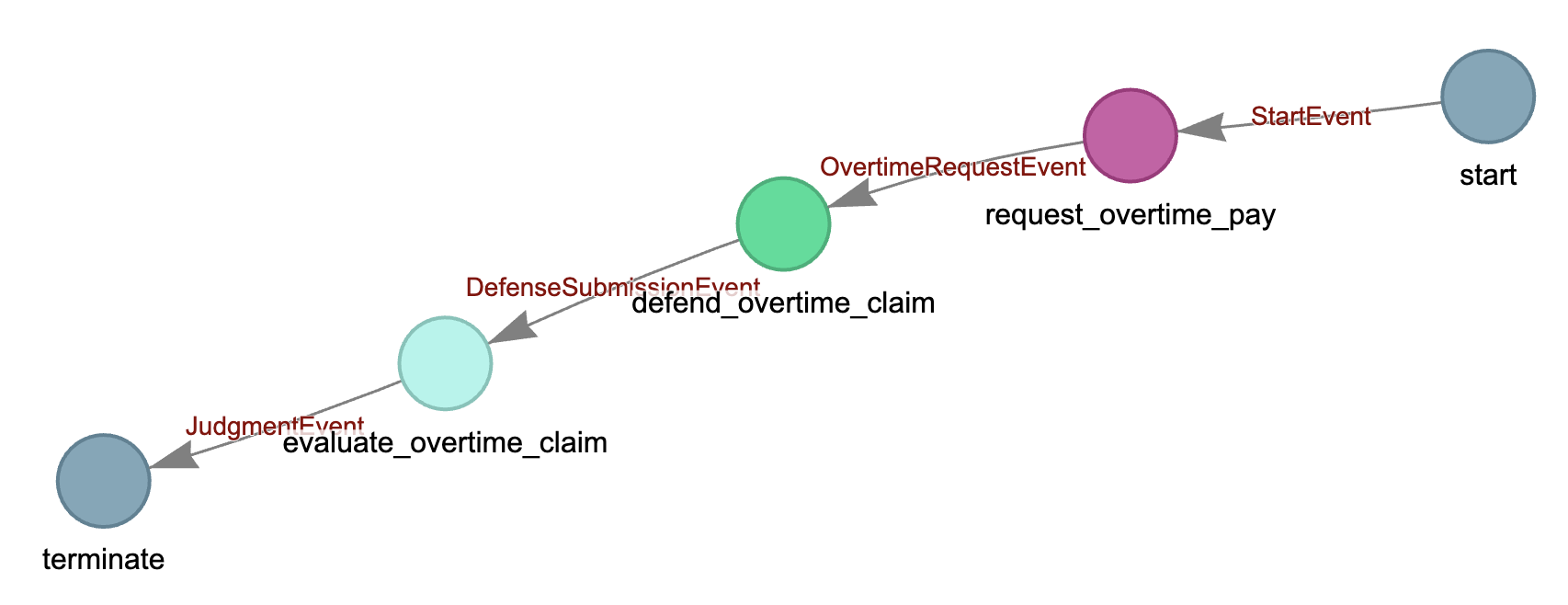}
        \caption{Behavior graph of Work Hours and Overtime in Labor Law.}
    \end{figure}

    \subsection{Communication}

    \subsubsection{Information Cascade and Silence}

    $\bullet$ \textbf{Detailed description}

    \begin{textbox}
     \underline{\textcolor{red}{Information Cascade and Silence}}\\
     
        System Goal:

        Simulate the process of rumor propagation and the formation of the spiral of silence in social networks, and study how mechanisms such as information verification, social pressure, opinion climate perception, and media environment influence information diffusion and willingness to express opinions publicly.\\

        Agent Types:

        The system includes five main agent types:

        (1) Ordinary Users (User): Represent individual users in social networks with attributes such as belief, expressed\_opinion, certainty, verification\_tendency, conformity\_tendency, and isolation\_fear.
        
        (2) Opinion Leaders: Represent influential users with additional attributes like influence, followers, and content\_creation, inheriting all attributes of ordinary users.
        
        (3) Media Organizations: Represent traditional or social media platforms with attributes like reach, credibility, editorial\_policy, and fact\_checking\_rigor.
        
        (4) Fact-Check Organizations: Represent third-party organizations dedicated to verifying information, with attributes like detection\_ability, response\_speed, credibility, and reach.
        
        (5) Platform Regulators: Represent social media platform content management teams with attributes such as monitoring\_ability, intervention\_strength, and response\_time.\\

        Environment Description:

        The environment consists of a social network structure defining connections between users, an information environment characterized by noise and diversity, and the degree of social polarization reflecting the division between different opinion groups.
    \end{textbox}

    $\bullet$ \textbf{Behavior graph}

    \begin{figure}[h]
        \centering
        \includegraphics[width=0.6\textwidth]{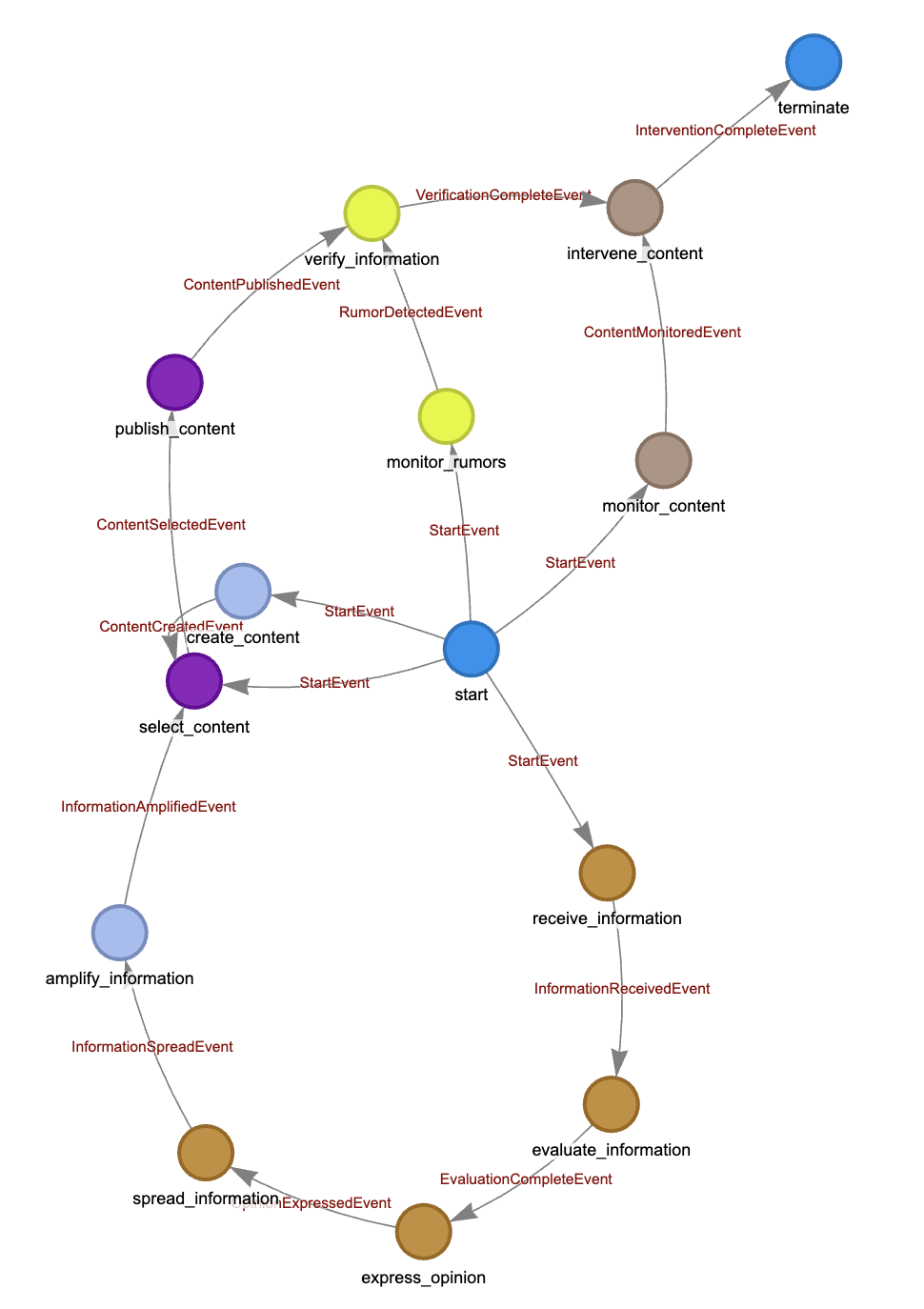}
        \caption{Behavior graph of Information Cascade and Silence.}
    \end{figure}

    \subsubsection{Two Step Flow Model}

    $\bullet$ \textbf{Detailed description}

    \begin{textbox}
     \underline{\textcolor{red}{Two Step Flow Model}}\\
     
        System Goal:
        
        Simulate how information spreads in society and validate the two-step flow model hypothesis.\\

        Agent Types:

        Media agent, opinion leader agents, and public agents.\\

        Environment Description:

        A social network where agents are connected, with stronger connections between opinion leaders and weaker ones among public agents.

        Information Flow:

        Information flows from media to opinion leaders and then to the public.

        Role of Opinion Leaders:

        Opinion leaders act as filters and interpreters of information, influencing public perception.
    \end{textbox}

    $\bullet$ \textbf{Behavior graph}

    \begin{figure}[h]
        \centering
        \includegraphics[width=0.6\textwidth]{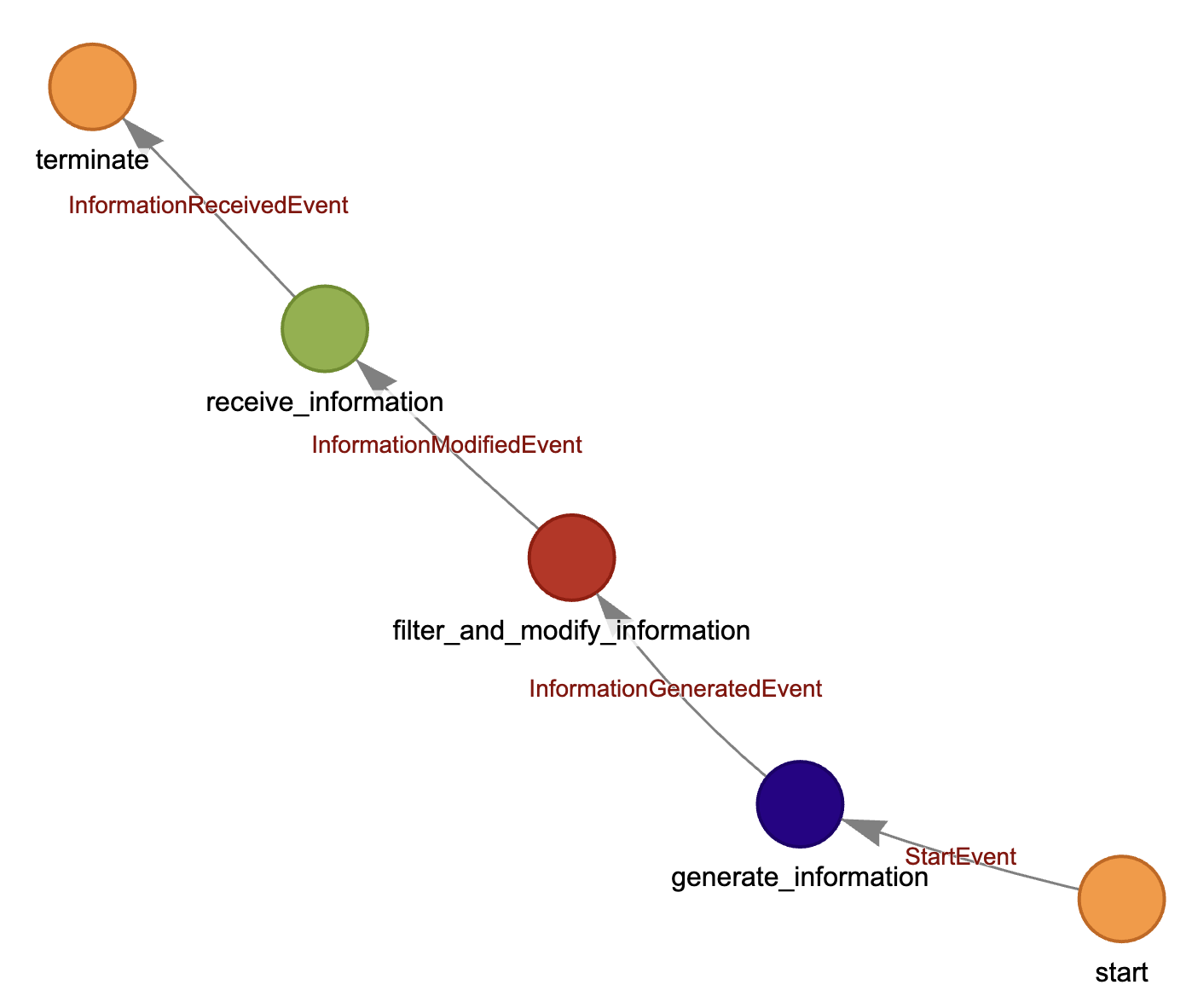}
        \caption{Behavior graph of Two Step Flow Model.}
    \end{figure}

    \subsubsection{Uses and Gratifications Theory}

    $\bullet$ \textbf{Detailed description}

    \begin{textbox}
     \underline{\textcolor{red}{Uses and Gratifications Theory}}\\
     
        System Goal:
        
        Simulate how audiences choose and use different types of media based on personal needs to validate the basic assumptions of the uses and gratifications theory.\\

        Agent Types:

        Audience Agents representing individuals with varying needs and preferences, and Media Agents representing different media types.\\

        Environment Description:

        A diverse media environment where agents interact with multiple media types to satisfy varying needs such as entertainment, information, and social interaction.

        System Assumptions:

        Agents actively choose media content based on personal needs, with media diversity and content differentiation influencing choices and emotional responses.
    \end{textbox}

    $\bullet$ \textbf{Behavior graph}

    \begin{figure}[h]
        \centering
        \includegraphics[width=0.6\textwidth]{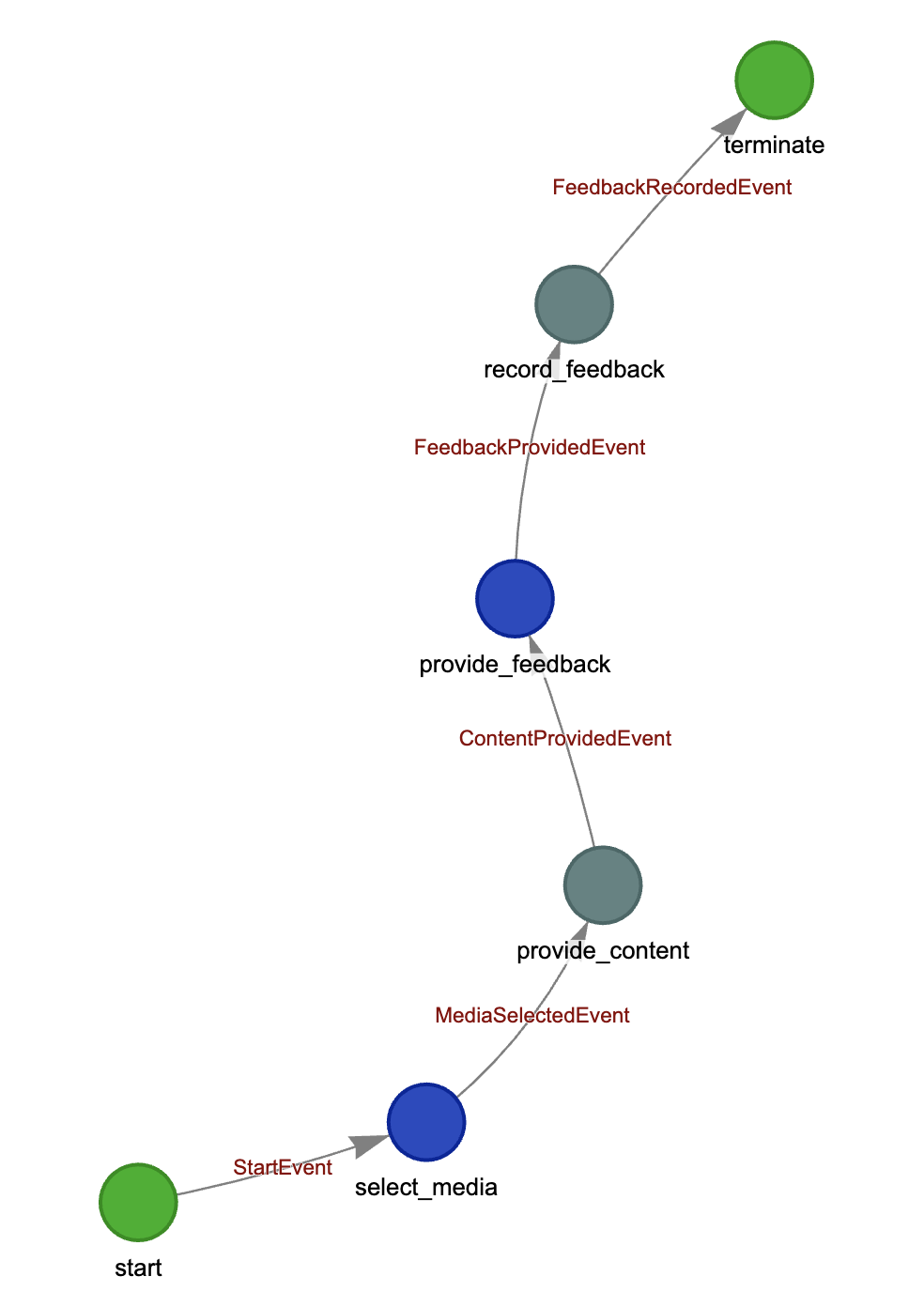}
        \caption{Behavior graph of Uses and Gratifications Theory.}
    \end{figure}

    \subsubsection{Diffusion of Innovations}

    $\bullet$ \textbf{Detailed description}

    \begin{textbox}
     \underline{\textcolor{red}{Diffusion of Innovations}}\\
     
        System Goal:
        
        Simulate how innovations spread in a social system, analyze how different types of individuals influence the diffusion process, and validate the basic assumptions of information dissemination models.\\

        Agent Types:

        Innovators, Early Adopters, Early Majority, Late Majority, Laggards\\

        Environment Description:

        A social system consisting of various individuals connected through a social network where the diffusion of innovation is influenced by network structure, social status, and trust in innovation.

        Individual Influence:

        Different types of individuals (innovators, early adopters, etc.) affect the speed of innovation acceptance.

        Diffusion Factors:

        Speed of innovation diffusion depends on network structure, social status, and trust in the innovation.
    \end{textbox}

    $\bullet$ \textbf{Behavior graph}

    \begin{figure}[h]
        \centering
        \includegraphics[width=0.7\textwidth]{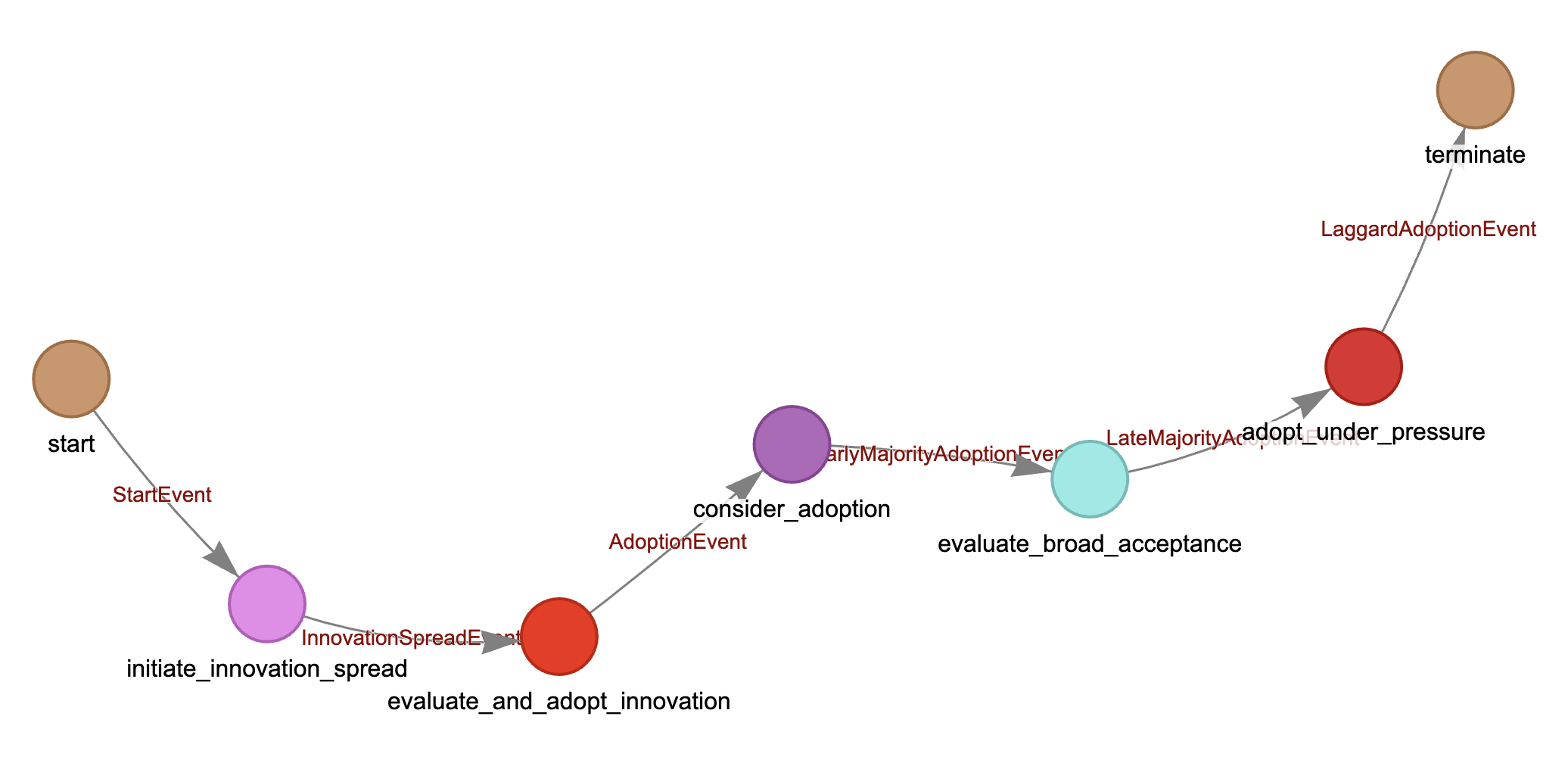}
        \caption{Behavior graph of Diffusion of Innovations.}
    \end{figure}

    \subsubsection{Cultural Globalization}

    $\bullet$ \textbf{Detailed description}

    \begin{textbox}
     \underline{\textcolor{red}{Cultural Globalization}}\\
     
        System Goal:
        
        Simulate the process of cultural globalization and study how cultural elements spread across different cultures and regions through social interactions, technological dissemination, and globalization processes. Explore the impact of cultural globalization on local cultures, values, and social behaviors.\\

        Agent Types:

        Cultural Agents represent individuals with specific cultural backgrounds, including language, customs, values, and behavior patterns. Cultural Product Agents represent different types of cultural products or symbols such as movies, music, food, fashion, etc.\\

        Environment Description:

        The environment consists of different regions with varying degrees of cultural openness, economic development levels, and information flow. Agents interact through social networks, media, tourism, and multinational corporations.

        Cultural Elements:

        Cultural elements like entertainment, customs, food, and language cross borders through channels like social networks, media, tourism, and multinational corporations.

        Cultural Influence Factors:

        The process of cultural globalization is influenced by factors such as regional cultural openness, economic development, and information flow.
    \end{textbox}

    $\bullet$ \textbf{Behavior graph}

    \begin{figure}[h]
        \centering
        \includegraphics[width=0.8\textwidth]{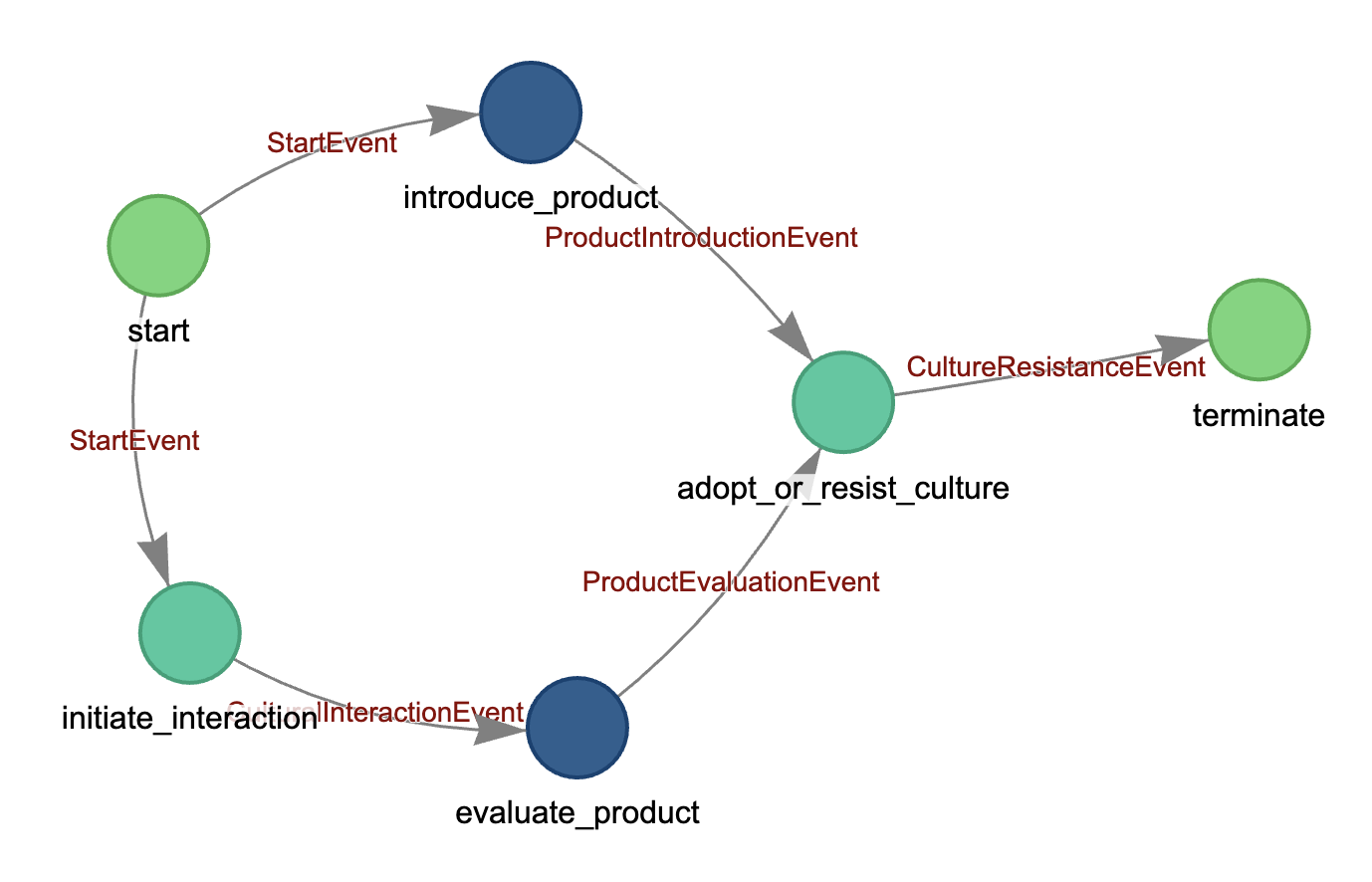}
        \caption{Behavior graph of Cultural Globalization.}
    \end{figure}

    \subsubsection{Agenda Setting Theory}

    $\bullet$ \textbf{Detailed description}

    \begin{textbox}
     \underline{\textcolor{red}{Agenda Setting Theory}}\\
     
        System Goal:
        
        Simulate how media influences public agenda by selecting specific topics and study the process of media agenda setting and its impact on public attention.\\

        Agent Types:

        Media Agents representing media institutions or news platforms and Public Agents representing audiences or public groups.\\

        Environment Description:

        The environment includes a topic library with various issues like social problems, political events, and economic phenomena, allowing media to select topics for reporting.

        System Assumptions:

        Media influences public focus by selecting and emphasizing specific topics, shaping public attitudes and behaviors. Interaction and feedback mechanisms exist between media agenda and public agenda.
    \end{textbox}

    $\bullet$ \textbf{Behavior graph}

    \begin{figure}[h]
        \centering
        \includegraphics[width=0.6\textwidth]{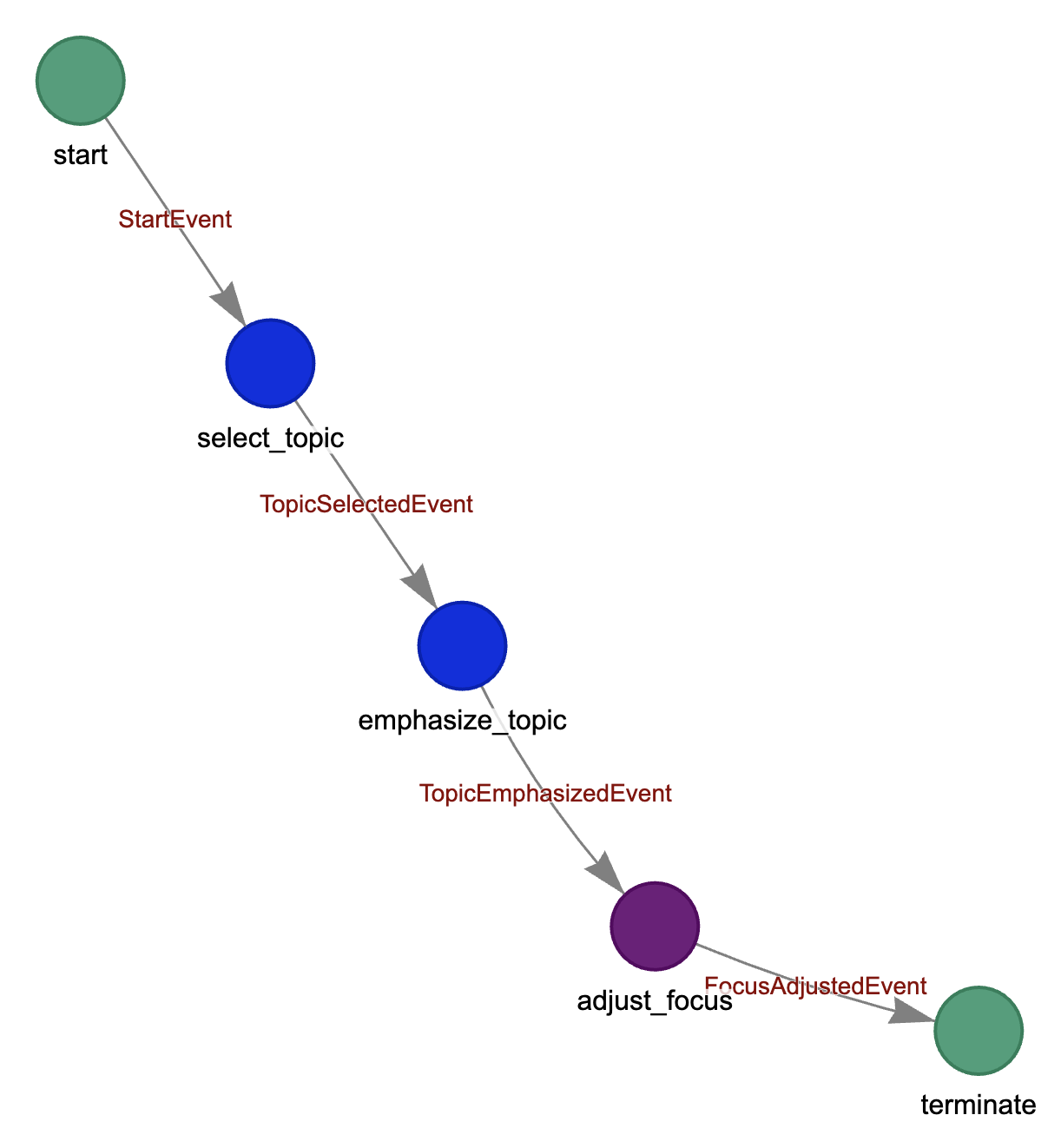}
        \caption{Behavior graph of Agenda Setting Theory.}
    \end{figure}

\section{A Complete Example of Running YuLan-OneSim}
    To provide an intuitive understanding of how our simulator operates, this section presents a complete example covering simulation scenario construction, execution, and result feedback.
    In specific, we take job market simulation as an example, and introduce how to leverage our simulator to conduct social simulation.
    
\subsection{Simulation scenario construction}
\begin{figure}[htbp]
    \centering
    \includegraphics[width=0.9\textwidth]{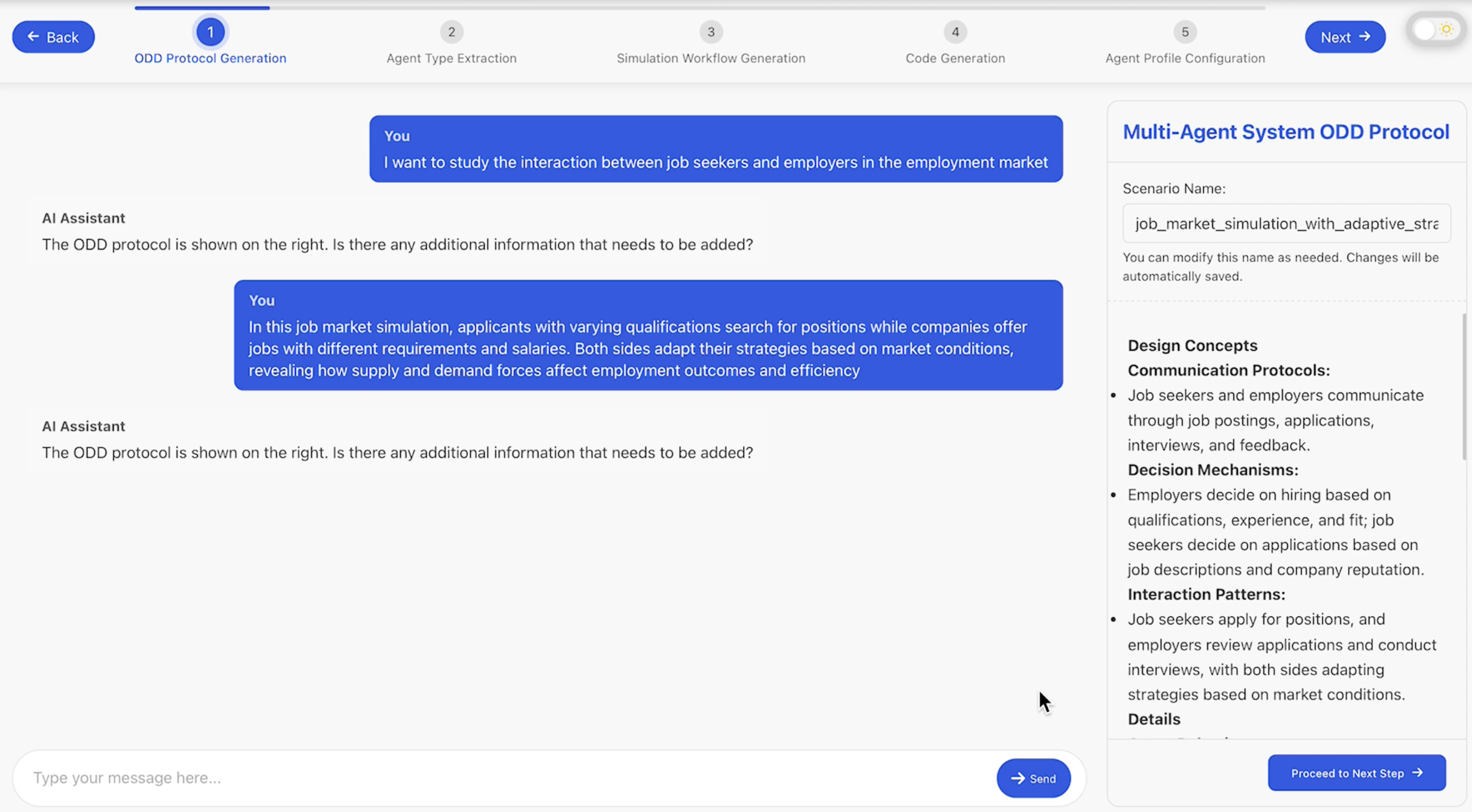}
    \caption{ODD protocol generation through natural language interaction. The user describes their simulation objectives, and the system extracts and refines the ODD protocol based on dialogue.}
    \label{fig:odd_generation}
\end{figure}

To construct a social simulation scenario, users can first engage in dialogue with the ODD Agent, describing their requirements in natural language. The LLM will extract the ODD protocol based on the current description and display it on the right side. If the description is incomplete, the agent will ask questions to clarify. This process is illustrated in Figure~\ref{fig:odd_generation}.

\begin{figure}[htbp]
    \centering
    \includegraphics[width=0.9\textwidth]{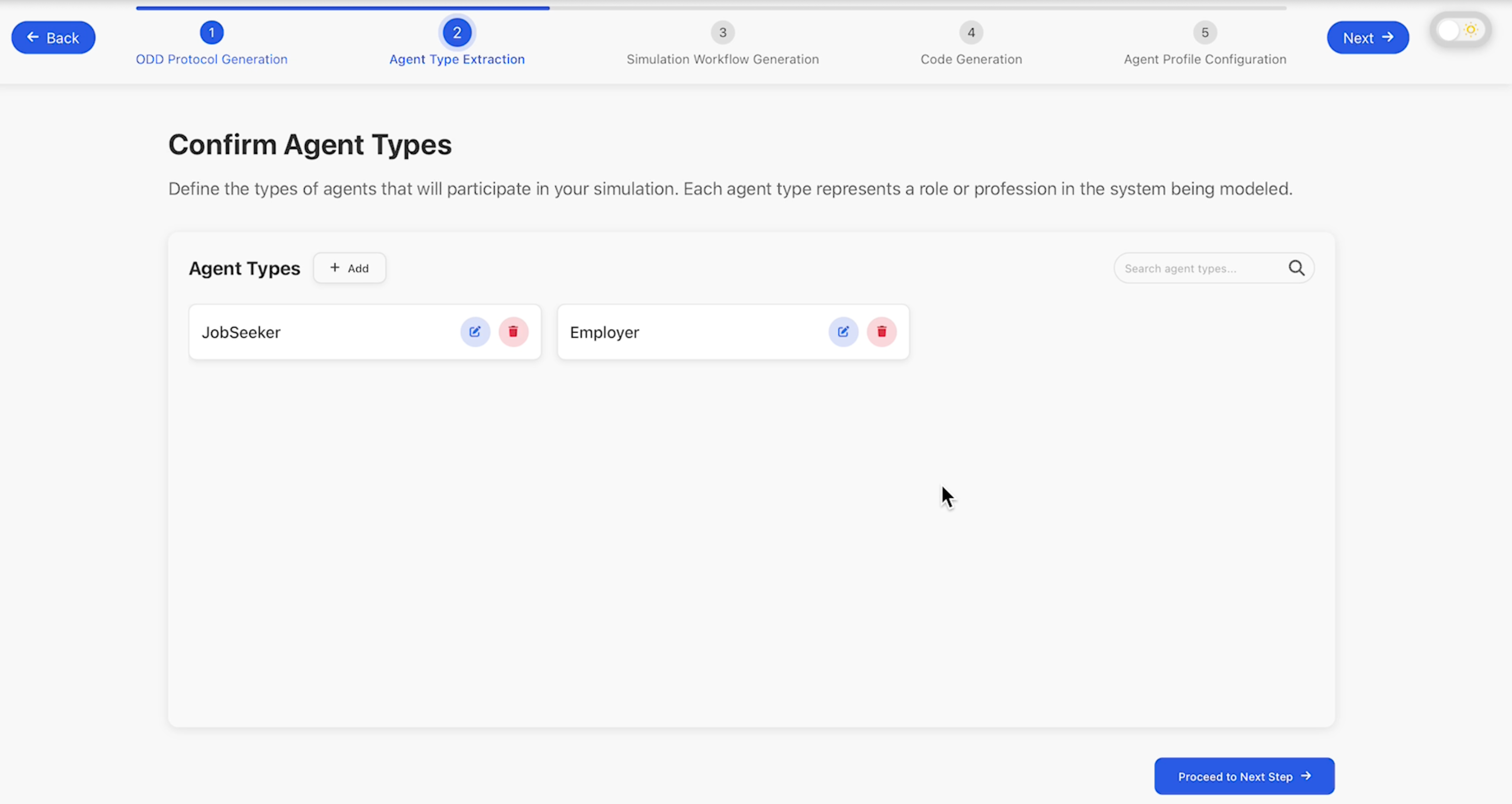}
    \caption{Agent type extraction and confirmation interface. Users can review and adjust the agent types identified from the ODD protocol.}
    \label{fig:agent_types}
\end{figure}

After confirming the ODD protocol, the LLM extracts agent types and their corresponding descriptions from the simulation scenario. Users can add, remove, or modify agent types and descriptions at this stage, as shown in Figure~\ref{fig:agent_types}.

\begin{figure}[htbp]
    \centering
    \includegraphics[width=0.9\textwidth]{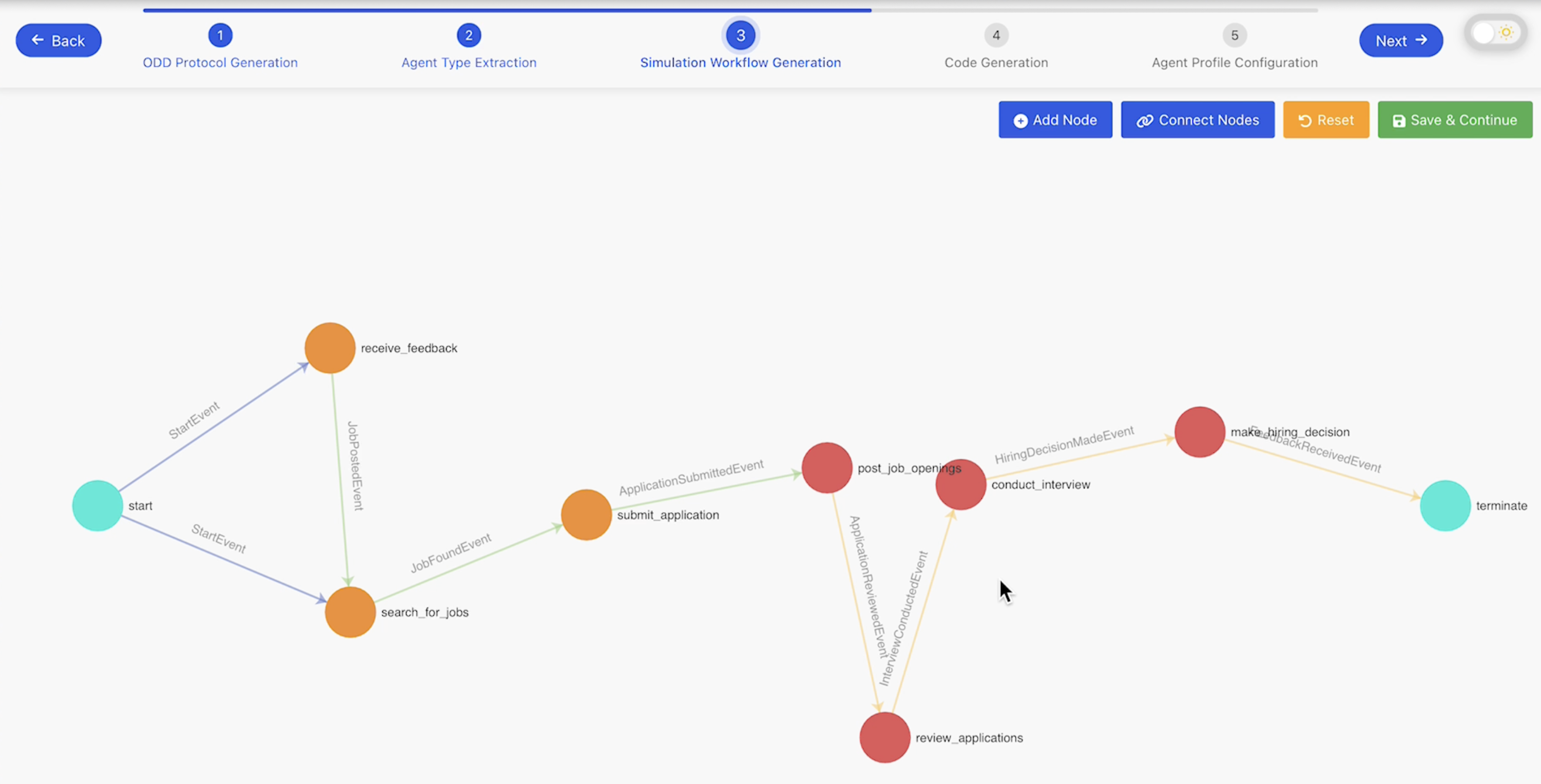}
    \caption{Behavior graph generation interface. The system visualizes agent actions as nodes and events as edges, with different colors representing different agent types.}
    \label{fig:behavior_graph}
\end{figure}

Next, the LLM analyzes the ODD protocol and extracted agent types to determine agent behaviors and interaction logic, generating an agent behavior graph. Each node specifically displays information about the current agent's action, while edges show event information. Different colors represent actions from different types of agents. Users can modify the generated behavior graph, creating new nodes or edges, as shown in Figure~\ref{fig:behavior_graph}.

\begin{figure}[htbp]
    \centering
    \includegraphics[width=0.9\textwidth]{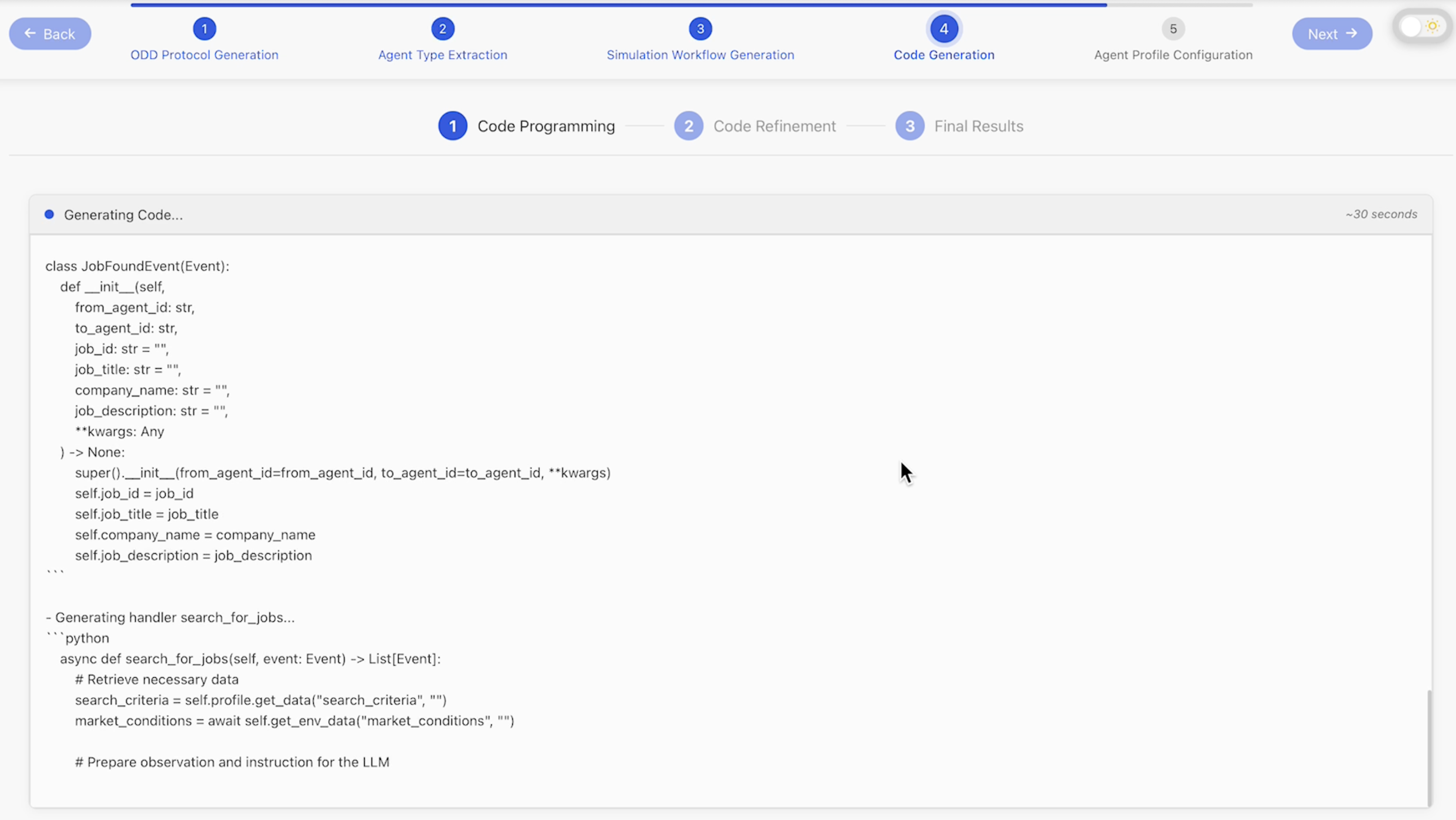}
    \caption{Code generation interface showing real-time code generation for simulation components.}
    \label{fig:code_generation}
\end{figure}

After verifying the behavior graph, the LLM generates corresponding code based on the graph, displaying the generation process in real-time. The system undergoes iterative checking and repair processes. Once validation is complete, users can review and modify the code for each node and edge, as shown in Figure~\ref{fig:code_generation}.

\begin{figure}[htbp]
    \centering
    \includegraphics[width=0.9\textwidth]{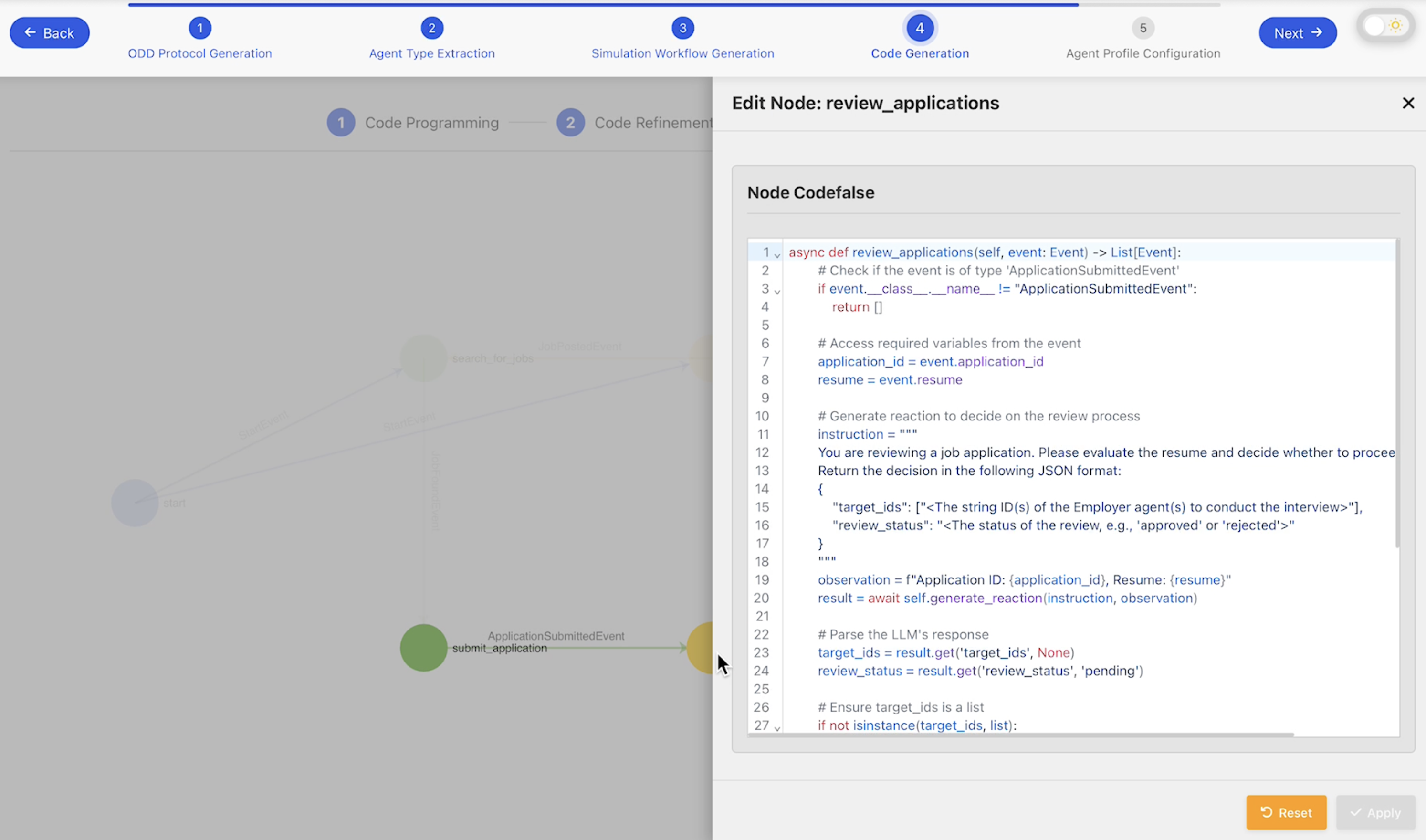}
    \caption{Detailed view of generated code for specific nodes in the behavior graph, allowing users to review and modify the implementation.}
    \label{fig:code_detail}
\end{figure}

\begin{figure}[htbp]
    \centering
    \includegraphics[width=0.9\textwidth]{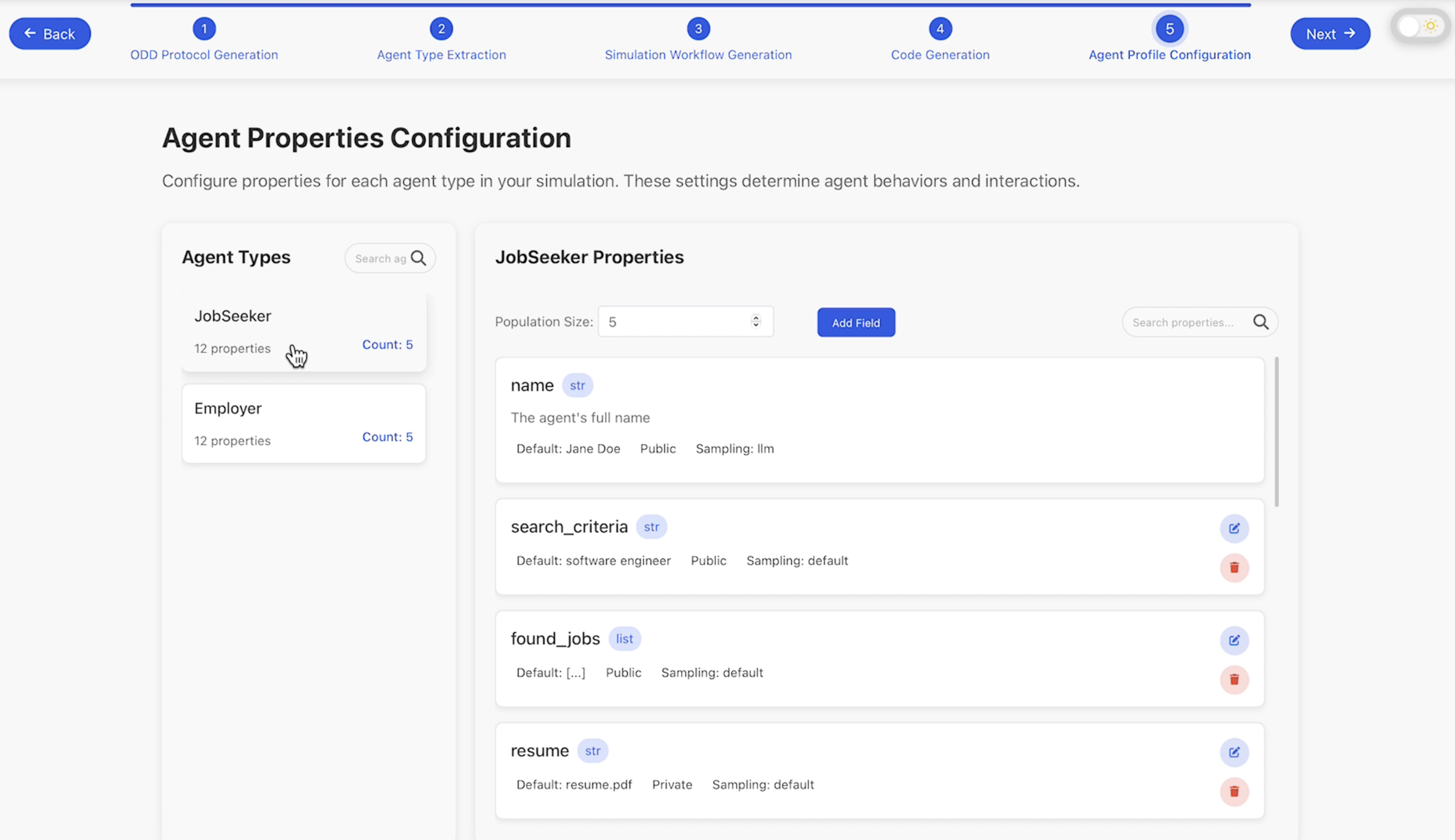}
    \caption{Agent profile configuration interface, allowing users to define agent attributes and population sizes.}
    \label{fig:profile_configuration}
\end{figure}

After code generation, the final step in scenario construction is generating data for simulation execution. The LLM creates an agent profile schema based on the ODD protocol and generated code. Users can add, remove, or modify agent profile attributes and specify the number of agent profiles to generate. Environmental data and relationship data between agents are automatically generated along with the profiles, as shown in Figure~\ref{fig:profile_configuration}.

\subsection{Simulation execution}
Once scenario construction is complete, the simulation can begin. Users can select scenarios from the pre-built scenario library to start simulation.

\begin{figure}[htbp]
    \centering
    \includegraphics[width=0.9\textwidth]{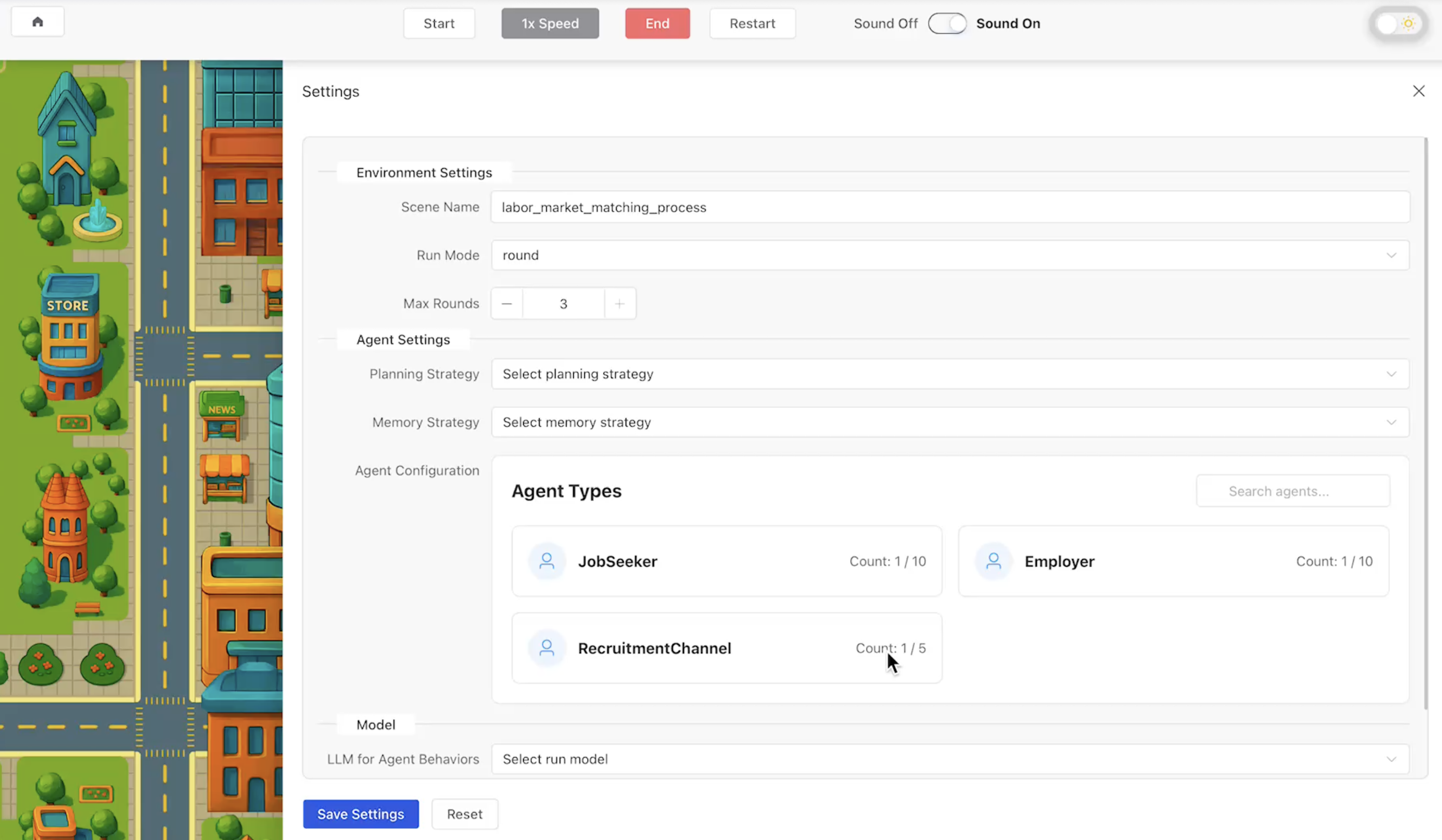}
    \caption{Simulation configuration interface allowing users to set execution parameters such as running mode, planning strategy, and agent counts.}
    \label{fig:simulation_settings}
\end{figure}

Before starting, users need to configure the simulation settings. On the settings page, users can configure the simulation running mode (round or tick), number of simulation steps, agent planning and memory strategies, number of agents, and the LLM driving the agents, as shown in Figure~\ref{fig:simulation_settings}.

\begin{figure}[htbp]
    \centering
    \includegraphics[width=0.9\textwidth]{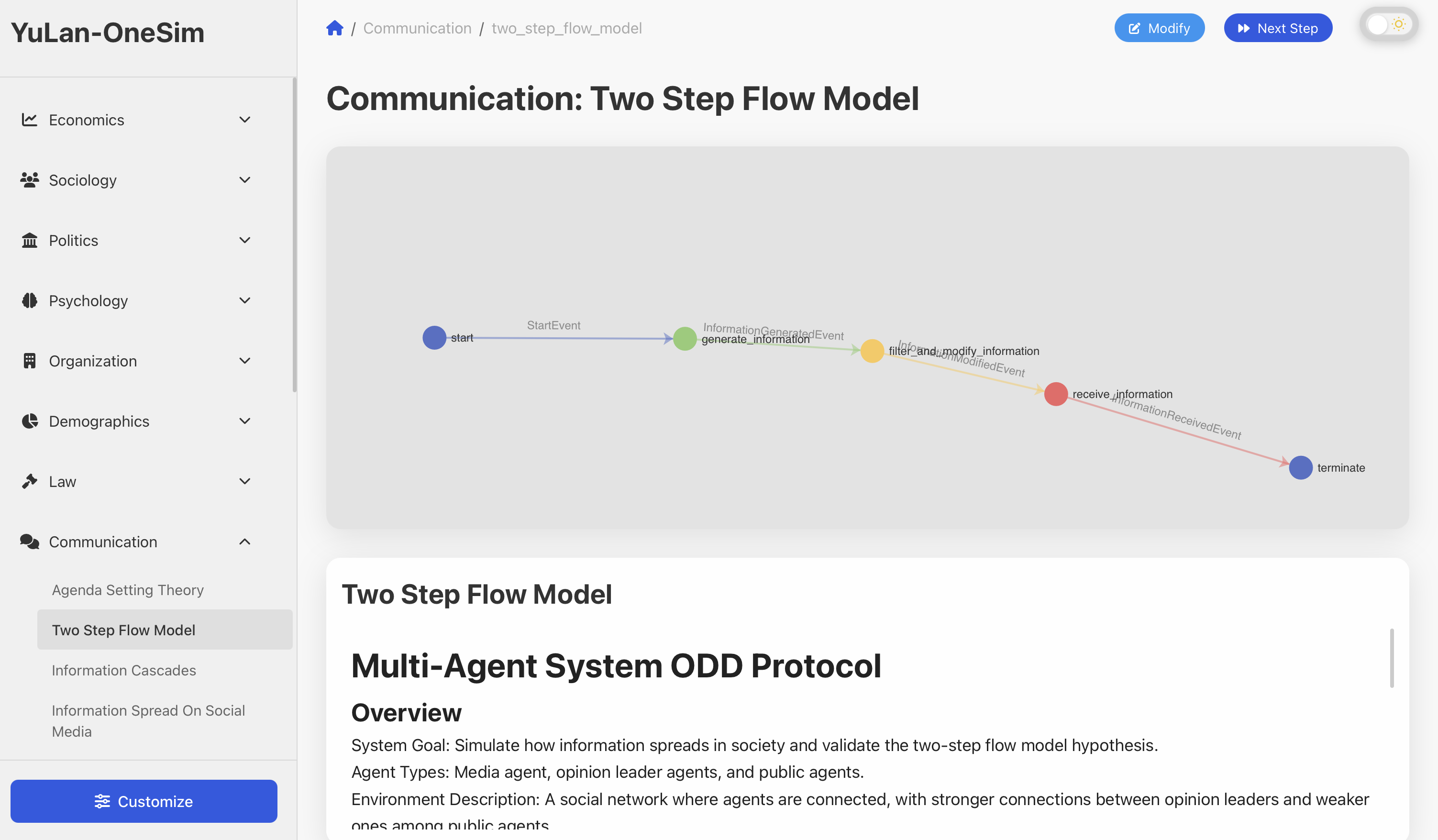}
    \caption{Scenario selection interface showing pre-built scenarios organized by domain.}
    \label{fig:scenario_selection}
\end{figure}

\begin{figure}[htbp]
    \centering
    \includegraphics[width=0.9\textwidth]{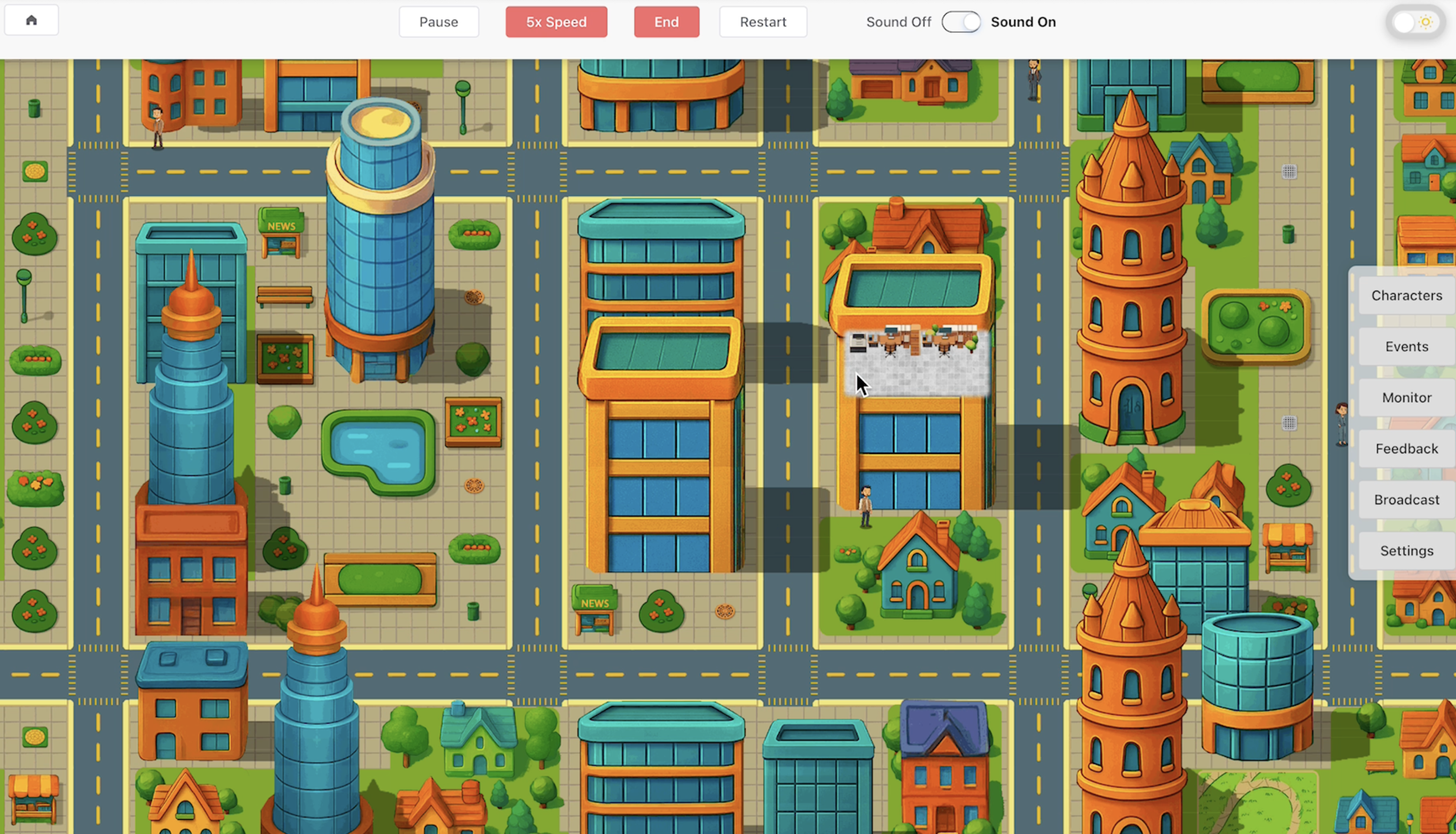}
    \caption{Simulation execution interface showing the virtual environment with buildings and agents.}
    \label{fig:simulation_interface}
\end{figure}

After the simulation begins, agents move and interact within the constructed virtual city. Buildings and streets are populated with agents, as shown in Figure~\ref{fig:simulation_interface}.

\begin{figure}[htbp]
    \centering
    \includegraphics[width=0.9\textwidth]{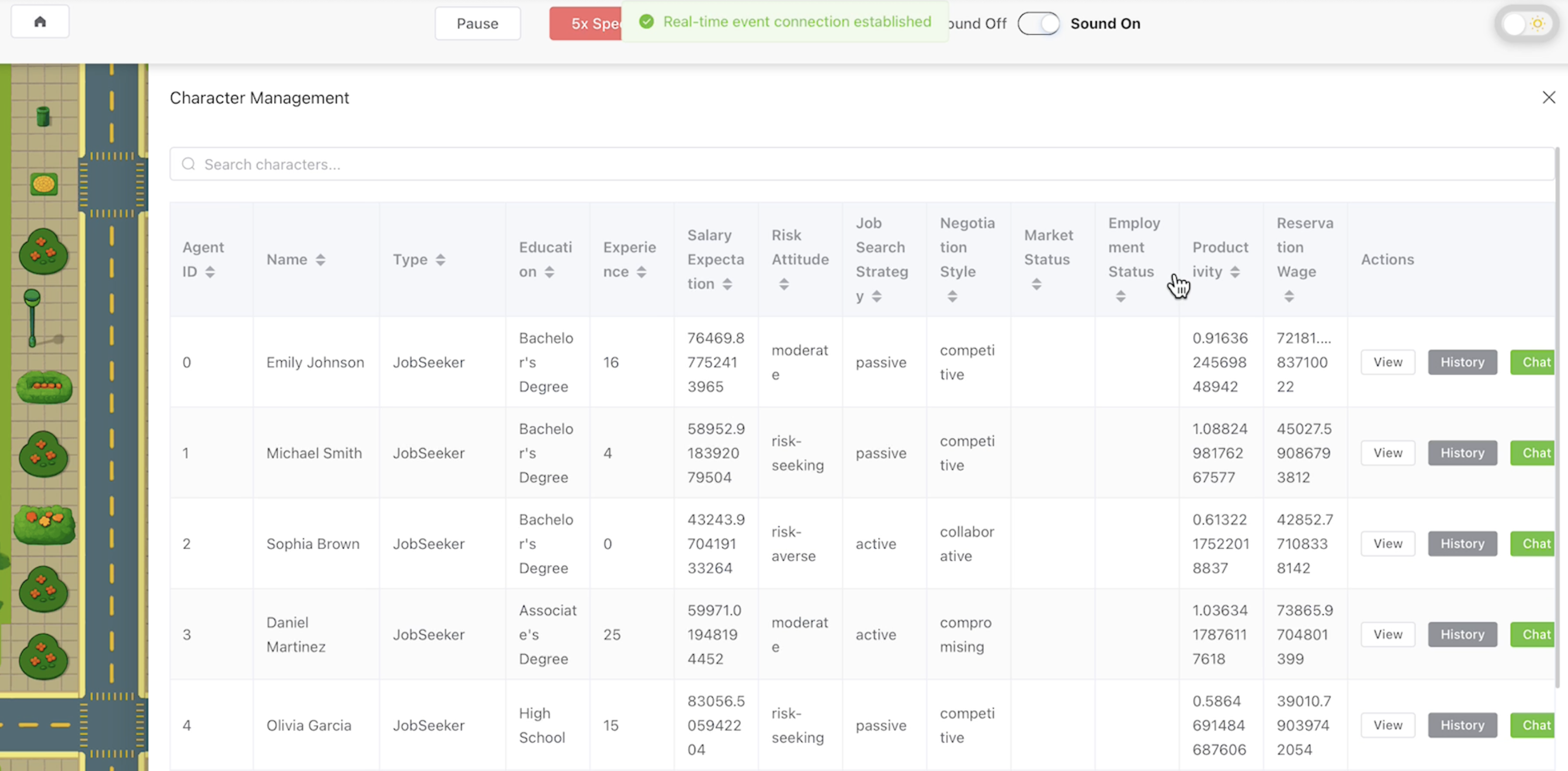}
    \caption{Character management interface showing detailed information about all agents in the simulation.}
    \label{fig:agent_profiles}
\end{figure}

We can view information about all agents in the simulation or focus on individual agents. When viewing individual information, the view locks onto the target, displaying the agent's ID, type, and profile. Agent profiles can be manually modified at any time, and users can individually converse with each agent, as shown in Figure~\ref{fig:agent_profiles}.

\begin{figure}[htbp]
    \centering
    \includegraphics[width=0.9\textwidth]{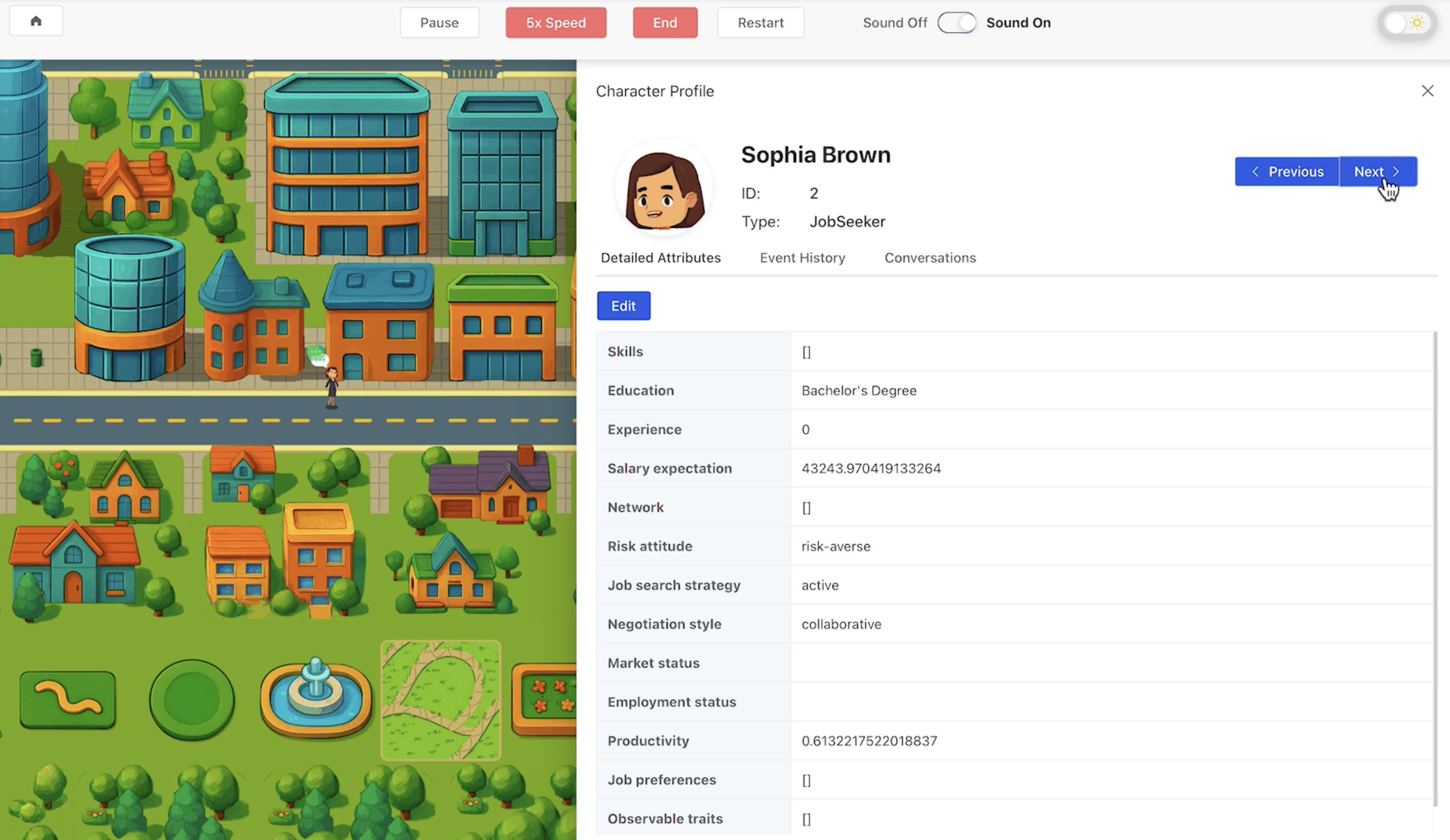}
    \caption{Individual agent profile view showing detailed attributes and allowing for profile editing.}
    \label{fig:individual_profile}
\end{figure}

\begin{figure}[htbp]
    \centering
    \includegraphics[width=0.9\textwidth]{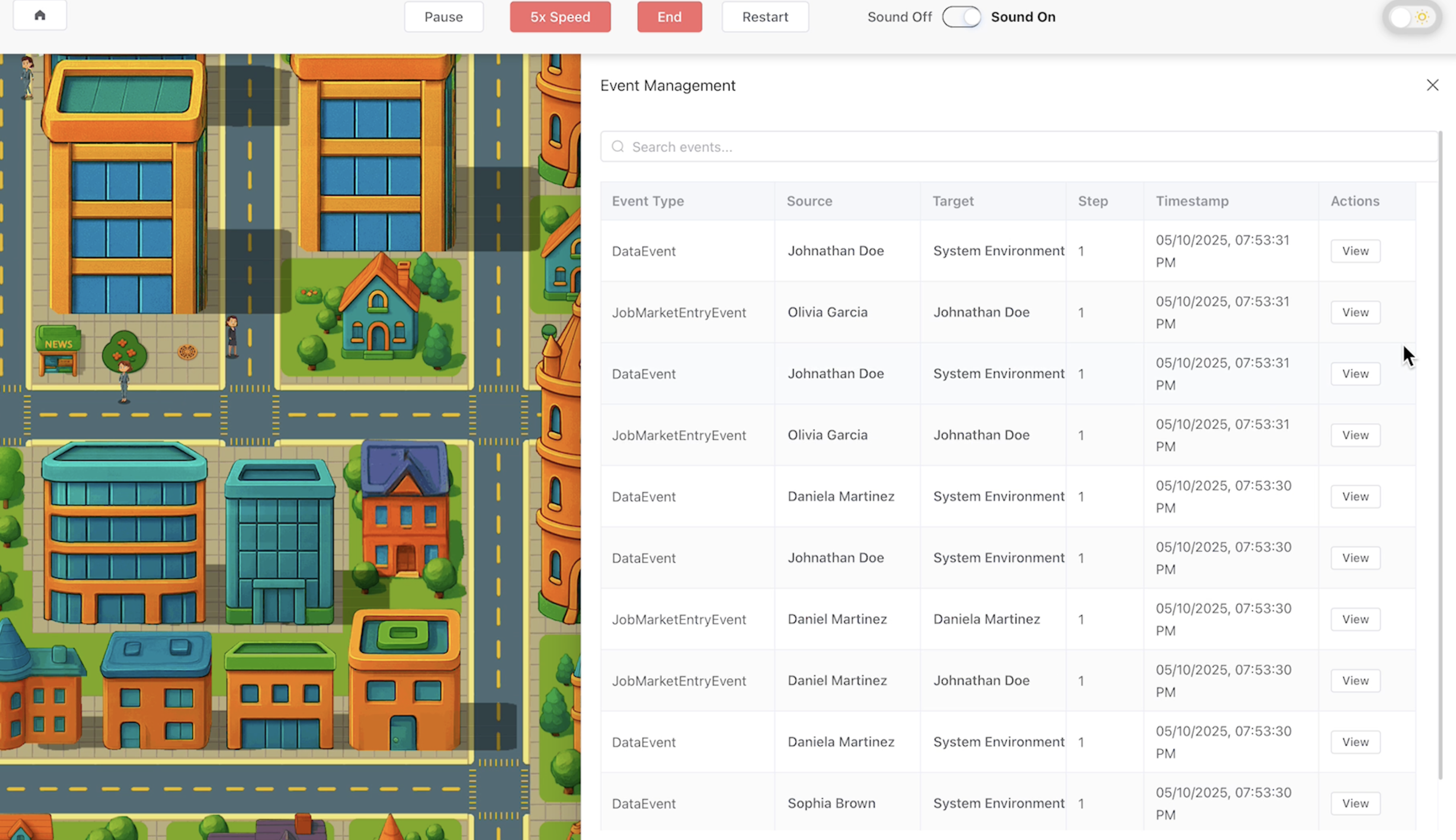}
    \caption{Event management interface displaying all events occurring within the simulation in real-time.}
    \label{fig:event_details}
\end{figure}

Our framework is event-driven and responsive. Users can see real-time details of events occurring in the simulation scenario, as shown in Figure~\ref{fig:event_details}.

\subsection{Simulation result feedback}
\begin{figure}[htbp]
    \centering
    \includegraphics[width=0.9\textwidth]{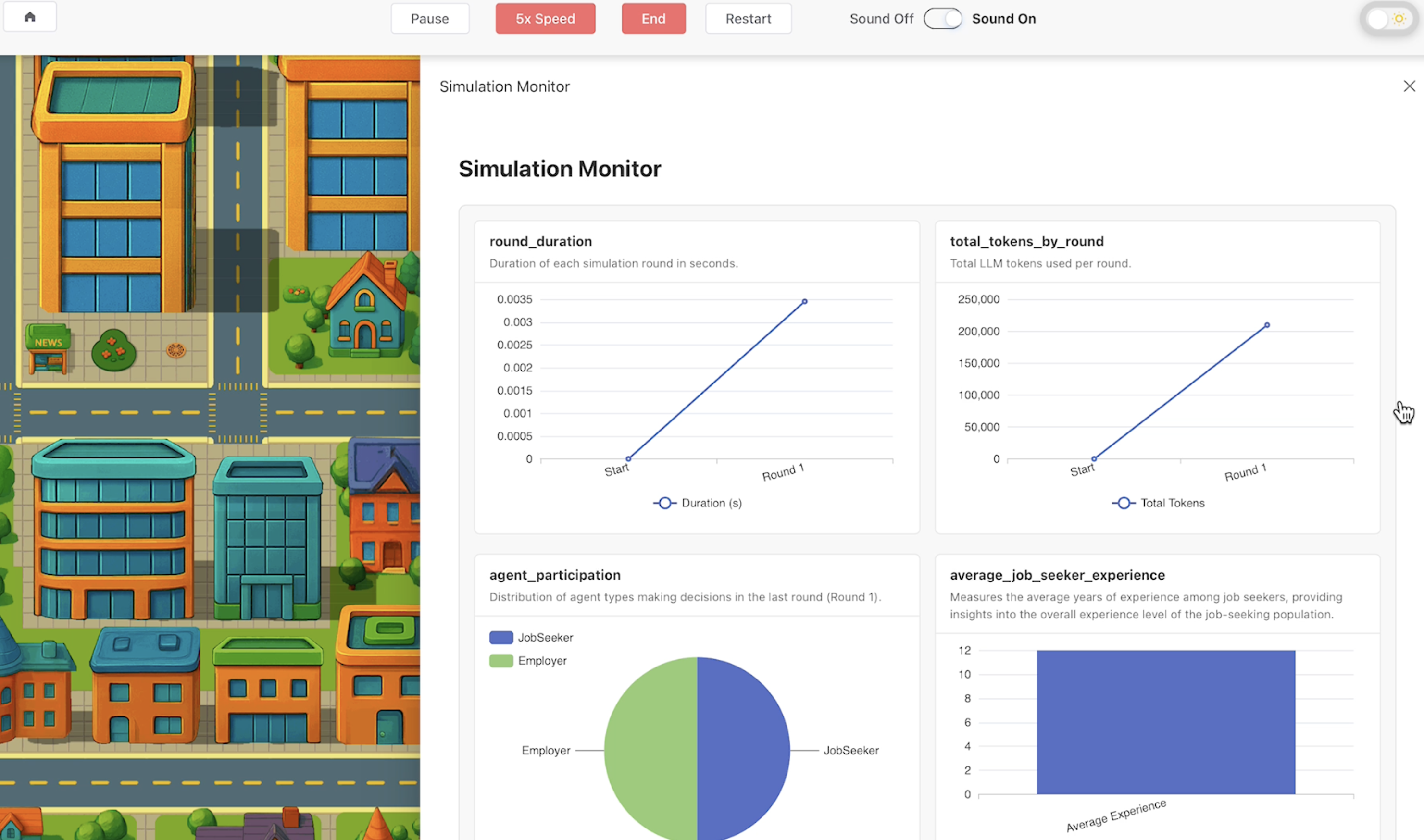}
    \caption{Simulation monitoring interface showing real-time metrics of the simulation progress.}
    \label{fig:simulation_monitor}
\end{figure}

During simulation, we monitor general metrics in real-time, such as simulation time and token consumption. We also monitor scenario-specific metrics like employment rate, displaying them in chart form, as shown in Figure~\ref{fig:simulation_monitor}.

\begin{figure}[htbp]
    \centering
    \includegraphics[width=0.9\textwidth]{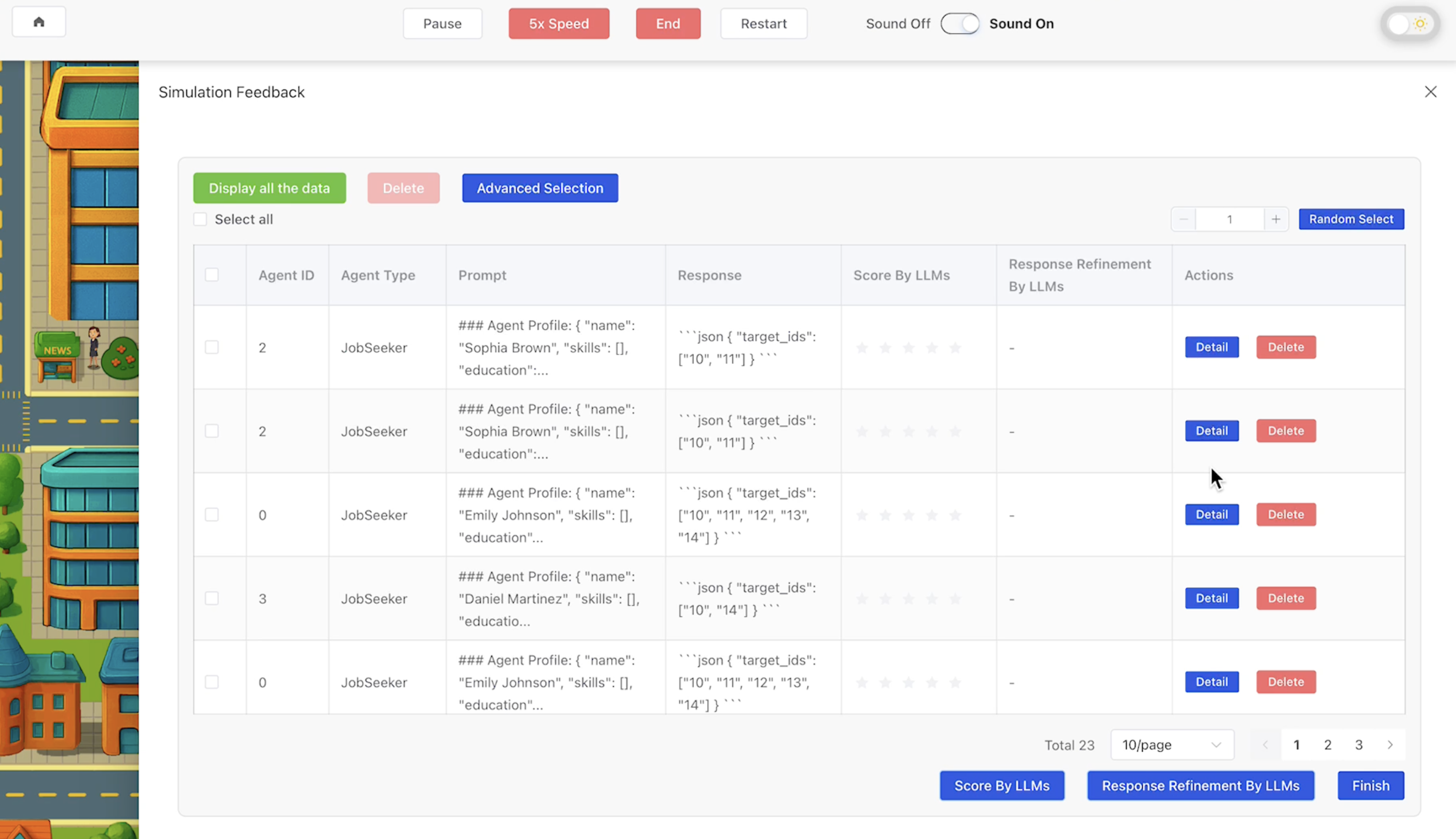}
    \caption{Feedback collection interface allowing for the evaluation and refinement of LLM responses.}
    \label{fig:feedback_interface}
\end{figure}

We also collect prompts received by LLMs and their responses during simulation. Users can select data for LLM scoring or modification to obtain improved responses, which can be exported in dataset format for storage, as shown in Figure~\ref{fig:feedback_interface}.

\begin{figure}[htbp]
    \centering
    \includegraphics[width=0.9\textwidth]{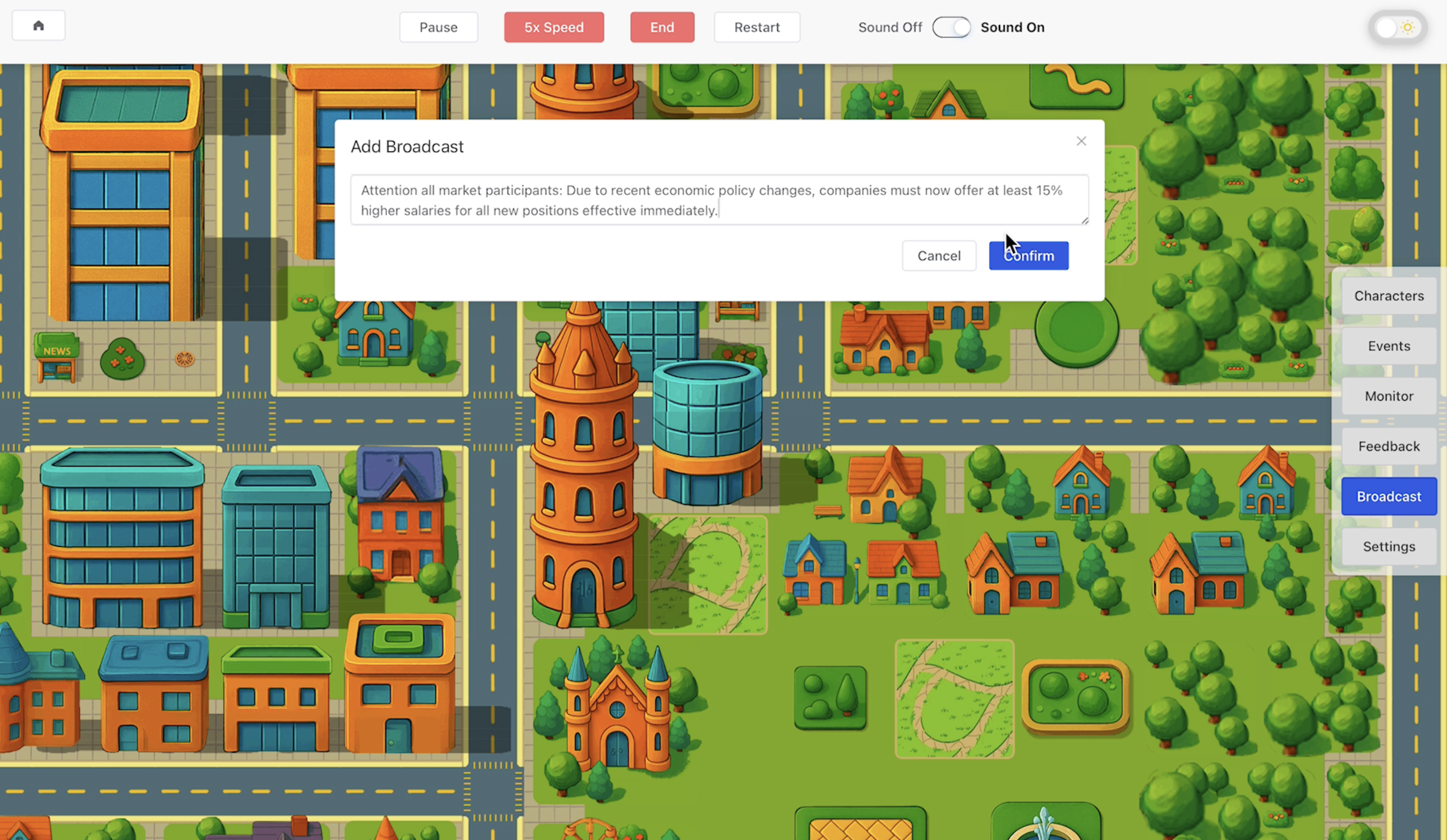}
    \caption{Broadcast interface for sending global announcements to all agents in the simulation.}
    \label{fig:broadcast_interface}
\end{figure}

Additionally, the simulation system supports flexible control mechanisms, allowing users to pause, resume, or stop at any time. Users can also broadcast events to all agents, as shown in Figure~\ref{fig:broadcast_interface}.

    % \subsection{Simulation Result Feedback}

\end{document}